\documentclass{article}

\PassOptionsToPackage{numbers, compress, sort}{natbib}
    \usepackage[final]{neurips_2021}

\usepackage[utf8]{inputenc} % allow utf-8 input
\usepackage[T1]{fontenc}    % use 8-bit T1 fonts
\usepackage{hyperref}       % hyperlinks
\usepackage{url}            % simple URL typesetting
\usepackage{booktabs}       % professional-quality tables
\usepackage{amsfonts}       % blackboard math symbols
\usepackage{nicefrac}       % compact symbols for 1/2, etc.
\usepackage{microtype}      % microtypography
\usepackage{xcolor}         % colors
\usepackage{dsfont}         % colors
\usepackage{amsmath}         % colors
\usepackage{graphicx}
\usepackage{subcaption}
\usepackage{wrapfig}

\newcommand{\rococo}{\textsc{NOOCh}}

\newcommand{\one}{\mathds{1}}

\title{Identifying and Benchmarking Natural Out-of-Context Prediction Problems}
\author{%
  David Madras\\
  University of Toronto\\
  Vector Institute \\
  \texttt{madras@cs.toronto.edu} \\
  \And
  Richard Zemel\\
  University of Toronto\\
  Vector Institute \\
  Columbia University \\
  \texttt{zemel@cs.toronto.edu} \\
}

\begin{document}

\maketitle

\begin{abstract}
  Deep learning systems frequently fail at out-of-context (OOC) prediction, the problem of making reliable predictions on uncommon or unusual inputs or subgroups of the training distribution.
%   , which is of paramount importance in practical applications.
  To this end, a number of benchmarks for measuring OOC performance have recently been introduced.
  In this work, we introduce a framework unifying the literature on OOC performance measurement, and demonstrate how rich auxiliary information can be leveraged to identify candidate sets of OOC examples in existing datasets.
  We present {\rococo}: a suite of naturally-occurring ``challenge sets'', and show how varying notions of context can be used to probe specific OOC failure modes.
  Experimentally, we explore the tradeoffs between various learning approaches on these challenge sets and demonstrate how the choices made in designing OOC benchmarks can yield varying conclusions.
\end{abstract}

\section{Introduction} \label{sec:intro}

People often find context useful for prediction, both for improving accuracy and processing efficiency \citep{chun2000contextual}.
However, deep learning systems frequently over-rely on context cues \citep{geirhos2020shortcut,geirhos2018imagenet,lovering2021predicting}, which can lead to poor performance on out-of-context (OOC) examples, when contextual information is misleading.
By OOC examples, we mean inputs which are uncommon or unusual with respect to the training distribution;
these can be thought of as sampled from under-represented subgroups, or low (non-zero) density regions, of the training distribution.
OOC is a fundamental problem, a key facet of system robustness.
In safety-critical situations, errors can be problematic; as such, it is important to have reliable methods for measuring how well a model can perform OOC.
Furthermore, given the rise of larger models and datasets \citep{kaplan2020scaling}, there is a need for scalable approaches to OOC evaluation --- even if manual evaluation of ``corner cases'' by domain experts may (always) be the gold standard.

A key pre-requisite task to \textit{evaluating} OOC performance is \textit{identifying} which examples should be considered OOC.
This identification task is a challenging one in and of itself: in a natural image, ``context'' can be varied, complex and high-dimensional \citep{torralba2003contextual,xiao2020noise,peters2016causal,bobick1995using}.
Therefore, any evaluation method intending to measure a model's OOC performance must (implicitly) select a specific notion of ``OOC performance''.
Indeed, since deep learning typically yields underspecified models \citep{d2020underspecification}, it is plausible that different choices may yield different measurements.
Common approaches include generating semi-synthetic data to simulate the effect of a shift in some salient feature \citep{kim2019learning,xiao2020noise,sagawa2019distributionally}, or using some auxiliary information to guide choices about what a reasonable OOC set should be \citep{koh2020wilds,hendrycks2021natural}.

In this work, we develop a conceptual framework for identifying sets of OOC examples in existing datasets.
We show how our framework unifies and generalizes prior literature on OOC performance measurement, and allows us to utilize more complex, structured annotated data to define various formulations of OOC performance.
%notions of ``OOC performance''.
We demonstrate this framework's effectiveness through uncovering two OOC ``challenge sets'' \citep{isabelle2017challenge} within an existing benchmark, each corresponding to differing notions of context.
We show how our framework enables scalable and targeted measurement of models' OOC performance through clarifying the relationship between the concept of 
OOC performance
%``OOC performance'' 
and its implementation, allowing for clearer insight on current approaches as well as opportunities for improvement.
Our contributions are as follows:
\begin{itemize}
    \item We present {\rococo} (Naturally-Occurring Out-of-context Challenge sets), a suite of 
    %``challenge sets'' 
    challenge sets for evaluating performance on naturally-arising OOC problems, available at \url{https://github.com/dmadras/nooch};
    \item We develop a conceptual framework for automatically identifying OOC challenge sets from existing data by leveraging known underlying structure;
    \item  We demonstrate and contrast two instantiations of this framework using two notions of %``context''
    context, defining concepts of ``hard positives'' and ``hard negatives'' in the OOC setting; and
    \item We quantitatively analyze the performance of several methods from the robust learning literature on these challenge sets, exploring the tradeoffs inherent in different approaches to OOC performance measurement; and qualitatively demonstrate how rich notions of context can yield insight into system robustness.
    %rich investigation of OOC errors.
\end{itemize}

\section{Measuring Out-of-Context Performance} \label{sec:measurement}

Intuitively, a model which has good OOC performance should be able to maintain good performance under unusual or perturbed contextual conditions.
We distinguish this from the out-of-distribution (OOD) problem \citep{hendrycks2016baseline}, which is usually concerned with inputs from a different domain than the training set.
Rather, the OOC prediction problem is more similar to subgroup robustness or distributional shift, where a model must perform well on uncommon input regions at training time.
% One desirable property of an ML method is for it to output models which have good ``OOC performance''.
% Intuitively, this simply means that we would like a model, once learned, to be able to maintain good performance under unusual or perturbed contextual conditions.
However, even after drawing this distinction, the notion is ill-defined: context may refer to concepts as varied as object relationships \citep{torralba2003contextual}, image backgrounds \citep{xiao2020noise}, experimental settings \citep{peters2016causal}, or world models \citep{bobick1995using}. 
Furthermore, even fixing a notion of context, the criterion for what should make something out-of-context (OOC) is still unclear.
For instance, \citet{peters2016causal} focus on previously unobserved contexts (i.e. environments), whereas \citet{xiao2020noise} are concerned with unusual contexts given the class of interest (i.e. perturbations to image background).

Clearly, defining a benchmark to measure a method's OOC performance requires a number of design choices, which has fostered
%enabled 
a recent proliferation of OOC benchmarks.
We note in particular, one of the key choices is around the usage of \textit{auxiliary information}.
Across the literature on OOC performance measurement, there are a plethora of approaches to defining OOC criteria using some type of auxiliary information $C$.
For the purposes of algorithm design, $C$ may be assumed to be available at training, validation, and/or test time, or not at all --- however, at the the time of benchmark design, it is available on a sufficiently large portion of the collected dataset to guide the designers' choices about what a suitable OOC criterion should be.
Examining the current literature on measuring OOC performance, we identify the following as a unifying framework:
\begin{enumerate}
\item Identify some existing auxiliary information $C$, a variable which takes some value on many (or all) examples and specifies some underlying structure in the data.
\item Select a notion of ``OOC'' (e.g. ``images with misleading backgrounds are OOC'', ``examples from unfamiliar time periods are OOC'') and define an ``OOC criterion'' by choosing a binary function $\phi$ of $C$.
\item Restrict the test set to those examples where $\phi = 1$. Optionally, also restrict the training set to those examples where $\phi = 0$.
\end{enumerate}

\begin{table}[]
\centering
\begin{tabular}{ccc}
\hline
\textbf{Dataset}                                                           & \textbf{Auxiliary Information $C$ }                                                           & \textbf{OOC function $\phi$}                  \\ \hline
Waterbirds \citep{sagawa2019distributionally} & 1 if background is water (binary) & $C \neq Y$                 \\ 
iWildCam2020-Wilds\citep{koh2020wilds}        & camera trap  ID (categorical)                                                & $C \notin \{1 \dots 245\}$         \\ 
FMoW-Wilds \citep{koh2020wilds}               & time stamp (ordinal)                                        & $C_{time} \geq 2013$ \\ 
Imagenet-A \citep{hendrycks2021natural}       & max NLL of ensemble (continuous)                                           & $C \geq -\log{(0.15)}$ \\
Breeds \citep{santurkar2020breeds}       & subclass (categorical)                                          & $C \in \textrm{target subset}$ \\
\hline
\end{tabular}
\caption{Examples of OOC benchmarks from the literature under our framework.
The right-most column lists the condition under which $\phi = 1$.
The OOC function $\phi$ in \citep{hendrycks2021natural} has several additional filtering steps, some heavily manual.
}
\label{tab:framework}
\end{table}

We show in Table \ref{tab:framework} how a range of prior literature leverages auxiliary information to define OOC criteria.
We can think of $C$ as providing benchmark designers with some type of ``inductive bias'' around what should be considered OOC for a given benchmark.
The above framework implies that there is a diversity of OOC criteria which can be defined over any dataset, and this class is as broad as the class of functions $\phi$ which can be defined over the available auxiliary information.
% While past benchmarks are valuable, they tend to rely on one-dimensional or categorical auxiliary data, suggesting that the class of operationalizations for measuring OOC performance is much broader and more flexible than previously explored.
% \DM{Clarify here: we're presenting two examples of such an approach}
In the rest of the paper, we take advantage of the flexibility of this framework to give two examples of such an approach.
We show that by leveraging more complex annotated structure, we can create multiple OOC benchmarks from an existing dataset using multiple criteria for what should be considered ``OOC''.
% use multiple operationalizations of OOC performance to create multiple benchmarks from an existing dataset.
We trace out how the choices made in designing these criteria correspond to different notions of context, and demonstrate experimentally that these yield varying measurements of OOC performance.

\begin{figure}[t!] % "[t!]" placement specifier just for this example
\begin{subfigure}{0.33\textwidth}
\includegraphics[width=\linewidth]{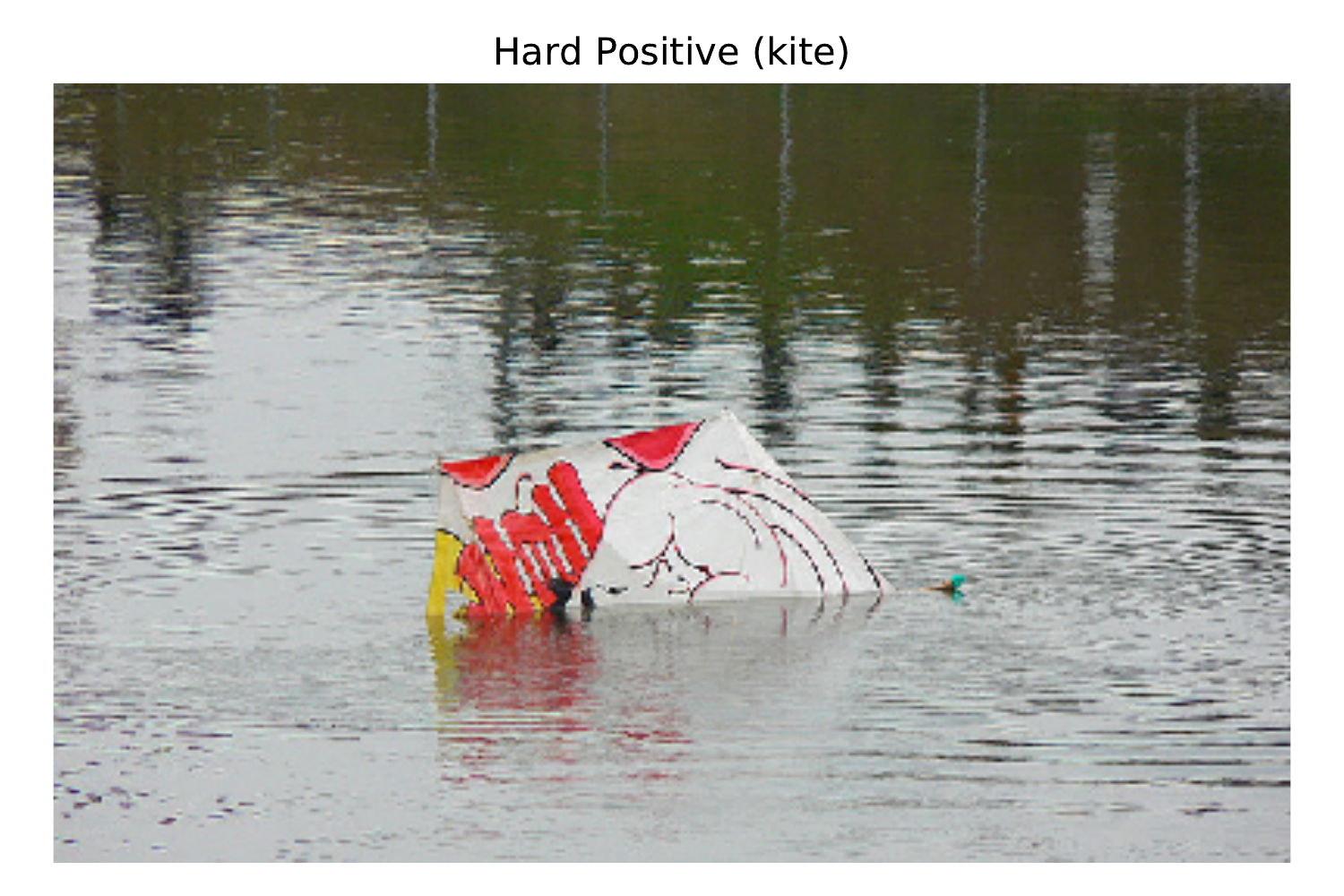}
% \caption{First subfigure} \label{fig:a}
\end{subfigure}\hspace*{\fill}
\begin{subfigure}{0.33\textwidth}
\includegraphics[width=\linewidth]{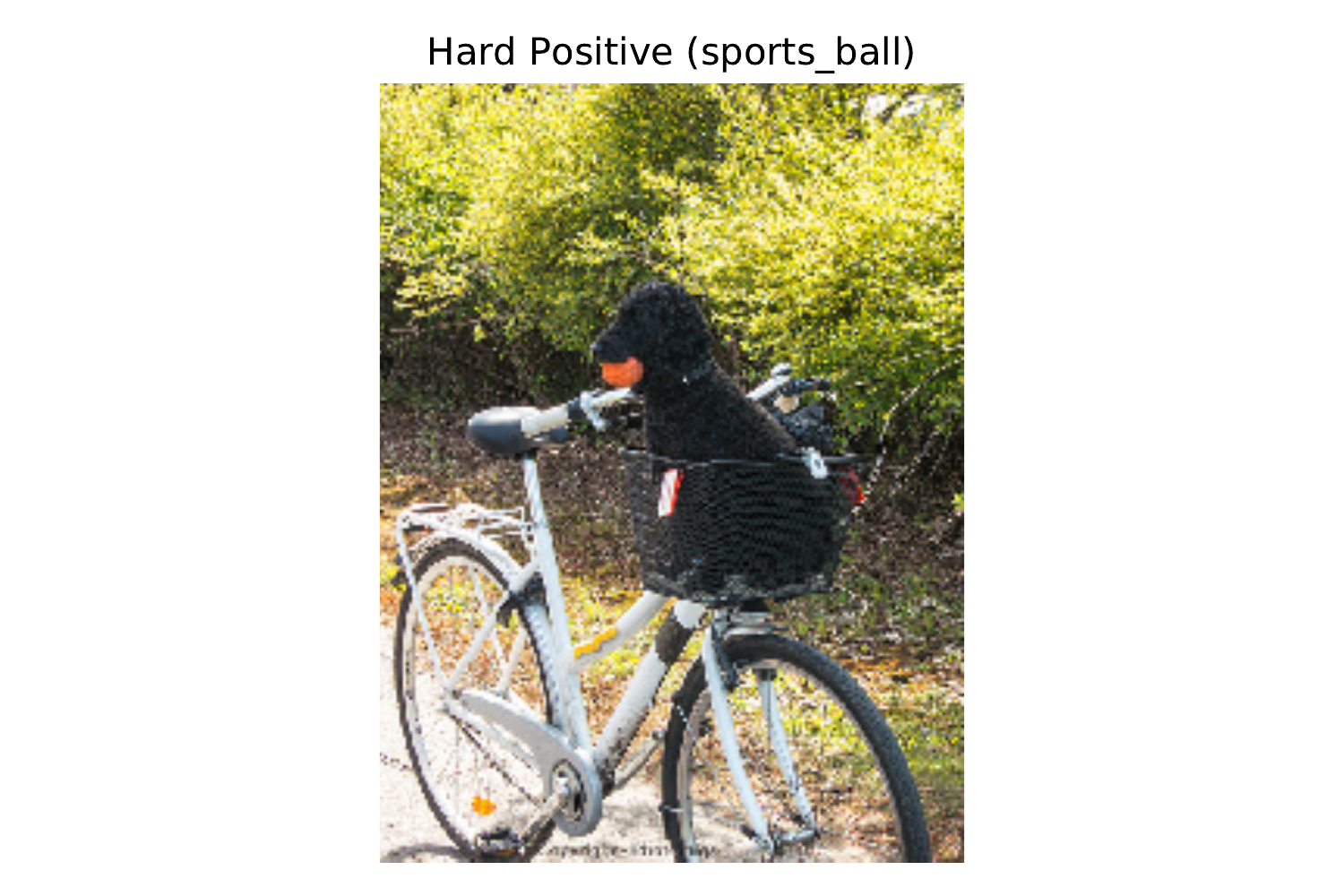}
% \caption{Second subfigure} \label{fig:b}
\end{subfigure}\hspace*{\fill}
\begin{subfigure}{0.33\textwidth}
\includegraphics[width=\linewidth]{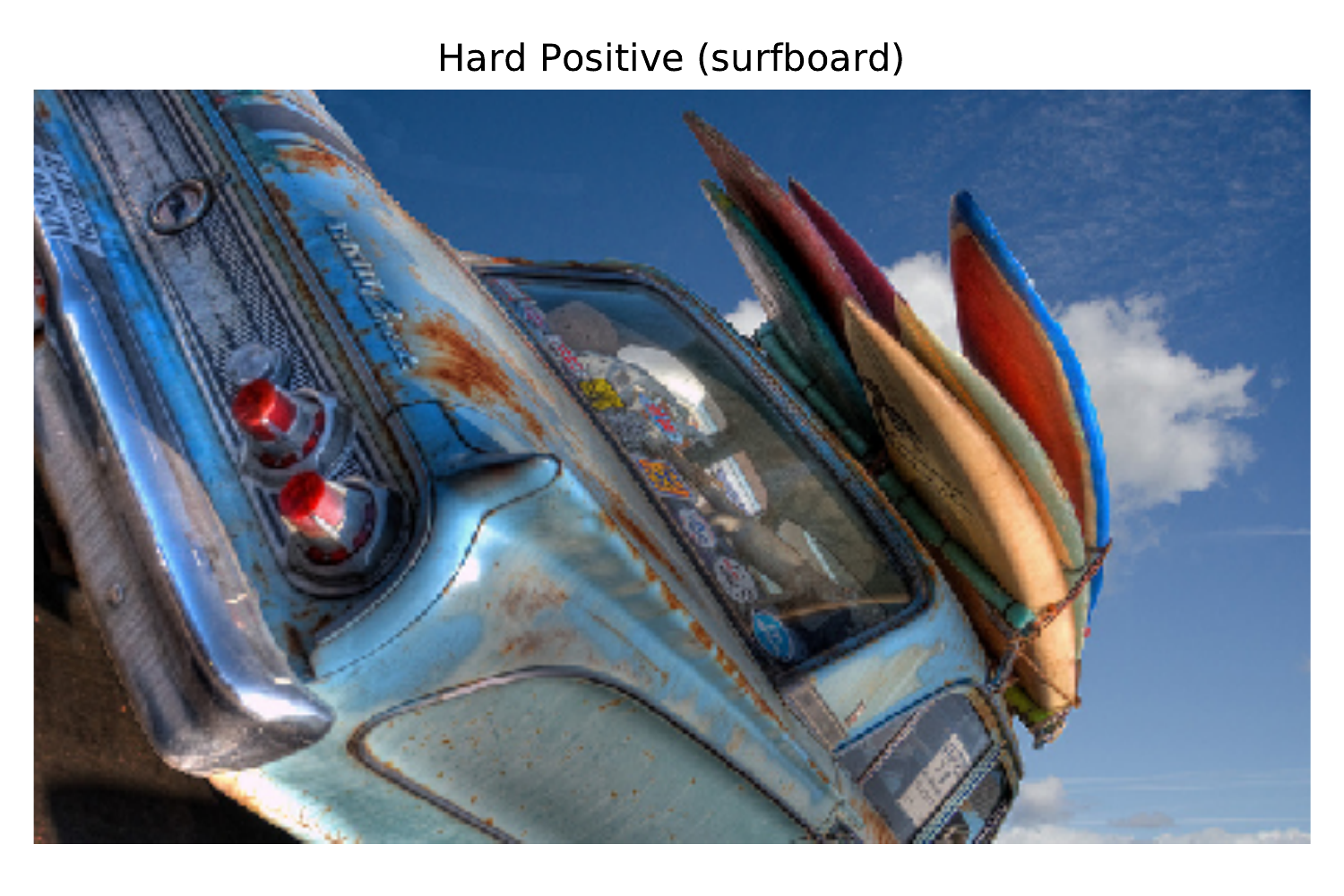}
% \caption{Second subfigure} \label{fig:b}
\end{subfigure}

\medskip
\begin{subfigure}{0.33\textwidth}
\includegraphics[width=\linewidth]{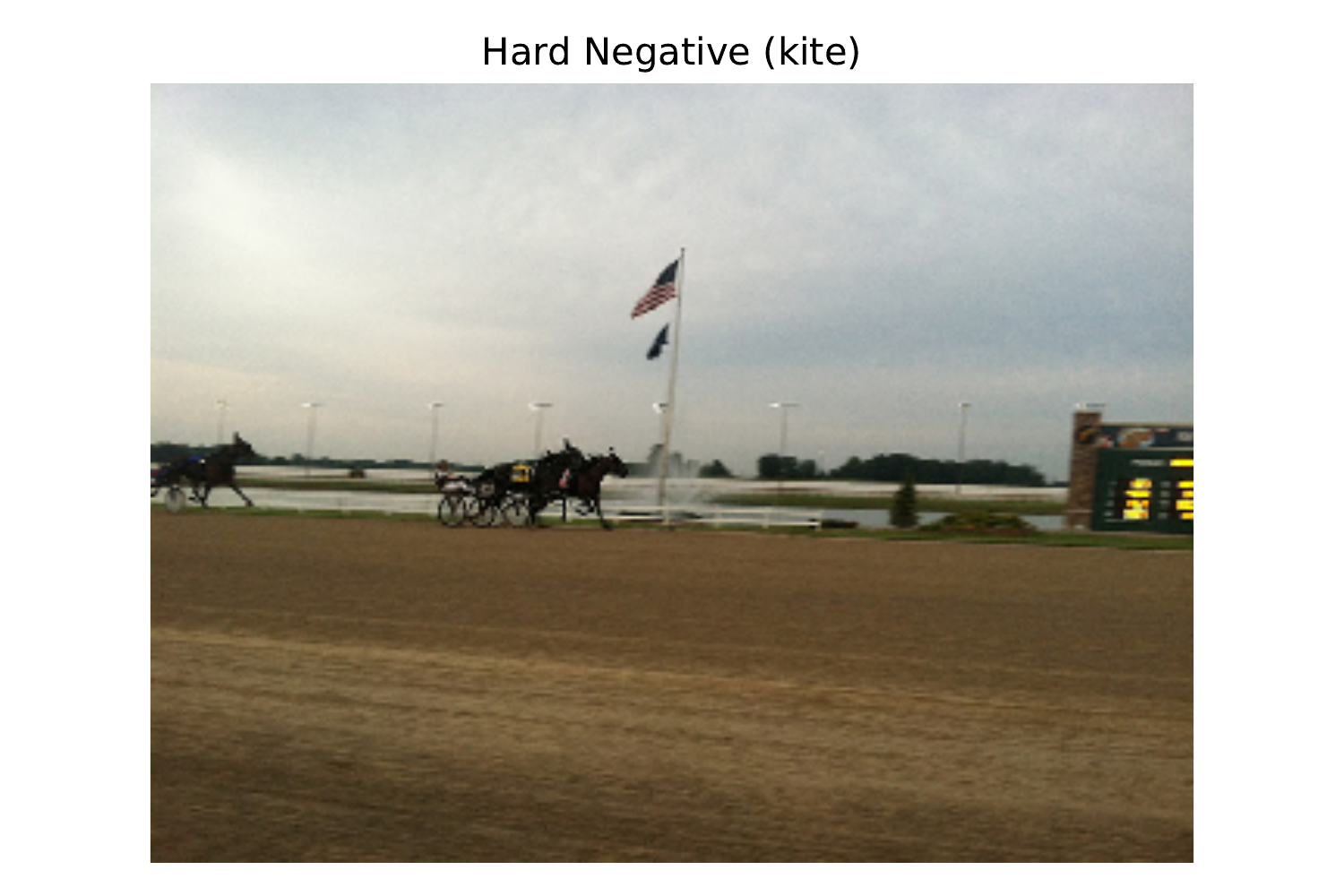}
% \caption{First subfigure} \label{fig:a}
\end{subfigure}\hspace*{\fill}
\begin{subfigure}{0.33\textwidth}
\includegraphics[width=\linewidth]{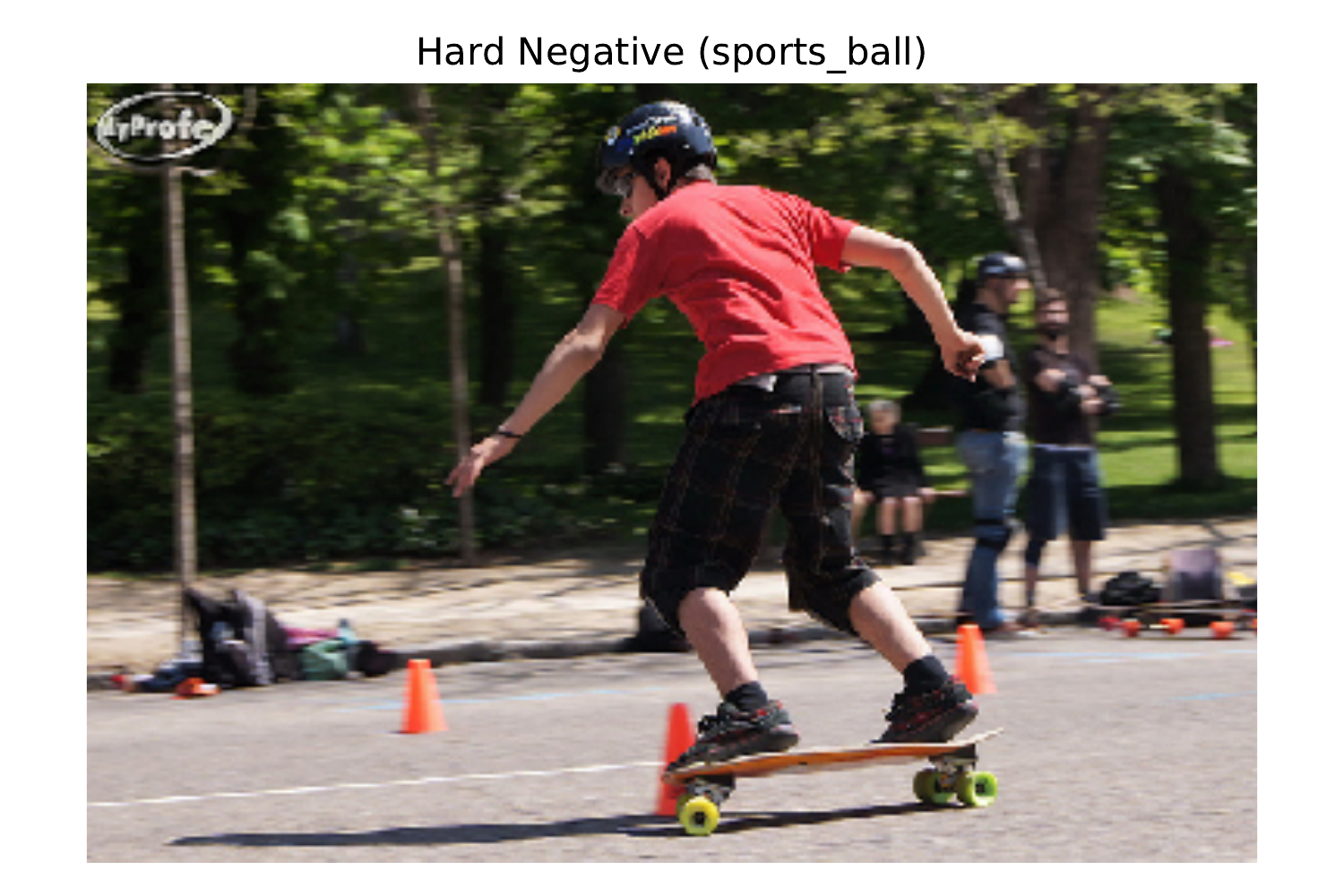}
% \caption{Second subfigure} \label{fig:b}
\end{subfigure}\hspace*{\fill}
\begin{subfigure}{0.33\textwidth}
\includegraphics[width=\linewidth]{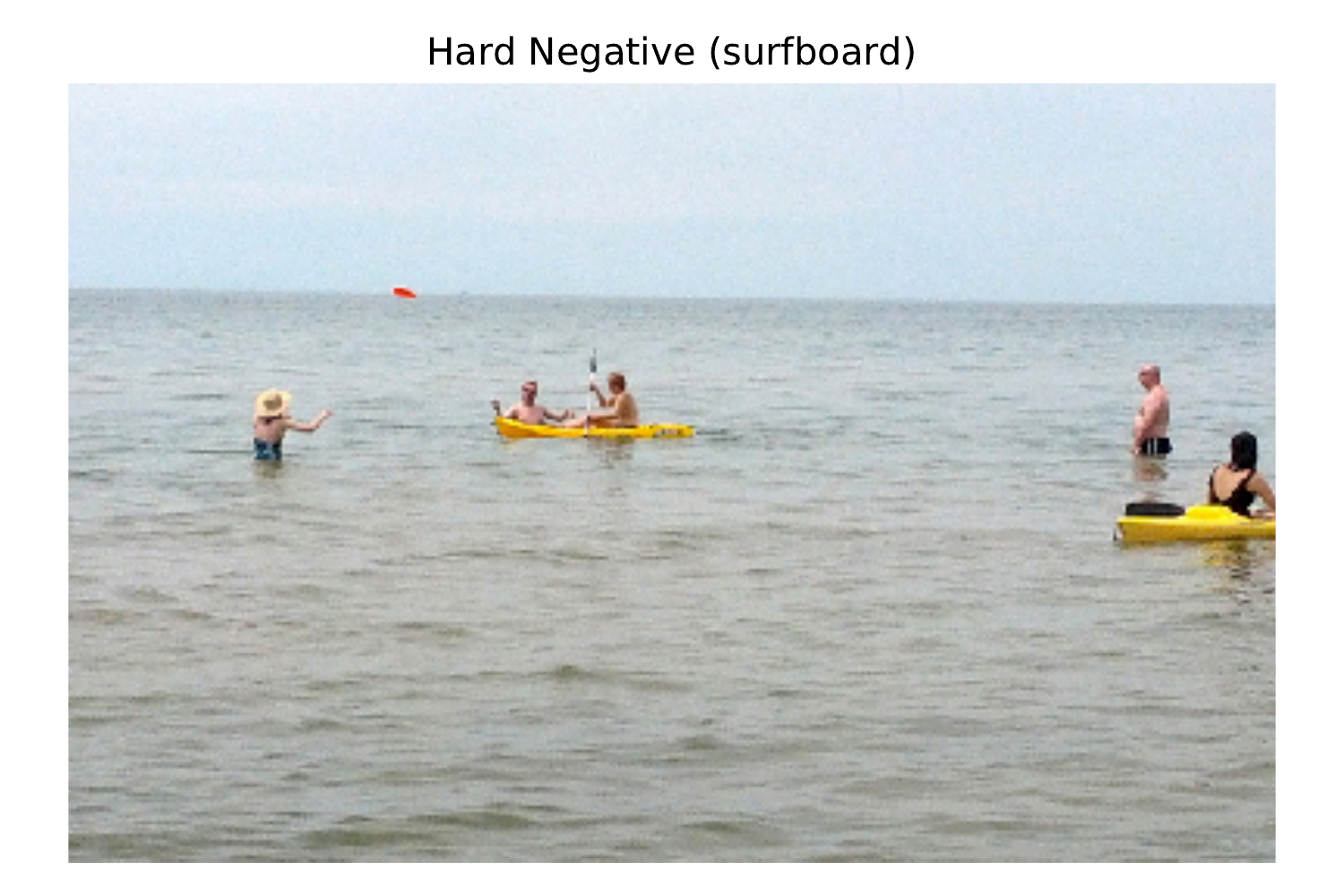}
% \caption{Second subfigure} \label{fig:b}
\end{subfigure}
\caption{Using the co-occurrence/extractibility (CE) criterion, examples of (0.05, 0.1)-hard positives (top row) and negatives (bottom row) for the classes (L to R):  \texttt{kite, sports\_ball, surfboard}.} \label{fig:benchmark-hard-pos-neg-examples}
\end{figure}

\section{Finding Naturally-Occurring OOC Problems} \label{sec:benchmark}

In this section, we demonstrate concretely how rich auxiliary information can be used to study the way that context shifts arise naturally within an existing computer vision benchmark, and provide two criteria for OOC performance that can be computed from these annotations. 
Throughout, we consider the binary prediction task of determining object presence, a problem where relationships between various objects naturally provide helpful context --- given an image $X$, is an object of class $Y$ present or not?

\subsection{Background: COCO and COCO-Stuff}

The Microsoft Common Objects in COntext dataset (COCO) \citep{lin2014microsoft} is a computer vision dataset consisting of images of natural scenes.
Each image is annotated with instance labels and segmentations for every ``thing'' in the image, as well as several captions describing the content of the scene.
Images usually contain multiple items and as such usually have multiple labels.

However, for the purposes of investigating OOC prediction, many relevant objects are not labelled in COCO; for instance, background objects such as ``sky'' or ``grass'' are not COCO classes.
Fortunately, the COCO-Stuff dataset \citep{caesar2018coco} provides labels and segmentations for all of the ``stuff'' in the images from COCO.
The ``thing'' vs ``stuff'' distinction is a technical one: a thing is an object with a specified size and shape, whereas stuff has no ``defined spatial extent'' \citep{forsyth1996finding}.
Having both thing and stuff labels is essential for understanding model behaviour on OOC examples, since it is exactly these ``stuff'' classes which often (but not always) provide important context cues.
Taken together, the thing and stuff annotations
yield a rich sandbox for queries about the role of context in prediction.
For our purposes, COCO-Stuff contains 171 binary tasks for determining object presence (81 thing classes and 90 stuff classes; we ignore the 1 ``unlabelled'' class).

\subsection{Automatically Identifying OOC Examples: Hard Positives and Negatives}

% As mentioned in Sec. \ref{sec:intro}, context is hard to define.
% We note that different definitions of context correspond to different OOC example-level failure modes.
Here, we consider two contrasting notions of ``OOC'': 1. the presence/absence of frequently co-occurring, easily extractible objects; and 2. an unusual ``gist'' of a scene.
We define these two notions of context below, presenting two criteria for identifying naturally-occurring OOC prediction problems within the existing COCO(-Stuff) dataset, and discussing how we can use annotations as proxies to define an OOC indicator $\phi$.
We identify two types of OOC examples: {\it hard positives}, where the class is present despite an unusual context, and {\it hard negatives}, where the class is \textit{not} present, despite a usual context.

\begin{figure}[t!] % "[t!]" placement specifier just for this example
\begin{subfigure}{0.25\textwidth}
\includegraphics[width=\linewidth]{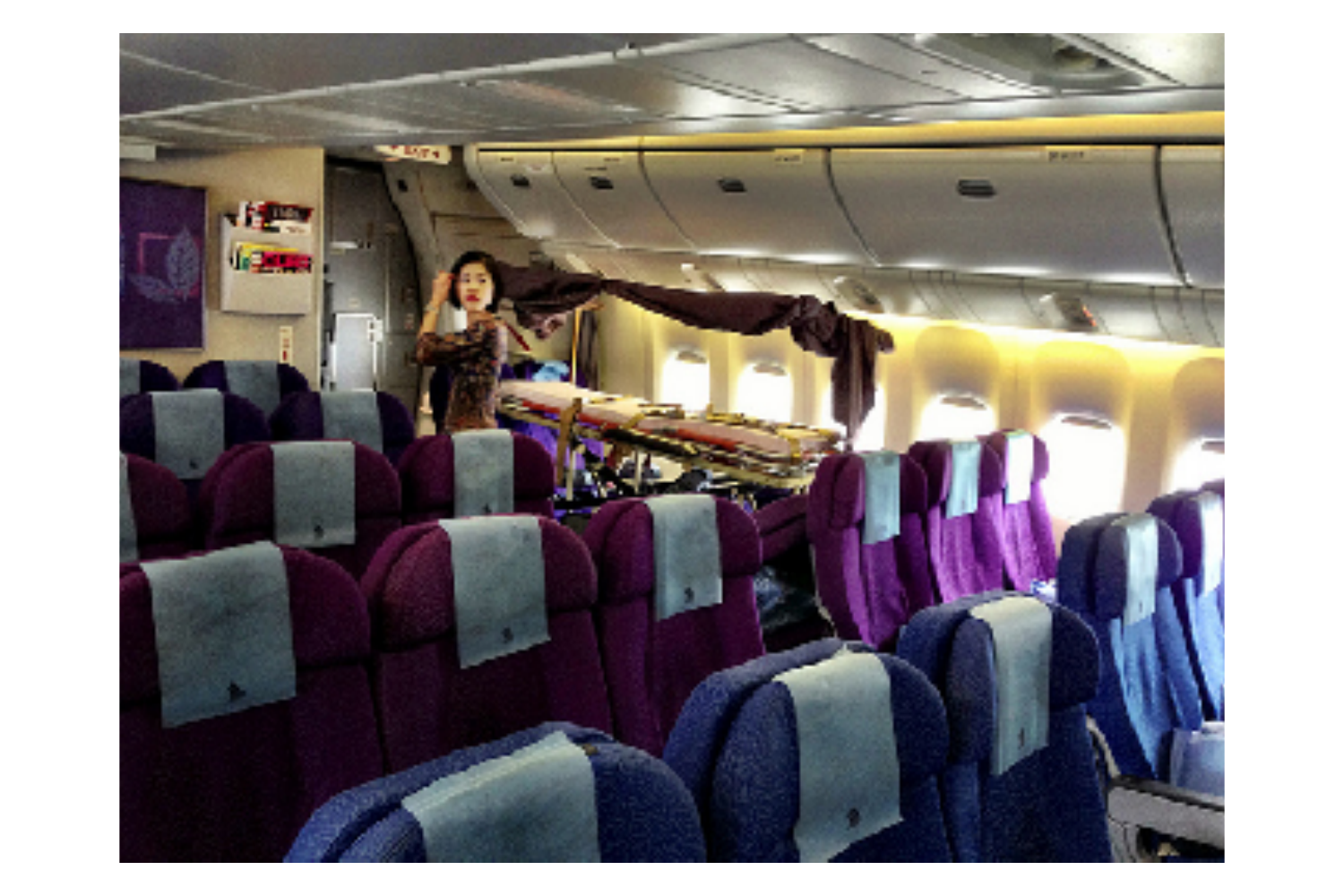}
\caption{hard positive (CE)} \label{fig:a}
\end{subfigure}\hspace*{\fill}
\begin{subfigure}{0.25\textwidth}
\includegraphics[width=\linewidth]{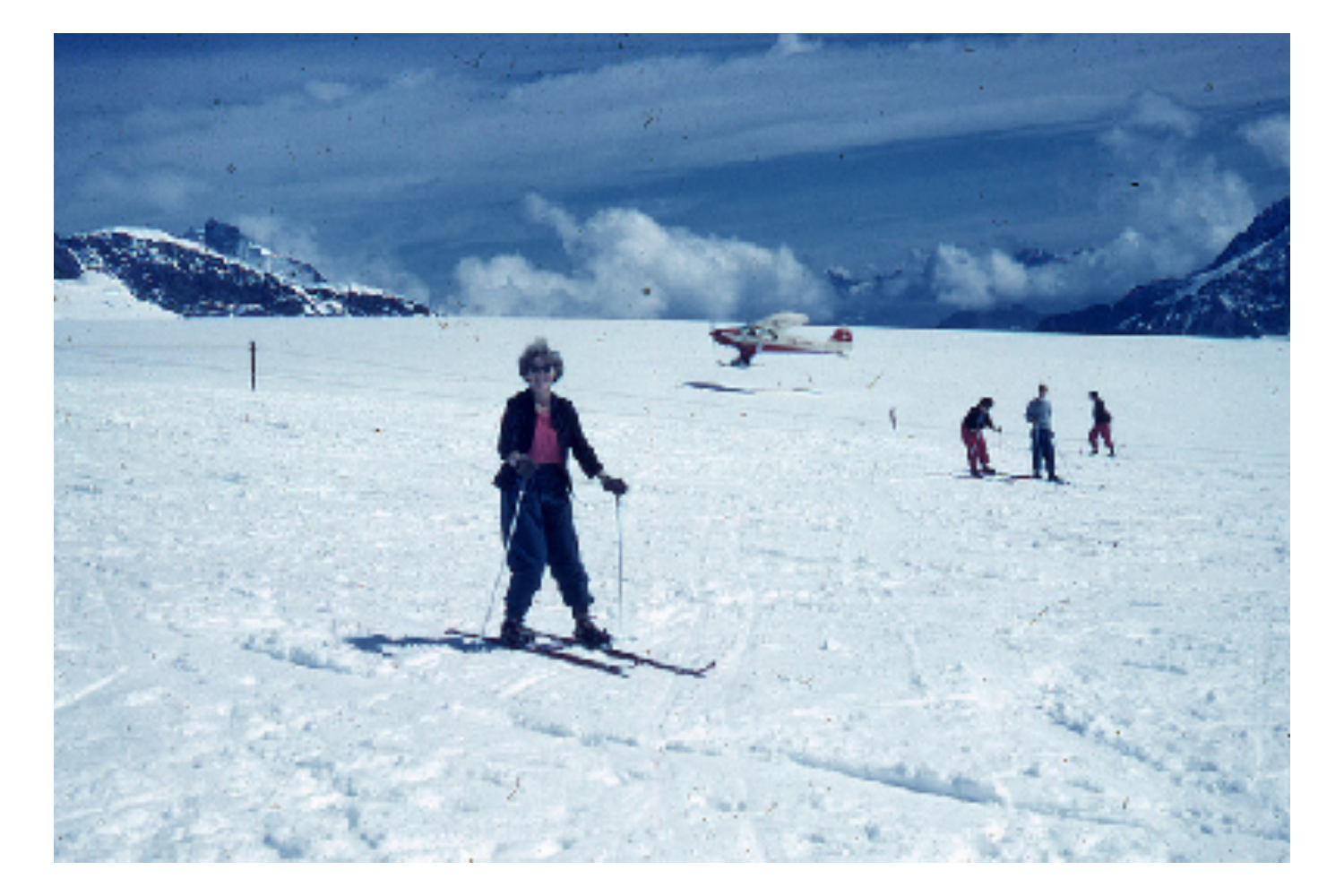}
\caption{hard positive (Gist)} \label{fig:b}
\end{subfigure}\hspace*{\fill}
\begin{subfigure}{0.25\textwidth}
\includegraphics[width=\linewidth]{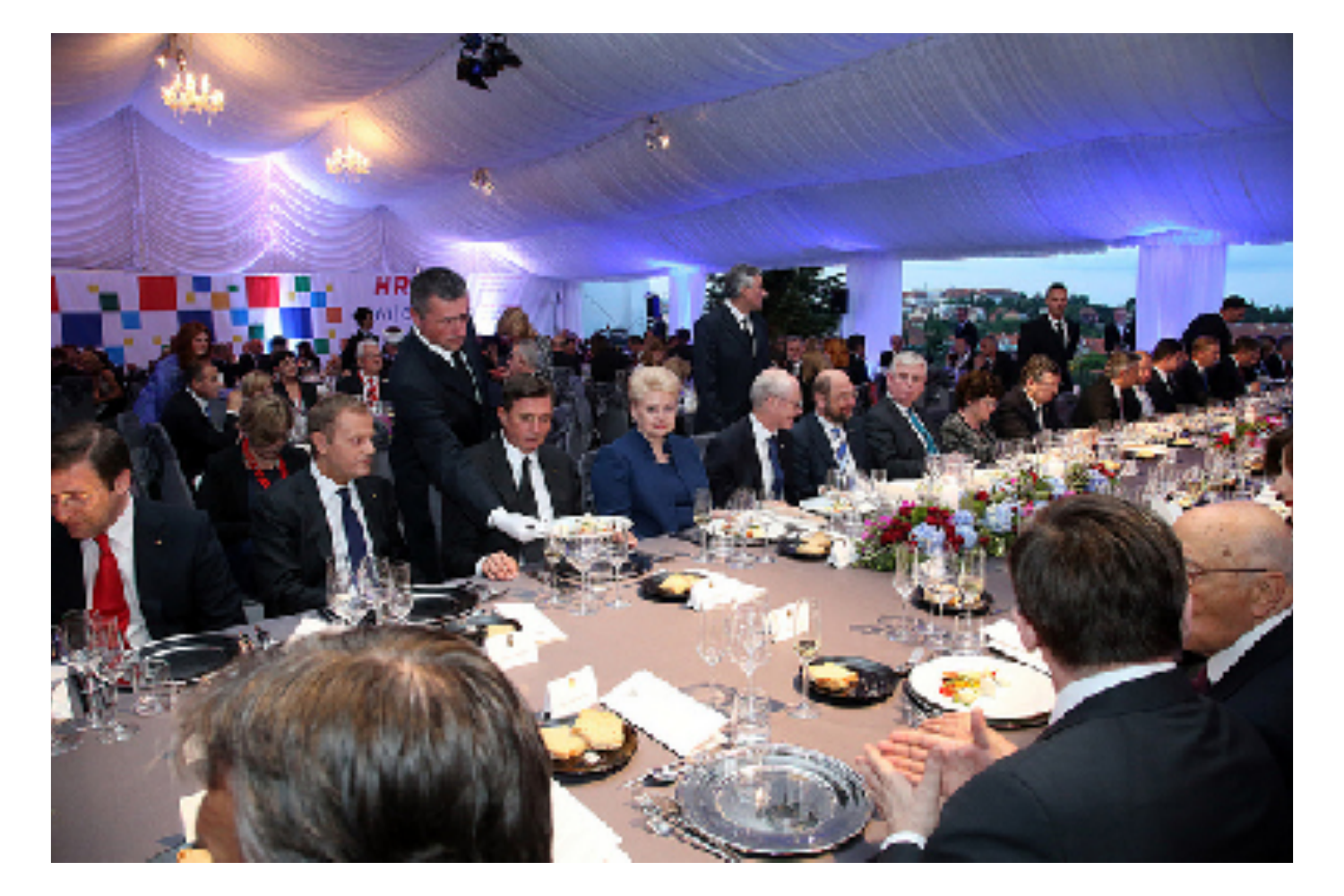}
\caption{hard negative (CE)} \label{fig:a}
\end{subfigure}\hspace*{\fill}
\begin{subfigure}{0.25\textwidth}
\includegraphics[width=\linewidth]{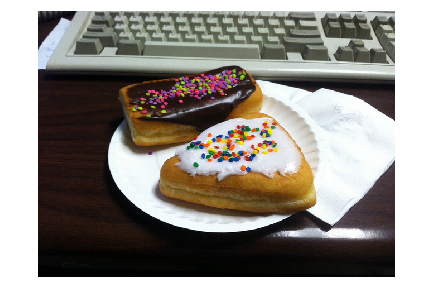}
\caption{hard negative (Gist)} \label{fig:b}
\end{subfigure}
\caption{\small{
To contrast the CE and Gist criteria, we show samples from the \texttt{airplane} (L) and \texttt{bowl} (R) tasks.}
} \label{fig:results-qual-two-criteria}
\end{figure}

\subsubsection{Defining Context Using Co-Occurrences and Extractibility}

For an object class $Y$, context cues often come in the form of another object class $C$ that has two properties (cf. \citep{lovering2021predicting}).
First, $C$ and $Y$ co-occur frequently \citep{biederman1982scene}.
Second, $C$ is  more \textit{extractible} than $Y$ --- easier to detect.
If $C$ were \textit{less} extractible than $Y$ it would not be a useful cue for detecting $Y$, as a model could detect $Y$ directly.

We can utilize these properties to create candidate context cues $C$ for a class of interest $Y$.
Given segmentations, we can use an object's size within an image as a proxy for extractibility (larger objects tend to be more extractible).
Let the $Area$ operator take the sum of the areas of all segmentations of instances of that object, or return 0 if the object is not present.
Then, to estimate from the training set how important of a context variable $C$ is, we can compute $A(C, Y) = \mathds{E}[Area(C) - Area(Y) | Y = 1]$.
When $A(C, Y)$ is larger, this means that when $Y$ is present, $C$ is usually also present, and on average, takes up more of the image than $Y$ does.
When $A(C, Y) > \alpha$, we say that $C$ is an $\alpha$-strong context cue for $Y$ (or just $\alpha$-context for brevity).
We find that many of the contexts identified using this method for large enough $\alpha$ are intuitive.
Some examples of (label, 0.05-context) pairs are: (\texttt{car}, \texttt{road}), (\texttt{bowl}, \texttt{dining\_table}), (\texttt{cow}, \texttt{grass}).

Using this notion of context, we can then define hard positive and hard negative examples.
We make the simplifying \textit{noisy-or} assumption: that each context cue provides evidence for $Y$, so that the presence of any cue supports $Y$ being present, while the absence of all provides evidence against $Y$'s presence.
Given some image, we can define an ($\alpha, \beta$)-hard positive or negative.
If $Y=1$, and for all $\alpha$-context cues $C$, we have $Area(C) < \beta$ in this image (and there is at least one $\alpha$-context cue), then the example is an ($\alpha, \beta$)-hard positive.
Alternatively, if $Y=0$, and there exists some $\alpha$-context variable $C$ such that $Area(C) > \beta$, then the example is an ($\alpha, \beta$)-hard negative.
We will call this method the co-occurrence/extractibility (CE) criterion (see Fig \ref{fig:benchmark-hard-pos-neg-examples} for examples).
Throughout, we use $\alpha=0.05, \beta=1$ unless otherwise noted; these parameters were chosen since they approximately equalize $P[(X, Y)\ \textrm{is}\ (\alpha, \beta)\textrm{-hard} \ | Y = y]$ across $y = 0, 1$.
See Appendix \ref{app:data} for more details.

\subsubsection{Defining Context Using Gist}

We now turn to a broader notion of context, that of the ``gist'' of a scene \citep{torralba2003contextual}, or its overall semantic content.
This is something that humans can recognize easily \citep{larson2014spatiotemporal}, but goes beyond object frequency and extractibility.
When an object is present in a scene whose gist is very different from the scenes the object was present in at training time, this may make prediction difficult. 
% create a large challenge for OOC prediction.

We describe our method for estimating gist shift, which we call the gist criterion.
We use caption annotation data in COCO (each image has 5 caption sentences), making the assumption that information in a caption captures the gist of a scene.
We then take an SBERT embedding \citep{reimers2019sentence} of each image caption for an image, and average these embeddings to get a single embedding for that image.
Then, for a given image and some target label $Y$, we find at the cosine similarity between that image's embedding and the average embedding across all training images with $Y=1$.
If this similarity is below some threshold $\tau$, and $Y = 1$ for the test image, it is a $\tau$-hard positive;
if this similarity is above $\tau$, and $Y = 0$ for the test image, it is a $\tau$-hard negative.
Note, we do not look at distance to $Y=0$ examples; we assume that captions for $Y=0$ images may have little in common, whereas the mean caption for $Y=1$ is a prototypical description of $Y$ ``in context''.
Throughout, we set the threshold $\tau$ for each task so that the number of hard positives and negatives is the same as the CE criterion for that task, to facilitate comparisons.
See Figure \ref{fig:results-qual-two-criteria} for examples of hard positive and negatives chosen by this criterion in comparison to the CE criterion, and Appendix \ref{app:data} for more details.

\subsection{\rococo-CE and \rococo-Gist: Selecting Challenge Sets }

\begin{figure}[t!] % "[t!]" placement specifier just for this example
\begin{subfigure}{0.25\textwidth}
\includegraphics[width=\linewidth]{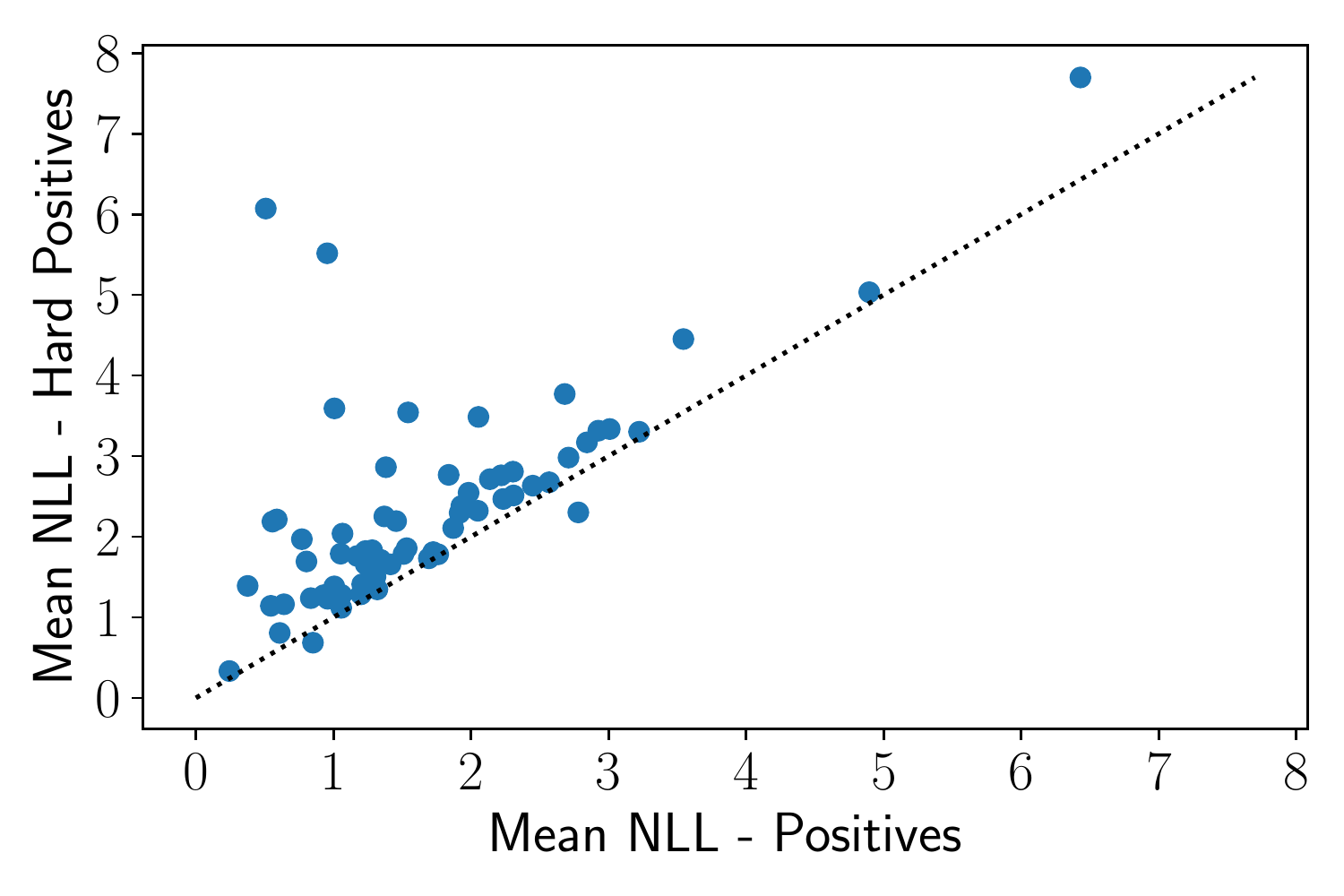}
% \caption{First subfigure} \label{fig:a}
\end{subfigure}\hspace*{\fill}
\begin{subfigure}{0.25\textwidth}
\includegraphics[width=\linewidth]{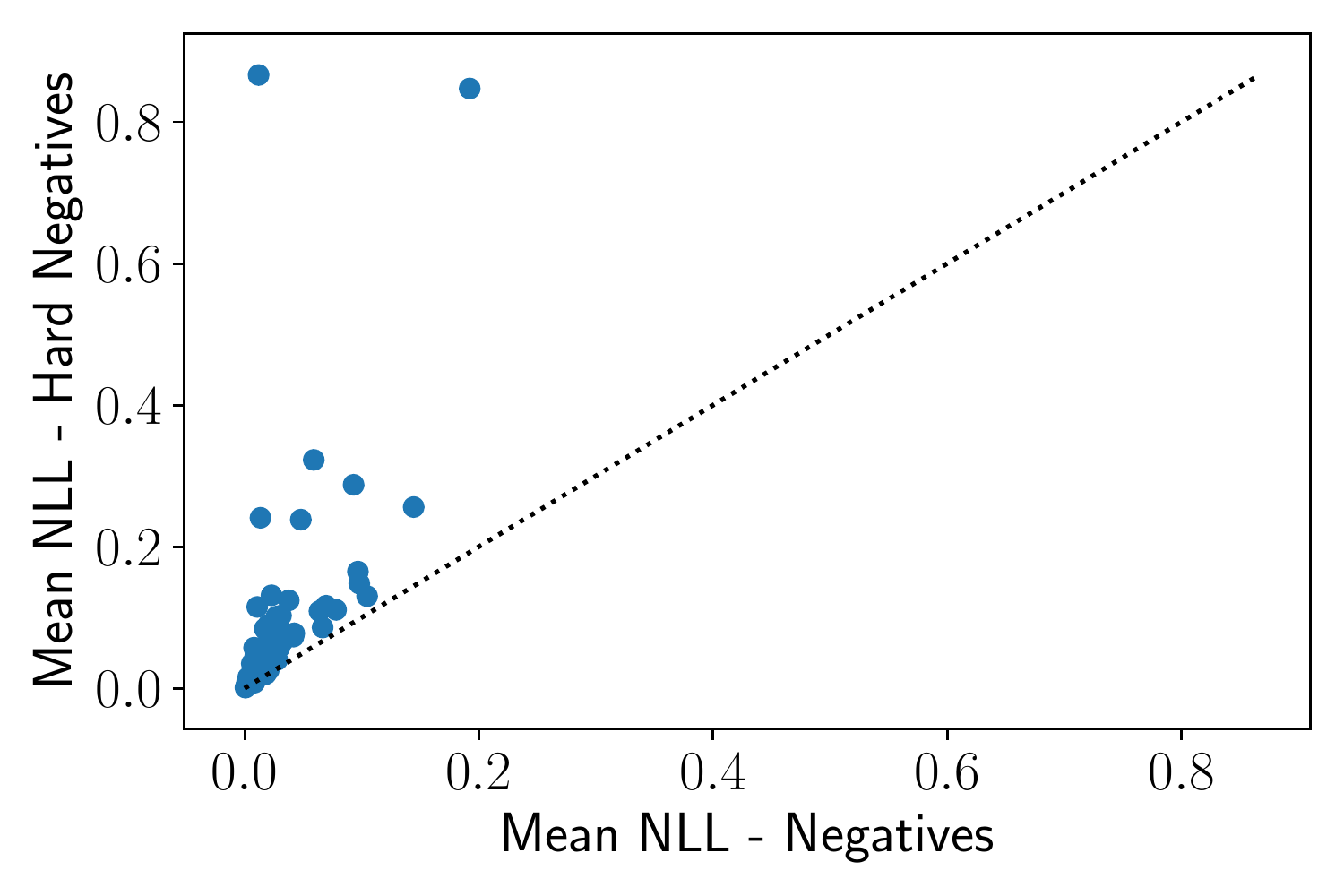}
% \caption{Second subfigure} \label{fig:b}
\end{subfigure}\hspace*{\fill}
\begin{subfigure}{0.25\textwidth}
\includegraphics[width=\linewidth]{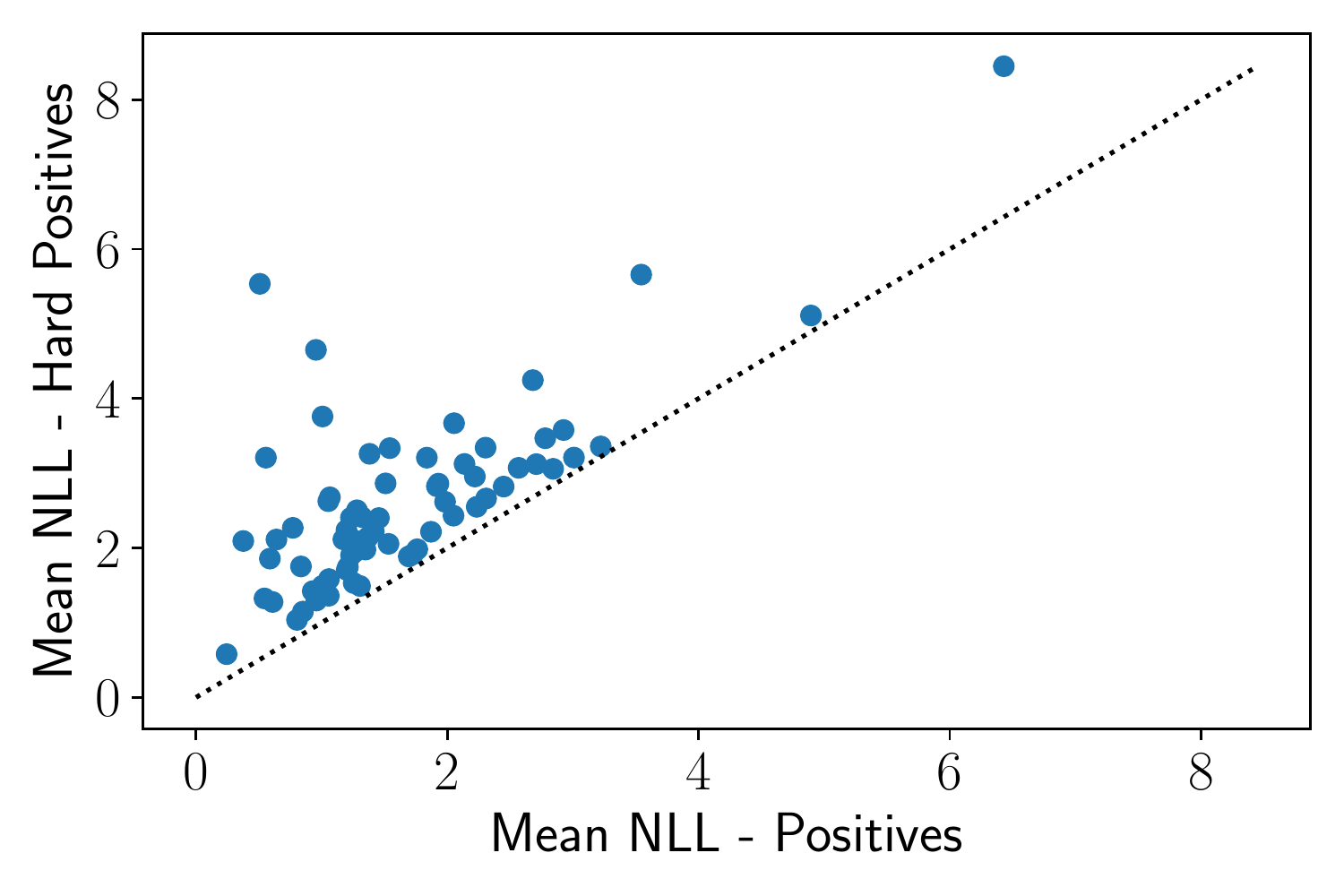}
% \caption{First subfigure} \label{fig:a}
\end{subfigure}\hspace*{\fill}
\begin{subfigure}{0.25\textwidth}
\includegraphics[width=\linewidth]{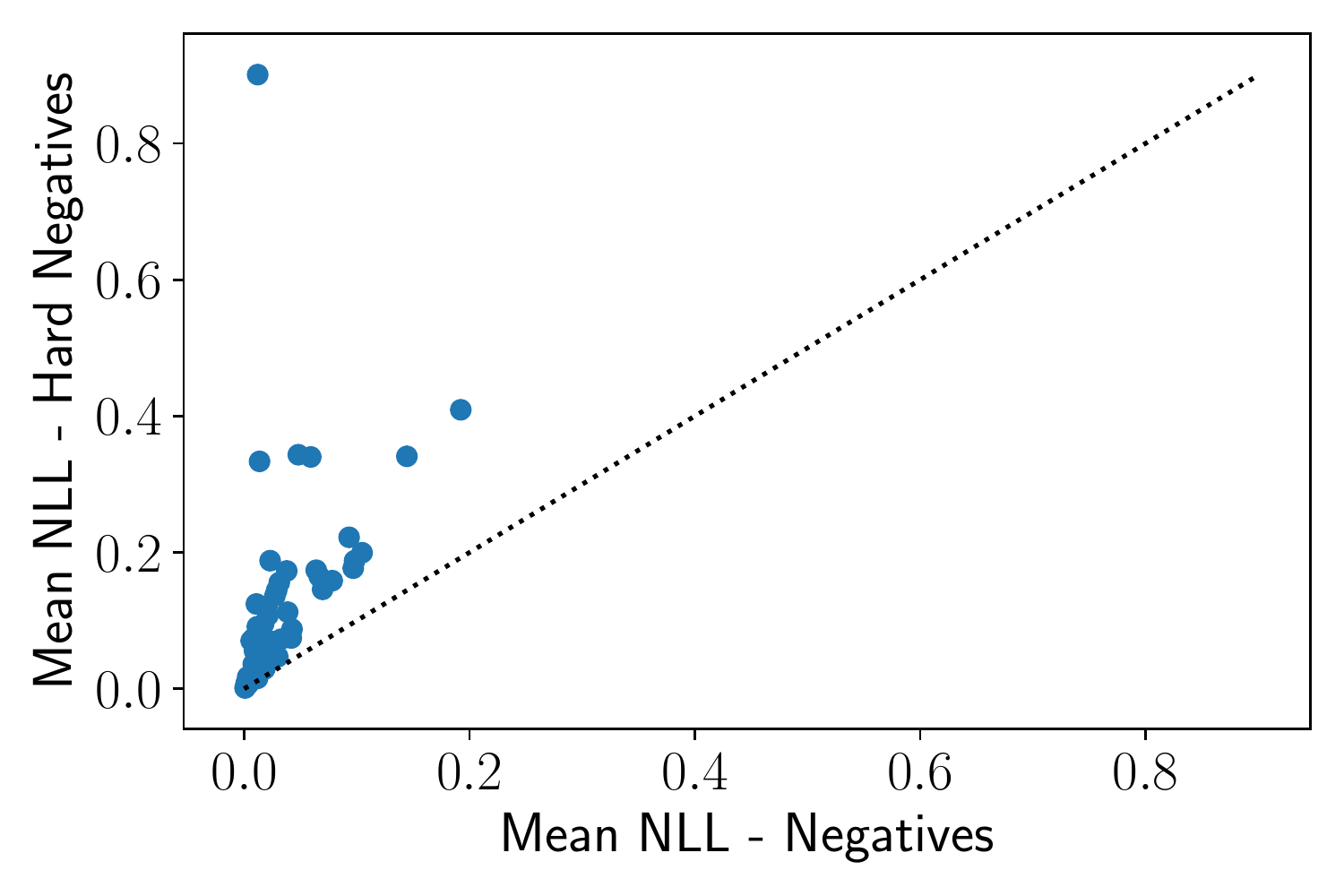}
% \caption{Second subfigure} \label{fig:b}
\end{subfigure}
\caption{
We find that hard positives/negatives induce higher average loss for both criteria.
Each point is a task in COCO-Stuff: the x- and y-axis values show the average loss achieved by an ERM model on all positives (negatives) and the average loss on hard positives (negatives).
From L to R: CE criterion (positives), CE criterion (negatives), gist criterion (positives), gist criterion (negatives).
The diagonal line represents where the hard example losses match marginal losses.
} \label{fig:benchmark-hard-pos-neg-task-scatter}
\end{figure}

We train binary classifiers to minimize average NLL on each of the 171 classes in COCO-Stuff.
We find that for nearly all tasks, the hard positives and negatives defined by our methods incur higher average loss than positive and negative examples respectively (Fig. \ref{fig:benchmark-hard-pos-neg-task-scatter}), for both the CE and Gist criteria.
This provides some evidence that our criteria are, in fact, identifying examples which are more difficult to classify correctly.

\begin{wraptable}{R}{0.45\linewidth}
    \centering
    \begin{tabular}{lrrr}
\hline
 Task         &   \# Hard + &   \# Hard - &   \# + \\
\hline
 car          &       1539 &        949 &  2585 \\
 bowl         &        944 &       1084 &  1463 \\
 boat         &        361 &        756 &   649 \\
 fire-hydrant &        189 &       1577 &   332 \\
 airplane     &        258 &       3622 &   635 \\
 cow          &        194 &       2360 &   445 \\
 backpack     &        607 &       6413 &  1183 \\
 cup          &        743 &       6852 &  1926 \\
 surfboard    &        147 &       4011 &   750 \\
 tie          &        136 &       6347 &   778 \\
 sports-ball  &        151 &       6574 &   860 \\
 kite         &         55 &       6725 &   451 \\
\hline
\end{tabular}

    \caption{Counts of hard positive, hard negative \& positive examples in the test set per \rococo\ task.}
    \label{tab:n_hard_examples}
\end{wraptable}

To select candidate OOC tasks for our challenge sets, we select the 12 tasks with the largest difference between average NLL on hard examples (by the CE criterion) and average NLL on all examples, filtering out those tasks without at least 50 hard positives and hard negatives.
We call these tasks the {\rococo \ } (Naturally-Occurring Out-of-context Challenge) suite.
% (Table \ref{tab:n_hard_examples}). 
We then identify two groups of challenge sets: \rococo-CE, which consists of the hard positive and negative examples on each of the 12 tasks in {\rococo \ } as identified by the CE criterion; and \rococo-Gist, which is the analogous set for the gist criterion; see Table \ref{tab:n_hard_examples} for a list of tasks.
This gives us 24 total challenge sets on which to evaluate an ML model's OOC performance.

\paragraph{Contrasting the Two Criteria (Fig. \ref{fig:results-qual-two-criteria}).}
The left two images in Fig. \ref{fig:results-qual-two-criteria} are hard positives on the \texttt{airplane} task.
On the far left, we see the inside of an airplane: this is selected as a hard positive by the CE criteria because there is no sky visible.
On the second left, the sky is visible but the overall scene is unusual; this is a hard positive by the Gist criteria.
The right two images are hard negatives on the \texttt{bowl} task.
On the second right, we see a large dinner with many plates: this is a hard negative by the CE criteria because there is a prominent dining table but no bowls.
On the far right, we see a paper plate on a desk (not a dining table): 
this is a hard negative by the Gist criteria since there is no bowl but it is similar to images where you might expect a bowl.

\section{Evaluating Robustness Approaches on {\rococo }} \label{sec:methods}
We now turn to evaluating various ML methods on the {\rococo \ } benchmarks.
We choose to focus on four categories of methods which provide a useful contrast between different approaches to OOC prediction. 
Expected risk minimization (\textbf{ERM}) is the standard loss function in ML systems; as shown by \citet{gulrajani2020search}, it can be difficult to beat on domain generalization problems.
Label-based adjustments (\textbf{Reweight} and \textbf{US}) use information about label imbalance at training time to adjust the loss function towards performing better on unusual labels.
Environment-based methods (\textbf{GDRO} \citep{sagawa2019distributionally}, \textbf{IRM} \citep{arjovsky2019invariant}, \textbf{Reweight-Envs}, \textbf{US-Envs}) use categorical auxiliary information (i.e. \textit{environments}) at training time to adjust the loss function towards performing better on unusual subgroups.
Finally, adaptive methods (\textbf{CVaR} (\citep{rockafellar2002conditional}) and \textbf{Focal} \citep{lin2017focal}) do not use auxiliary information at all, but instead try to infer adaptively from the input during training what the worst-performing subgroups should be.
See App. \ref{app:other-methods} for information on other methods we experimented with but did not include here.
We found the methods we included here had the best results overall, on the standard and hard examples.

Throughout, we use the following notation.
We are given a dataset $\{x_i, y_i\}_{i=1}^n$ with inputs and target labels respectively, and possibly some side information $\{ c_i \}_{i=1}^n$ as well (e.g. environment variables).
If side information is available, we assume it is available at training time but not necessarily test time.
We are aiming to learn a function $f: \mathds{X} \longrightarrow \mathds{Y}$.
We assume $\ell$ to be the the example-wise cross-entropy loss function.

\subsection{Expected Risk Minimization}
Expected risk minimization (ERM) is the standard paradigm for training ML models.
In ERM, we minimize the mean loss $\ell$ on the training set, plus potentially some data-independent regularizer $\mathcal{R}$:
\begin{equation}
    \mathcal{L}_{ERM}(f) = \frac{1}{n} \sum_{i=1}^n \ell(f(x_i), y_i) + \mathcal{R}(f)
\end{equation}

\subsection{Label-Based Adjustments}

Since COCO-Stuff contains significant label imbalance, we consider methods intended to correct for label imbalance.
We let $w_i = P(Y = y_i)$ in the training set.
To make a fairer comparison with the environment-based methods (below), we add a tuning parameter $\alpha$ to control the degree of adjustment.
$\alpha = 1$ represents the standard versions of these loss functions, but we frequently found other values $\alpha \in [0, 2]$ to be useful.

\paragraph{Label Reweighting.}
We reweight the loss of each training example in the ERM loss function, inversely proportional (when $\alpha = 1$) to the likelihood of observing that label:
\begin{equation}
    \mathcal{L}_{Reweight}(f) = \frac{1}{n} \sum_{i=1}^n (\frac{1}{w_i})^\alpha \ell(f(x_i), y_i) + \mathcal{R}(f)
\end{equation}

\paragraph{Label Undersampling.}
Rather than sampling training examples uniformly, we sample them with probability inversely proportional (when $\alpha = 1$) to the likelihood of observing that label.
We let $\mathcal{I}$ be a list of indices of size $n$, sampled independently from $[1 \dots n]$, such that for each $j \in \mathcal{I}$, $P(j = i)$ is proportional to $(\frac{1}{w_i})^\alpha$.
We redraw this for the training set every epoch and weight their losses equally as follows:
\begin{equation}
    \mathcal{L}_{Undersampling}(f) = \frac{1}{n} \sum_{i \in \mathcal{I}} \ell(f(x_i), y_i) + \mathcal{R}(f)
\end{equation}

\subsection{Environment-Based Methods}
One common setting is where the auxiliary information $c_i$ for each example is a categorical variable, representing an \textit{environment}.
The notion of environments for robust learning was originally given with causal motivation \citep{peters2016causal}, where each value of $c$ represents a different intervention on the underlying causal graph.
However, this type of grouping procedure can be used completely separately from the causal context, and so we just consider it to represent some informative partition of our data, which is defined as an input at training time.
Here, $c \in \{ 1 \dots C \}$ is an integer and
we use the following shorthand to describe the average loss for an environment, with $n_c$ referring to the number of examples in that environment: 
\begin{equation}
    \ell_c (f) = \frac{1}{n_c} \sum_{i=1}^{n} \ell(f(x_i), y_i) \one \{ c_i = c \}
\end{equation}
% We consider two popular environment-based learning algorithms.

\paragraph{Group DRO.}
The first environment-based method we consider is Group Distributionally Robust Optimization (GDRO) \citep{sagawa2019distributionally}, which aims to minimize the loss on the worst of $C$ partitions, with partition $c$ of size $n_c$, and group adjustment hyperparameter $K$:
\begin{equation}
    \mathcal{L}_{GDRO}(f) = \underset{c \in \{ 1 \dots C \}}{\max} \Big\{  \ell_c (f) + \frac{K}{\sqrt{n_c}}  \Big\} + \mathcal{R}(f)
\end{equation}
This loss aims to ensure that no group's loss is that bad, and the group adjustment term ensures greater focus on smaller groups, which may otherwise be ignored.

\paragraph{IRM.}
The second environment-based method we consider is invariant risk minimization (IRM) \citep{arjovsky2019invariant}:
% , which uses the following loss function
\begin{equation}
    \mathcal{L}_{IRM}(f) = \sum_{c=1}^C  \ell_c (f) + \lambda \| \nabla_{w | w=1} \ell_c (w \cdot f) \| + \mathcal{R}(f)
\end{equation}
The second term here is a gradient penalty on the output of $f$, $w$ is a constant multiplier on the output of $f$, and $\lambda$ is a hyperparameter.
The intuition for this is somewhat involved \citep{arjovsky2019invariant}; the overarching motivation is that each environment should learn a representation such that the same predictive classifier is optimal across environments.

\paragraph{Environment Reweighting and Undersampling.}
These are equivalent to label reweighting/undersampling above, but with $w_i = P(C = c_i)$.

\subsection{Adaptive Methods}

Finally, we consider a class of loss functions which, rather than using an injection of side information to specify which examples are OOC, focuses dynamically on the hardest examples at each step.
The first is conditional variance-at-risk optimization (CVaR).
This can be cast as a type of distributionally robust optimization (DRO) \citep{duchi2021statistics}, where we aim to minimize the loss over, rather than the whole training set, a worst-case distribution over training examples.
For some $p \in (0, 1)$, CVaR($p$) is defined as the average loss of the $p$-percent worst-loss examples.

We also consider focal loss \citep{lin2017focal}, which dynamically upweights high loss examples using a parameter $\gamma \geq 0$.
With binary $y$, $q(x, y) = f(x)$ if $y = 1$ and $q(x) = 1 - f(x)$ if $y = 0$, with $f(x) \in [0, 1]$ as the continuous-scored output of $f$, not the binary prediction.
Then, we have focal loss as a modification of the cross-entropy loss --- at $\gamma=0$ this reduces to cross-entropy; as $\gamma$ increases, it focuses more loss on the examples which have higher loss already:
\begin{equation}
    \mathcal{L}_{Focal}(f) = \frac{1}{n} \sum_{i=1}^n -(1 - q(x_i, y))^\gamma \log(q(x_i, y)) + \mathcal{R}(f)
\end{equation}

\section{Related Work}

A range of datasets have been proposed for the purposes of benchmarking OOC prediction.
Several focus on realistic OOC prediction: WILDS \citep{koh2020wilds} contains problems from across a range of applications; Imagenet-A \citep{hendrycks2021natural} select examples which perform worst on an ensemble of models; ObjectNet \citep{barbu2019objectnet} focuses on object recognition and shows objects varied by a range of attributes;  \citet{choi2011tree} isolate 26 images of objects in unusual contexts from a larger dataset.
Our work functions well as a complement to any of these datasets; we believe it is novel due to our ideas for scalably identifying challenge sets from annotated data, as well as its delineation of hard positives and hard negatives.
% novel due to its scale, its delineation of hard negatives and hard positives, and its focus on annotated data.
% its focus on object detection, where object relationships are paramount, its use of caption data, and its potential for interpretability due to a extensive annotations.
The notion of ``challenge sets'' \citep{isabelle2017challenge}, ``stress tests'' \citep{naik2018stress}, or ``contrast sets'' \citep{gardner2020evaluating} from the NLP literature is an inspiration for our work as well.
A range of primarily semi-synthetic datasets include those that center around: image corruption (Imagenet-C, Imagenet-P \citep{hendrycks2019benchmarking}; Pascal-C, Coco-C, Cityscapes-C \citep{michaelis2019benchmarking}); object hierarchy (BREEDS \citep{santurkar2020breeds}); synthetic shifts in background (Waterbirds \citep{sagawa2019distributionally}; Imagenet-9 \citep{xiao2020noise}); color (Colored MNIST \citep{kim2019learning,arjovsky2019invariant}); a group attribute (partitioned Civil Comments \citep{adragna2020fairness}); or purely synthetic data (Invariance Unit Tests \citep{aubin2021linear}).

Several works have discussed explicit examples where deep models failed to perform OOC prediction in practice. 
\citet{oakden2020hidden} and \citet{winkler2019association} discuss the risk of this occurring in the medical domain, and \citet{shetty2019not} in the autonomous driving domain.
Other works have detailed the challenge of OOC prediction for deep models, using frames of ``shortcuts'' \citep{geirhos2020shortcut}, ``simplicity'' \citep{shah2020pitfalls}, ``extractibility'' \citep{lovering2021predicting}, texture biases in CNNs \citep{geirhos2018imagenet}, or the challenge of out-of-place objects \citep{rosenfeld2018elephant}.

A range of work outside deep learning considers the OOC prediction problem from a different direction, focusing on how to improve prediction by taking context into account \citep{heitz2008learning,choi2012context,yao2010modeling,murphy2003using}.
Other work looks at the idea of using a latent variable to represent scene gist \citep{oliva2006building,wu2018learning}.
A number of newer methods not discussed elsewhere in the paper also aim to solve the OOC problem, including those that involve side information \citep{hu2018does,srivastava2020robustness,xie2020n,krueger2020out}, those that involve side information through causal underpinnings \citep{heinze2021conditional,rothenhausler2018anchor}, and those that ignore side information altogether \citep{creager2020environment,dagaev2021too,duchi2016variance}.

\section{Experiments} \label{sec:experiments}

In this section, we compare and contrast the various measurements of OOC performance yielded by {\rococo}, along with the semi-synthetic Waterbirds dataset \citep{sagawa2019distributionally}.
For all experiments we use a ResNet-50 \citep{he2016deep}, finetuned from ImageNet-pretrained features \citep{russakovsky2015imagenet}.
See App. \ref{app:data} and \ref{app:experiments} for further experimental details.
For the environment-based methods, we follow \citet{sagawa2019distributionally} and create 4 environments: 1 for each element of the cross-product of the label and its highest-$\alpha$ context class.
Many of the robust baselines from Sec. \ref{sec:methods} come with a hyperparameter which aims to trade off between average performance and OOC performance; we choose the hyperparameter which minimizes the maximum loss of hard positives and hard negatives on the validation set.

% TODO change tables
\begin{table}
\resizebox{\columnwidth}{!}{%
\centering

\begin{tabular}{lccccccccc}
\hline
 Task         & ERM            &   CVaR &   Focal & Reweight       & US             & Reweight (Envs)   & US (Envs)      & GDRO           & IRM            \\
\hline
 car          & 0.769          &  0.787 &   0.759 & 0.773          & 0.773          & \textbf{0.886}    & 0.846          & \textbf{0.891} & 0.868          \\
 bowl         & 0.749          &  0.781 &   0.734 & 0.751          & 0.759          & \textbf{0.864}    & 0.814          & \textbf{0.865} & 0.828          \\
 boat         & 0.869          &  0.888 &   0.823 & 0.866          & 0.877          & \textbf{0.954}    & 0.923          & \textbf{0.945} & 0.925          \\
 fire-hydrant & 0.913          &  0.933 &   0.908 & 0.927          & 0.913          & 0.933             & 0.920          & \textbf{0.946} & \textbf{0.942} \\
 airplane     & 0.986          &  0.984 &   0.983 & 0.986          & 0.985          & \textbf{0.991}    & 0.986          & \textbf{0.991} & 0.986          \\
 cow          & 0.935          &  0.937 &   0.932 & 0.939          & 0.938          & \textbf{0.963}    & 0.948          & \textbf{0.963} & 0.943          \\
 backpack     & 0.812          &  0.806 &   0.809 & 0.816          & 0.812          & 0.813             & 0.816          & \textbf{0.871} & 0.844          \\
 cup          & \textbf{0.870} &  0.863 &   0.867 & \textbf{0.876} & \textbf{0.873} & \textbf{0.879}    & \textbf{0.875} & 0.825          & \textbf{0.869} \\
 surfboard    & 0.939          &  0.947 &   0.933 & 0.940          & 0.939          & \textbf{0.960}    & 0.940          & \textbf{0.960} & \textbf{0.957} \\
 tie          & 0.742          &  0.752 &   0.728 & 0.760          & 0.763          & 0.756             & 0.761          & \textbf{0.806} & \textbf{0.775} \\
 sports-ball  & 0.867          &  0.869 &   0.871 & 0.869          & 0.868          & \textbf{0.911}    & 0.890          & \textbf{0.911} & 0.894          \\
 kite         & 0.932          &  0.932 &   0.937 & 0.940          & 0.928          & 0.949             & 0.936          & \textbf{0.960} & \textbf{0.950} \\
\hline
 Average      & 0.865          &  0.873 &   0.857 & 0.870          & 0.869          & 0.905             & 0.888          & \textbf{0.911} & 0.899          \\
\hline
\end{tabular}
% \caption{Foo}
}
\caption{AUC on hard test examples for all 12 \rococo-CE \ stress tests, after hyperparameter selection. 
Bold numbers have overlapping standard deviations with the highest observed mean's.
}
\label{tab:auc-by-task-ce}
% \end{table}
% \begin{table}
\resizebox{\columnwidth}{!}{%
\centering

\begin{tabular}{lccccccccc}
\hline
 Task         & ERM            & CVaR           & Focal          & Reweight       & US             & Reweight (Envs)   & US (Envs)      & GDRO           & IRM            \\
\hline
 car          & 0.766          & 0.785          & 0.763          & 0.768          & 0.773          & 0.845             & 0.823          & \textbf{0.863} & 0.832          \\
 bowl         & 0.642          & \textbf{0.686} & 0.635          & 0.656          & 0.678          & \textbf{0.692}    & \textbf{0.698} & \textbf{0.704} & \textbf{0.670} \\
 boat         & 0.826          & 0.855          & 0.771          & 0.823          & 0.823          & \textbf{0.897}    & 0.868          & 0.884          & 0.872          \\
 fire-hydrant & 0.840          & \textbf{0.865} & 0.822          & 0.849          & 0.841          & 0.834             & 0.836          & 0.823          & 0.845          \\
 airplane     & \textbf{0.977} & \textbf{0.979} & 0.975          & \textbf{0.978} & 0.976          & \textbf{0.978}    & 0.977          & \textbf{0.981} & \textbf{0.977} \\
 cow          & 0.901          & \textbf{0.912} & 0.877          & 0.903          & 0.906          & 0.908             & \textbf{0.908} & \textbf{0.911} & 0.896          \\
 backpack     & 0.716          & 0.717          & 0.725          & 0.731          & \textbf{0.749} & 0.729             & 0.736          & \textbf{0.732} & \textbf{0.750} \\
 cup          & 0.733          & 0.738          & 0.727          & 0.735          & 0.742          & 0.759             & \textbf{0.771} & 0.742          & \textbf{0.755} \\
 surfboard    & 0.913          & \textbf{0.920} & 0.900          & 0.912          & 0.914          & 0.909             & 0.912          & 0.882          & 0.894          \\
 tie          & \textbf{0.822} & 0.816          & 0.813          & \textbf{0.829} & \textbf{0.824} & \textbf{0.829}    & \textbf{0.831} & \textbf{0.835} & \textbf{0.840} \\
 sports-ball  & 0.830          & 0.832          & 0.828          & 0.830          & 0.822          & 0.866             & 0.859          & \textbf{0.880} & 0.855          \\
 kite         & \textbf{0.942} & \textbf{0.939} & \textbf{0.945} & \textbf{0.947} & 0.936          & \textbf{0.947}    & \textbf{0.939} & \textbf{0.947} & \textbf{0.943} \\
\hline
 Average      & 0.826          & 0.837          & 0.815          & 0.830          & 0.832          & \textbf{0.849}    & 0.847          & \textbf{0.849} & 0.844          \\
\hline
\end{tabular}
% \caption{Bar}
}
\caption{AUC on hard test examples for all 12 \rococo-Gist \ stress tests, after hyperparameter selection. 
Bold numbers have overlapping standard deviations with the highest observed mean's.
}
\label{tab:auc-by-task-gist}
% \end{table}

% \begin{table}[]
    \centering
    \resizebox{\columnwidth}{!}{%
    \begin{tabular}{lrrrrrrrll}
\hline
 Metric            &   ERM &   CVaR &   Focal &   Reweight &    US &   Reweight (Envs) &   US (Envs) & GDRO           & IRM            \\
\hline
 Worst-Group Error & 0.4   &  0.354 &   0.402 &      0.361 & 0.336 &             0.233 &       0.146 & \textbf{0.108} & 0.127          \\
 Worst-Group NLL   & 1.03  &  0.657 &   0.722 &      0.884 & 0.883 &             0.535 &       0.417 & \textbf{0.306} & \textbf{0.346} \\
 AUC-Hard          & 0.691 &  0.703 &   0.506 &      0.657 & 0.701 &             0.744 &       0.778 & \textbf{0.929} & 0.788          \\
\hline
\end{tabular}
    }
    \caption{AUC on ``hard'' test examples on the Waterbirds dataset.
    Hard examples are the union of two groups: land birds on water backgrounds and water birds on land backgrounds.
    Worst-group error or NLL takes the worse of the two; AUC-Hard calculates the AUC across the union.
    }
    \label{tab:waterbirds}
\end{table}

\subsection{Quantitative Analysis}

\subsubsection{Main Results: Different Criteria Yield Different Evaluations}
In Tables \ref{tab:auc-by-task-ce} and \ref{tab:auc-by-task-gist}, we show AUC (area under the ROC curve) on hard examples for each method we focus on and each of the 12 {\rococo\ } tasks individually.
We choose AUC as a metric since it is robust to label imbalance, and tasks in {\rococo \ } contain 1-10\% positive examples.

For comparison, we show in Table \ref{tab:waterbirds} an analogous table of results for Waterbirds \citep{sagawa2019distributionally}, a semi-synthetic dataset which is generated by pasting images of birds on top of either either land or water backgrounds; the goal is to classify land birds from water birds.
At training time, land birds are mostly shown on land (and water birds on water), but at test time we hope to perform well, regardless of background.
The information regarding place is made available through environments: the four environments are land/land, land/water, etc.
For Waterbirds, we show both AUC-Hard (our metric) and worst-group erorr/NLL (the metrics from \citet{sagawa2019distributionally}).

\paragraph{Environments are More Useful on Simpler Benchmarks.}
We first note that the three different benchmarks presented yield varying conclusions about the methods in question; in particular, the relative performance of the best environment-based methods vary greatly between the benchmarks.
We find that environment-based methods perform better on the benchmarks where the structure of the shift (i.e. the form of $\phi$) is better specified by the environments.
In Table \ref{tab:waterbirds}, we see that there is a particularly large gap in performance between the best environment-based methods (particularly GDRO) and the methods that do not use environments.
For instance, GDRO and IRM improve over ERM by about 0.3 in worst-group error (the metric of choice in \citet{sagawa2019distributionally}); and GDRO improves over \textit{all} other methods by about 0.14 in AUC on hard examples.
This difference is an order of magnitude greater than observed on either {\rococo } benchmark, suggesting that semi-synthetic datasets (such as Waterbirds), may overestimate the performance of current environment-based methods, and possibly GDRO in particular.

We see a smaller version of this phenomenon when just comparing \rococo-CE to \rococo-Gist: the environment-based methods are also the ones whose performance falls off the most between \rococo-CE and \rococo-Gist.
This could be because \rococo-CE is a simpler benchmark, more well specified by the given environments.
The contrast between Tables \ref{tab:auc-by-task-ce}, \ref{tab:auc-by-task-gist} and Table \ref{tab:waterbirds} documents the usefulness of having benchmarks for robustness across a range of complexity, and motivates the creation of benchmarks such as \rococo.

\paragraph{Gist Shift is Difficult.}
We further note that performance on \rococo-Gist is worse than on \rococo-CE.
This suggests that the more complex notion of context embodied by the gist yields a more difficult OOC task.
In some ways, would expect models to find these more holistic shifts harder, as these gist shifts go well beyond ``shortcuts'' \citep{geirhos2020shortcut}.
We also note that the gap between the environment-based methods and ERM is much smaller on \rococo-Gist, indicating that ERM is a strong baseline on this task, making gist shift a useful direction for OOC prediction research.

\paragraph{Access to Auxiliary Information is More Important than Algorithm.}
Overall, environment-based methods perform the best on all three benchmarks --- we expect this to be the case, since these methods are given access to structure which is relevant to the OOC task at hand.
This is mostly clearly indicated in the improvement between reweighting/undersampling methods when using environments rather than labels.
Interestingly, we find that on the more complex {\rococo } benchmarks, reweighting examples by environment performs similarly to  more specialized methods such as GDRO/IRM, given an equivalent amount of hyperparameter tuning.
This is true to a lesser degree for undersampling by environment as well.
This suggests that current environment-based methods have significant room to improve when it comes to more complex OOC benchmarks.

% The effect of environments and how to choose or create them is an open question,
% and an active area of research.

\subsubsection{Secondary Observations}

\begin{figure}[t!] % "[t!]" placement specifier just for this example
\begin{subfigure}{0.48\textwidth}
\centering
\includegraphics[width=0.9\linewidth]{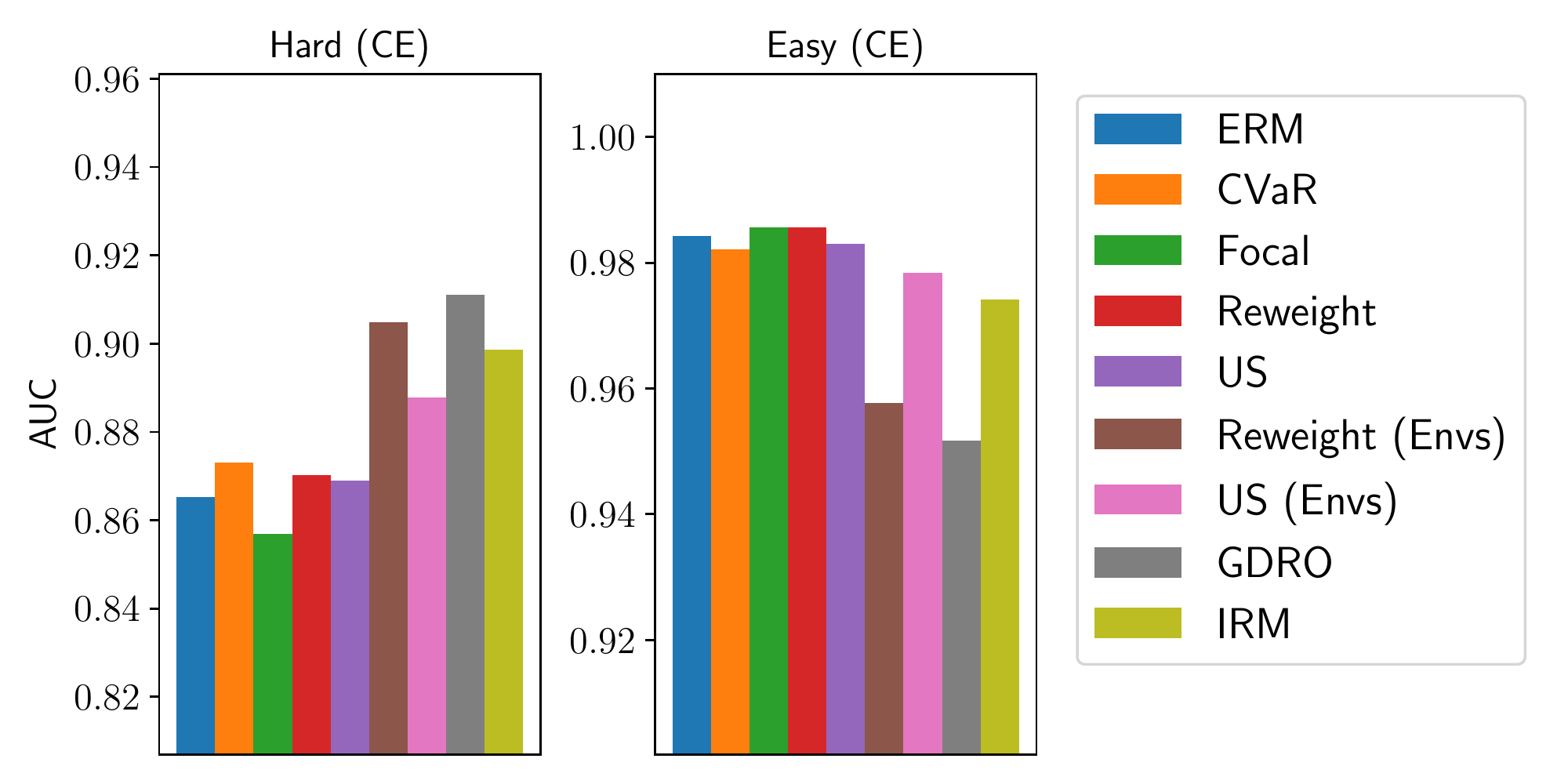}
% \caption{First subfigure} \label{fig:a}
\end{subfigure}\hspace*{\fill}
\begin{subfigure}{0.48\textwidth}
\centering
\includegraphics[width=0.9\linewidth]{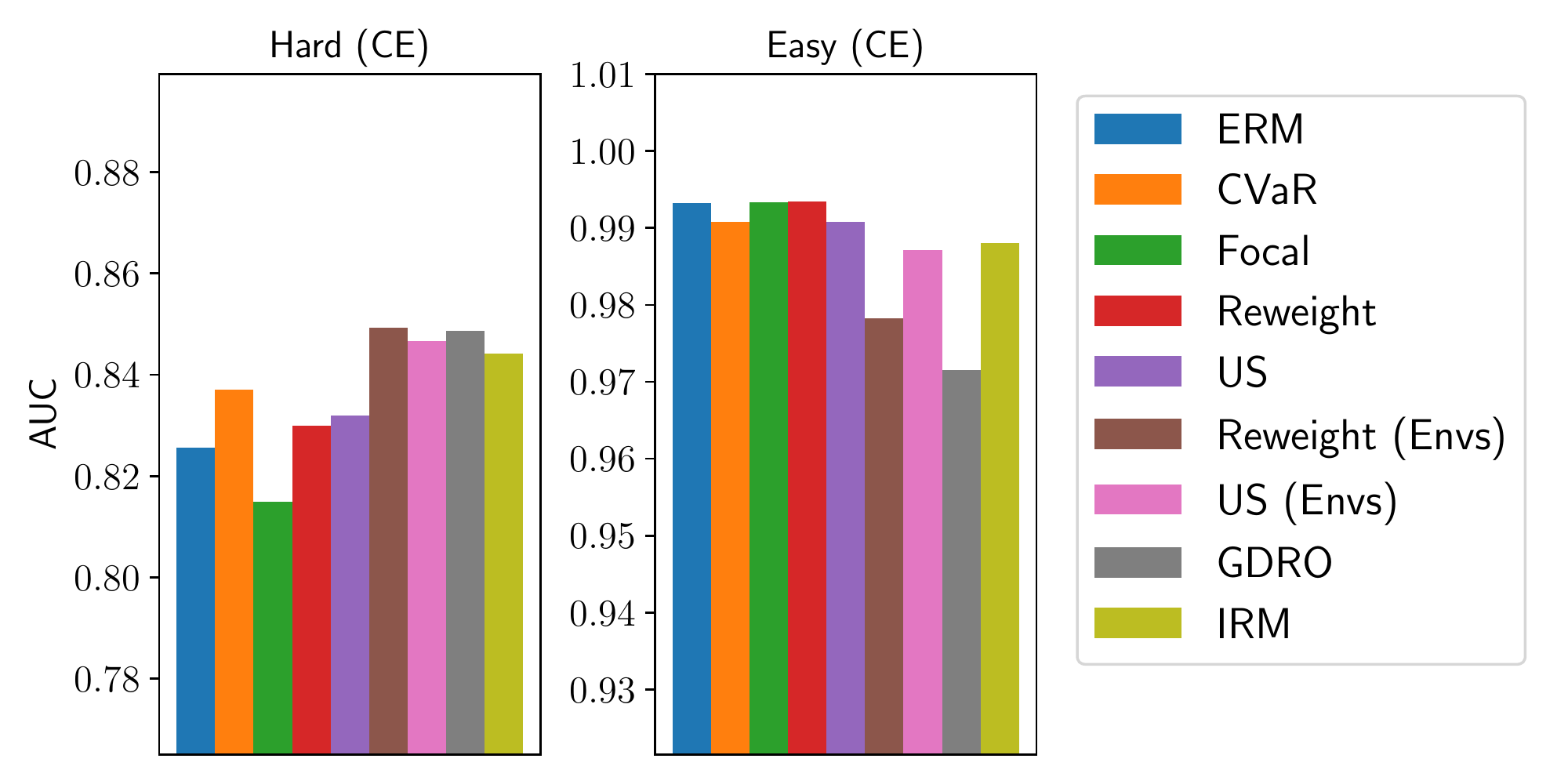}
% \caption{Second subfigure} \label{fig:b}
\end{subfigure}
\caption{
AUC (area under the ROC curve) achieved on hard and easy examples. Higher is better.
% L: \rococo-CE.
% R: \rococo-Gist.
} \label{fig:results-auc-hard-easy}
\end{figure}

\begin{figure}[t!] % "[t!]" placement specifier just for this example
\begin{subfigure}{0.48\textwidth}
\centering
\includegraphics[width=0.9\linewidth]{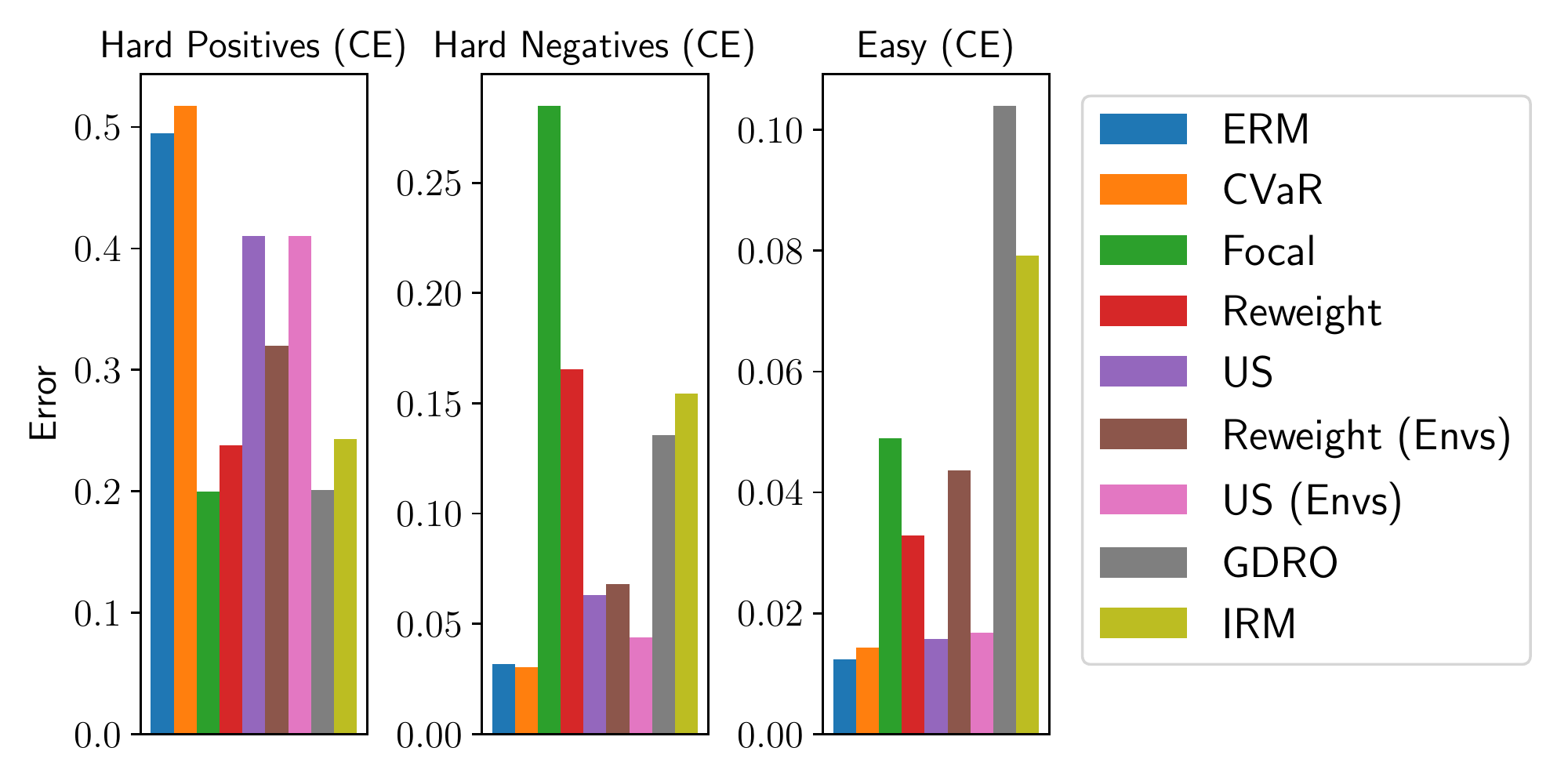}
% \caption{First subfigure} \label{fig:a}
\end{subfigure}\hspace*{\fill}
\begin{subfigure}{0.48\textwidth}
\centering
\includegraphics[width=0.9\linewidth]{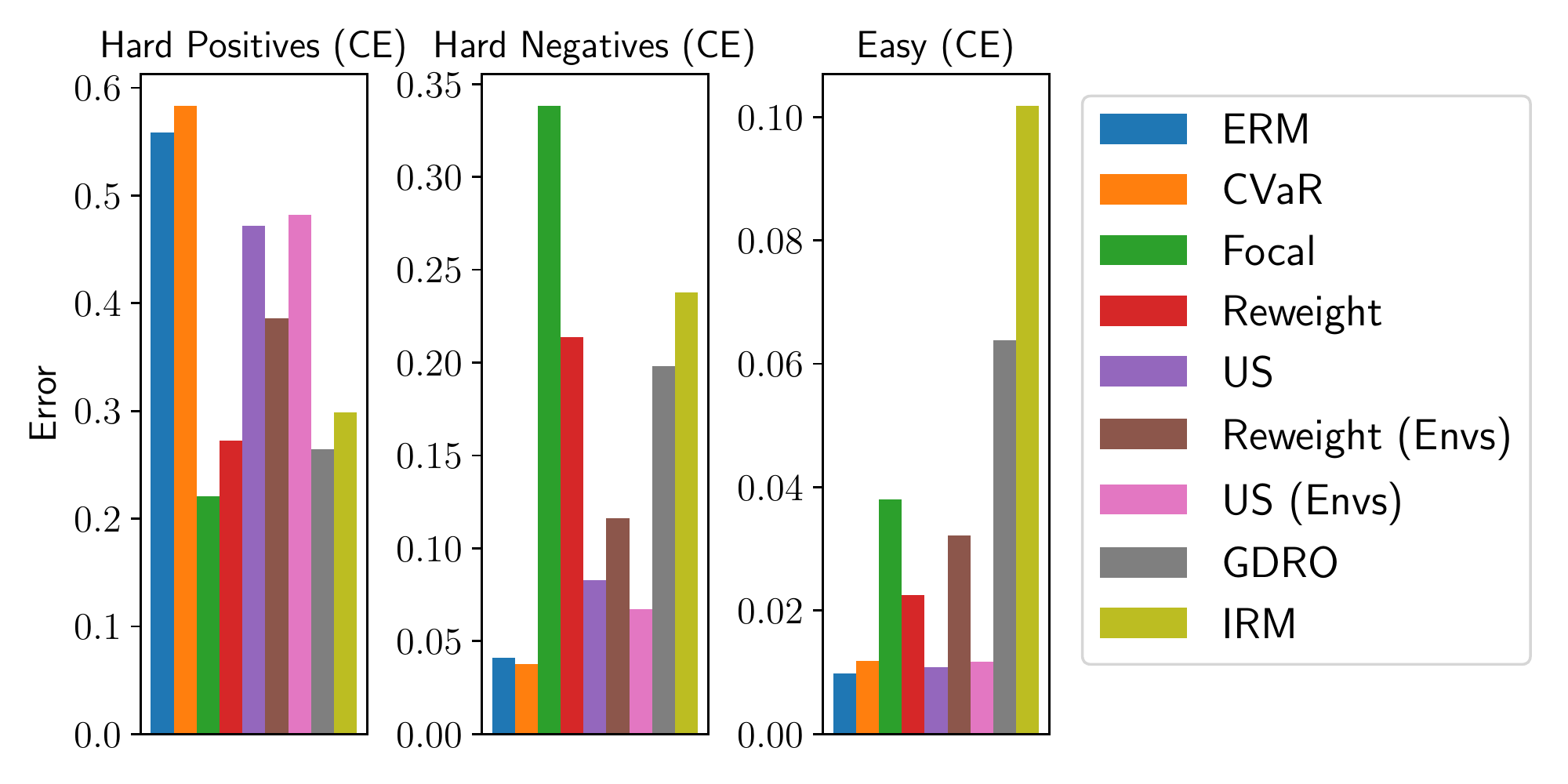}
% \caption{Second subfigure} \label{fig:b}
\end{subfigure}
\caption{
Classification error achieved on hard positive, hard negative, and easy examples.
% L: \rococo-CE.
% R: \rococo-Gist.
} \label{fig:results-err-hard-posneg-easy}
\end{figure}

\begin{figure}[t!] % "[t!]" placement specifier just for this example
\begin{subfigure}{0.48\textwidth}
\centering
\includegraphics[width=0.9\linewidth]{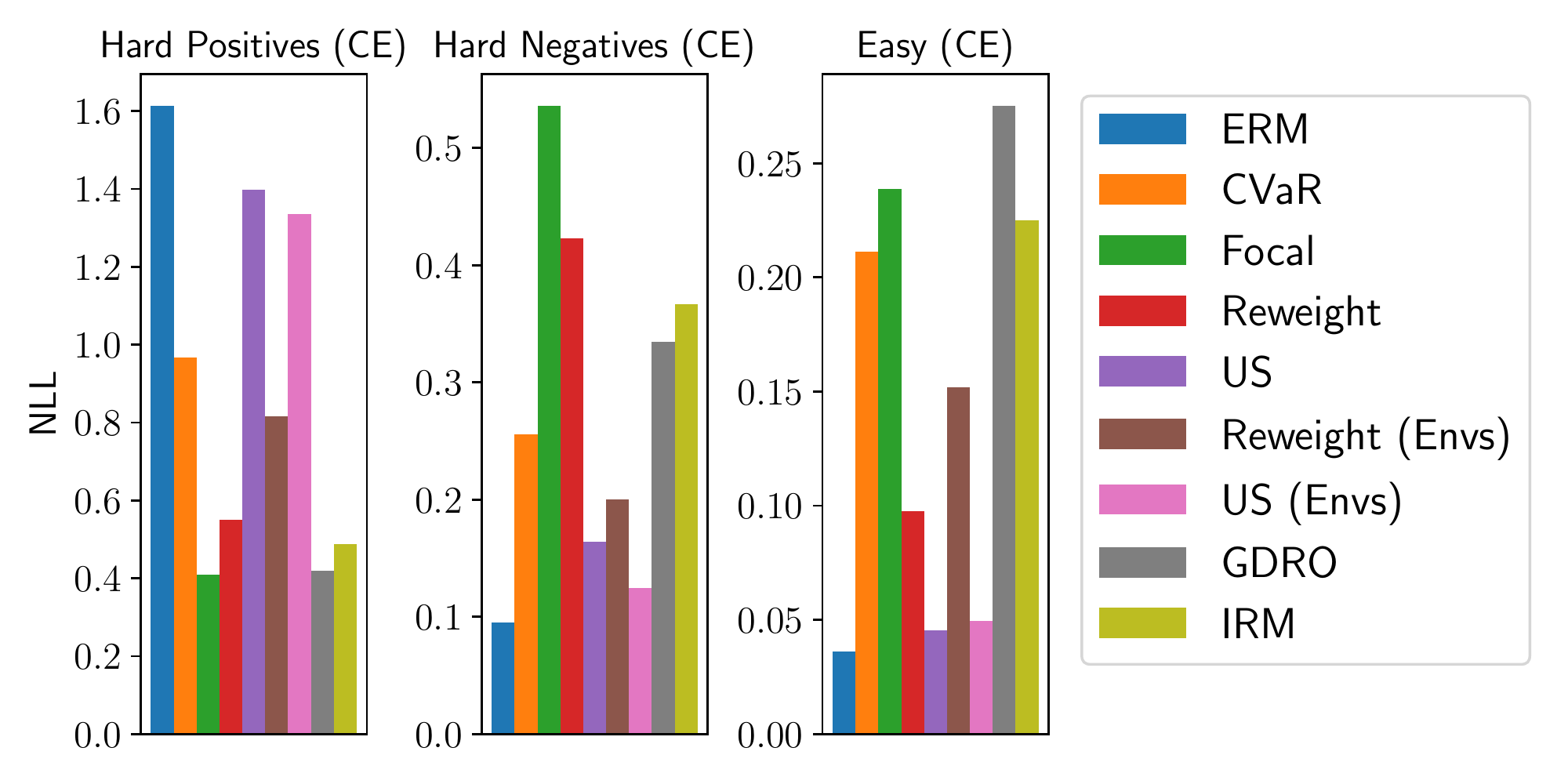}
% \caption{First subfigure} \label{fig:a}
\end{subfigure}\hspace*{\fill}
\begin{subfigure}{0.48\textwidth}
\centering
\includegraphics[width=0.9\linewidth]{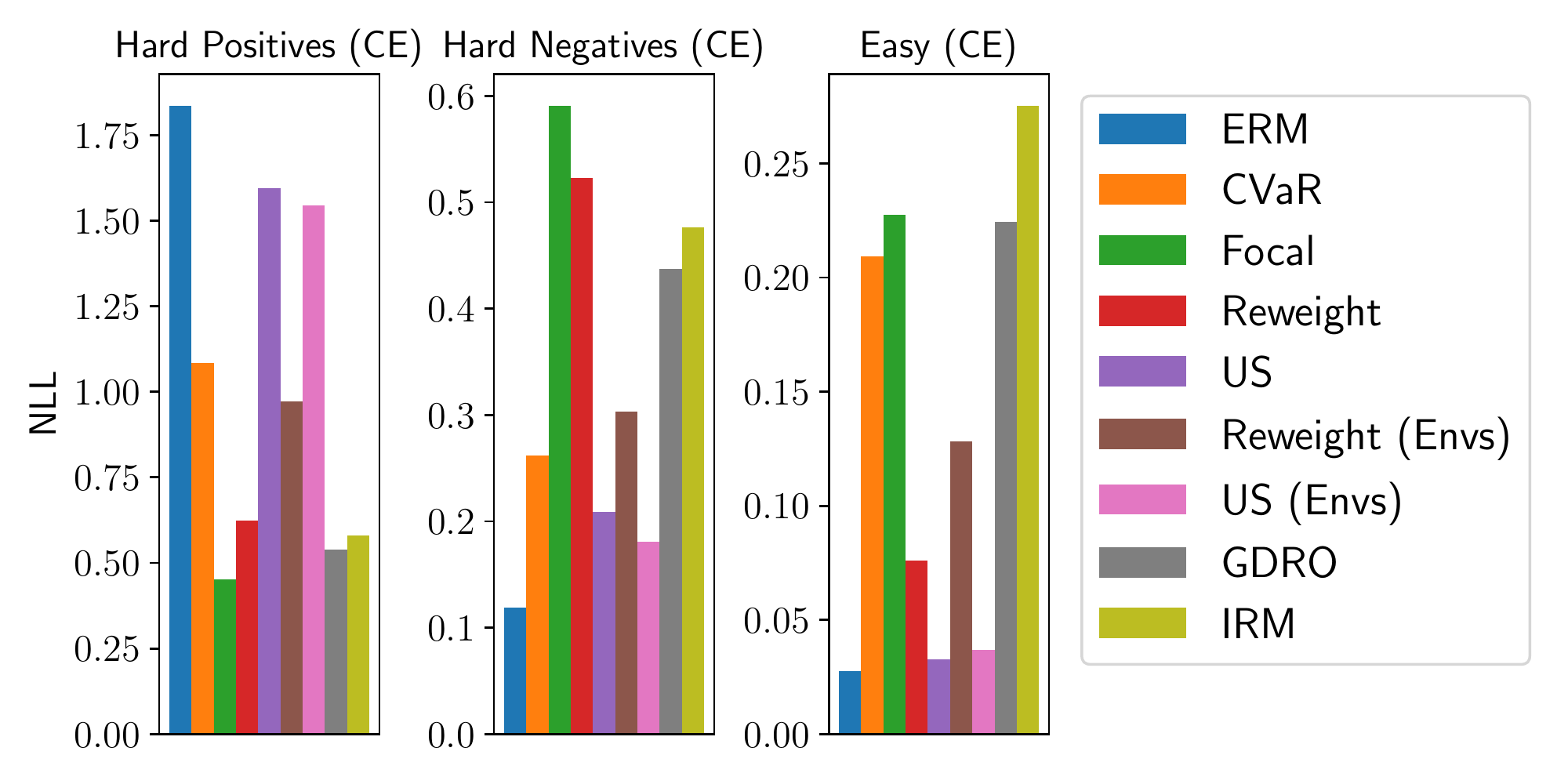}
% \caption{Second subfigure} \label{fig:b}
\end{subfigure}
\caption{
Negative log-likelihood (NLL) achieved on hard positive, hard negative, and easy examples.
% L: \rococo-CE.
% R: \rococo-Gist.
} \label{fig:results-nll-hard-posneg-easy}
\end{figure}

\begin{figure}[t!] % "[t!]" placement specifier just for this example
\begin{subfigure}{0.48\textwidth}
\centering
\includegraphics[width=0.9\linewidth]{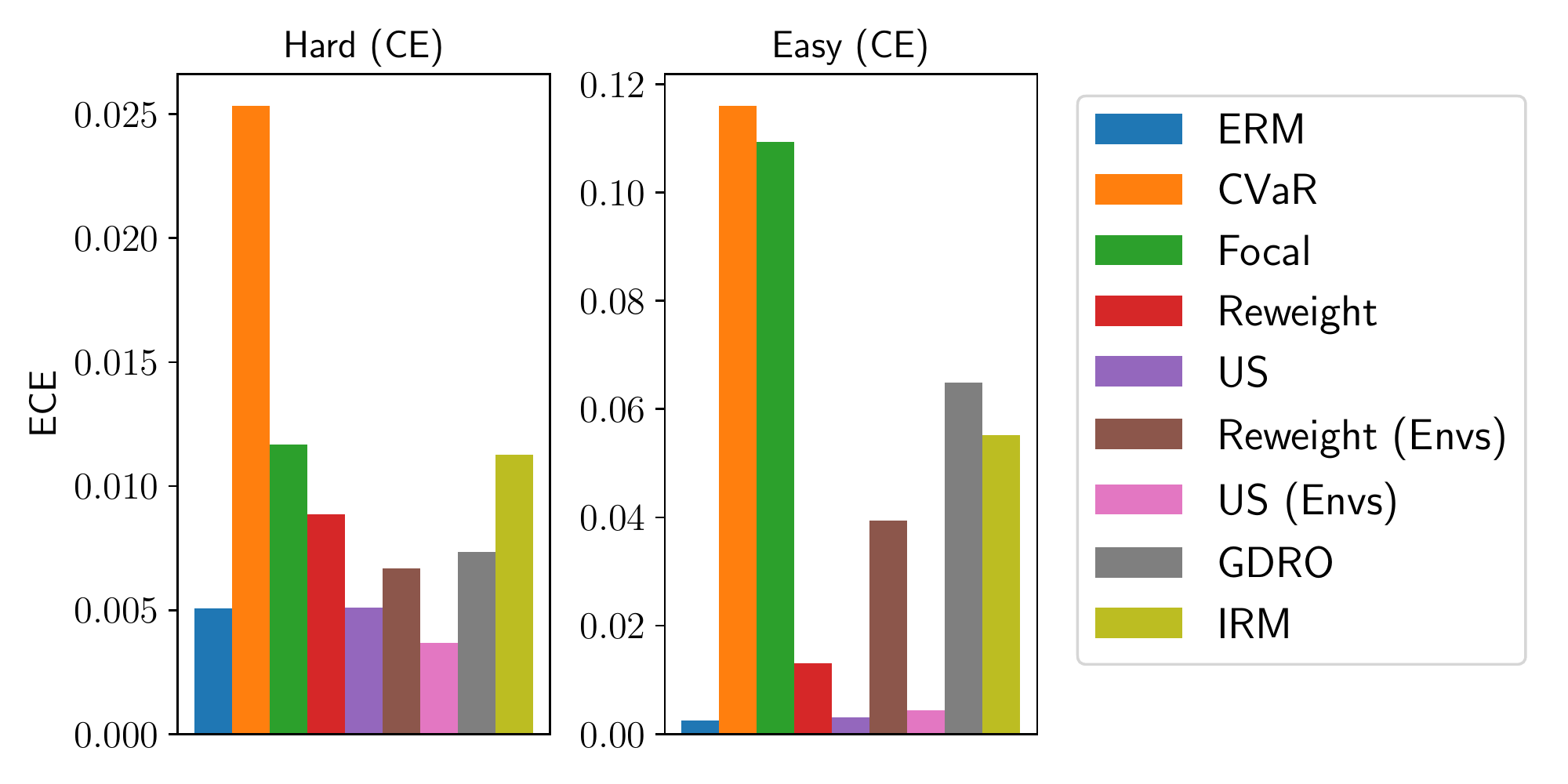}
% \caption{First subfigure} \label{fig:a}
\end{subfigure}\hspace*{\fill}
\begin{subfigure}{0.48\textwidth}
\centering
\includegraphics[width=0.9\linewidth]{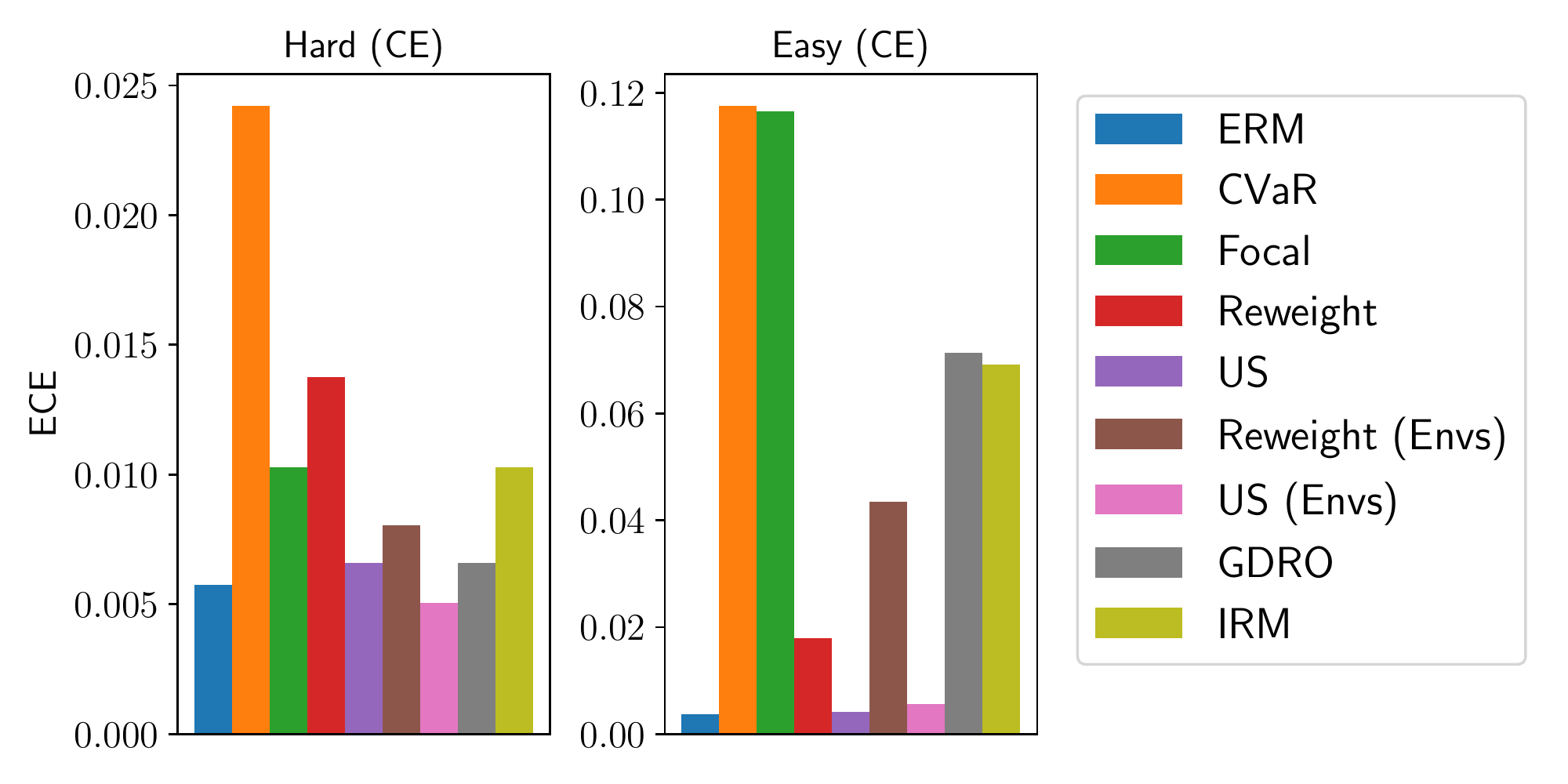}
% \caption{Second subfigure} \label{fig:b}
\end{subfigure}
\caption{
Expected Calibration Error (ECE) achieved on hard and easy examples. Lower is better.
% L: \rococo-CE.
% R: \rococo-Gist.
} \label{fig:results-ece-hard-easy}
\end{figure}

In Figures \ref{fig:results-auc-hard-easy}, \ref{fig:results-err-hard-posneg-easy}, \ref{fig:results-nll-hard-posneg-easy}, and \ref{fig:results-ece-hard-easy}, we show AUC, classification error, mean negative log-likelihood (NLL), and expected calibration error (ECE) \citep{guo2017calibration} respectively, averaged across the 12 tasks.
For AUC and ECE, we display results for hard and easy examples separately (where an ``easy'' example is defined as one which is not ``hard'', i.e. not a hard positive or hard negative).
For error and NLL, we further break out the results on hard examples into hard positives and hard negatives.
For all, we show \rococo-CE (L) and \rococo-Gist (R) separately.

\paragraph{Tradeoffs Exist Between Harder and Easier Examples.}
We note two tradeoffs across AUC, error, and NLL results: models that are better on hard examples tend to be worse on easy examples (and vice versa), and models that are better on hard positives tend to be worse on hard negatives.
ERM is better on easy examples and hard negatives: since the dataset is imbalanced, the hard negatives are the easier ``hard'' examples, and we expect ERM to perform better here since negatives are the majority class.

\begin{wrapfigure}{L}{0.5\linewidth} % "[t!]" placement specifier just for this example
% \hspace*{-0.5cm}
\begin{subfigure}{0.25\textwidth}
\includegraphics[width=\linewidth]{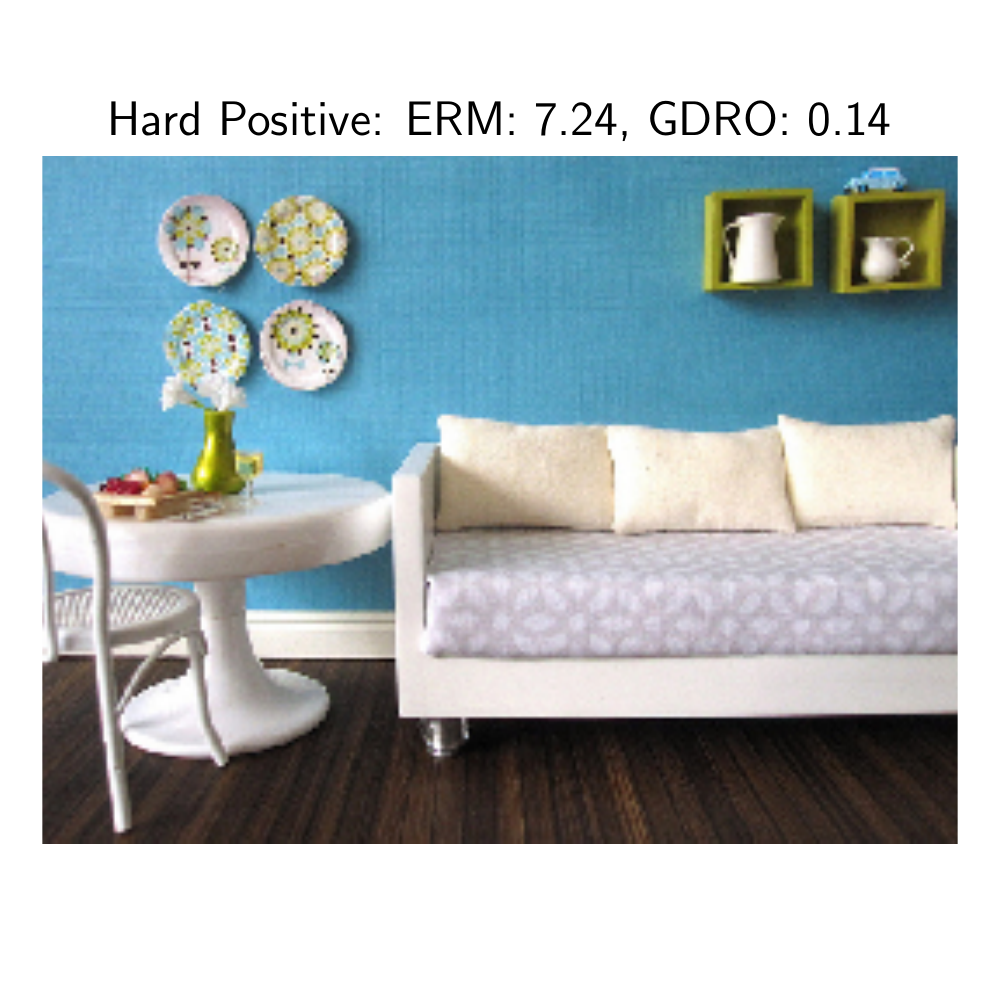}
% \caption{First subfigure} \label{fig:a}
\end{subfigure}%\hspace*{\fill}
% \hspace*{-2cm}
\begin{subfigure}{0.25\textwidth}
\includegraphics[width=\linewidth]{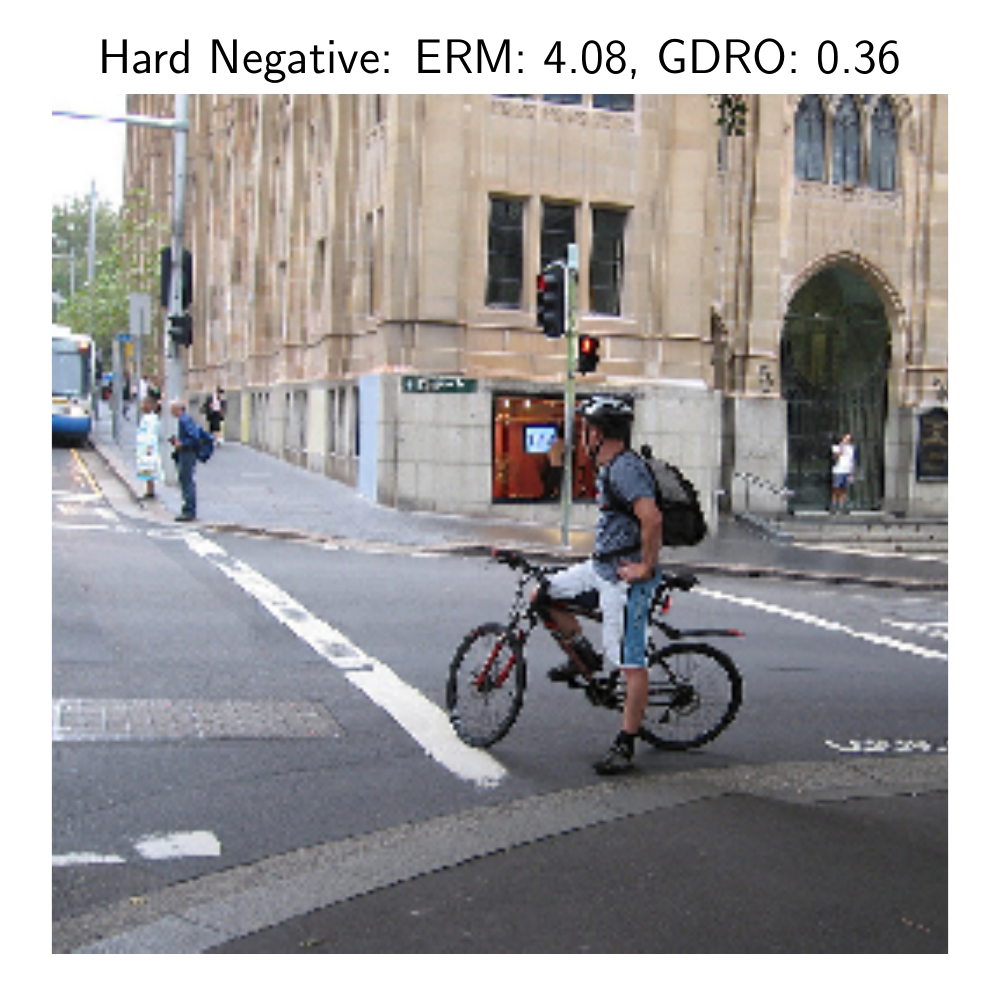}
% \caption{Second subfigure} \label{fig:b}
\end{subfigure}%\hspace*{\fill}
% \hspace*{-2cm}
% \begin{subfigure}{0.33\textwidth}
% \includegraphics[width=\linewidth]{figs/results/qual_examples/example_loss_diff_GDRO_over_ERM_car_Easy_Positive.pdf}
% % \caption{Second subfigure} \label{fig:b}
% \end{subfigure}
\caption{Test set examples where GDRO most improves over ERM (L) hard positives and (R) hard negatives
from the \rococo-CE \texttt{car} category.
Titles show NLL for each method on that image.
% From L to R: .
} \label{fig:results-qual-examples-car}
\end{wrapfigure}

\paragraph{Adaptive Methods Find Different Tradeoffs.}
We find that CVaR and focal loss find interesting tradeoffs between ERM and environment-based methods.
CVaR performs comparably to ERM on both hard and easy examples by AUC and error measures; and by ECE it is by far the worst of any method.
However, CVaR's NLL on hard positives is between ERM and the environment-based methods'.
Focal loss, on the other hand, performs similarly to ERM by AUC on both hard and easy examples, but it performs similarly to the environment-based methods on hard positives and negatives by error and NLL, providing a strong tradeoff between overall and OOC performance.

\paragraph{Two Surprising Calibration Results.}
We found that there was \textit{not} the same tradeoff with calibration as there was with the other metrics.
In particular, ERM and undersampling had the best calibration on \textit{both} hard and easy examples, suggesting there may be a path forward to avoid some of the tradeoffs between OOC and average performance by thinking more about calibration.
Secondly, we found the environment-based methods were not better calibrated, even on hard examples.

\subsection{Qualitative Analysis}

\paragraph{Where Environment-based Learning Improves Over ERM.}
In Fig. \ref{fig:results-qual-examples-car}, we show the images with the largest gaps in NLL on the \texttt{car} task among two groups: hard positives and hard negatives (in \rococo-CE).
The hard positive and hard negative where GDRO most overperformed are both images we might expect, where the object label and context do not match: a living room with a tiny toy car on the shelf, and a normal street scene that happens to not have any cars.
See App. \ref{app:qual} for analysis on where GDRO underperforms ERM.

\begin{wrapfigure}{L}{0.6\linewidth} % "[t!]" placement specifier just for this example
% \hspace*{-0.5cm}
\begin{subfigure}{0.32\textwidth}
\includegraphics[width=\linewidth]{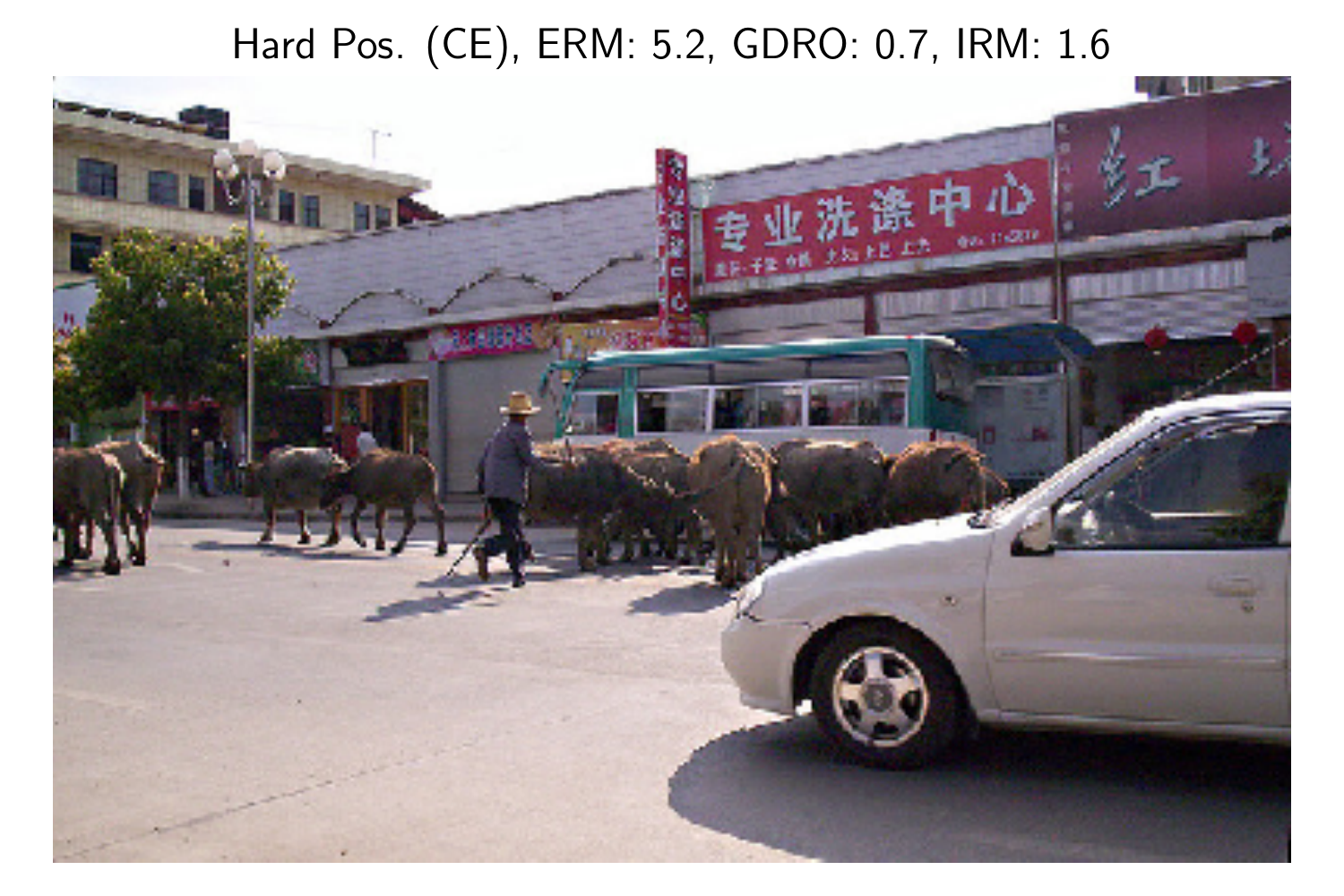}
% \caption{First subfigure} \label{fig:a}
\end{subfigure}%\hspace*{\fill}
% \hspace*{-2cm}
\begin{subfigure}{0.35\textwidth}
\includegraphics[width=\linewidth]{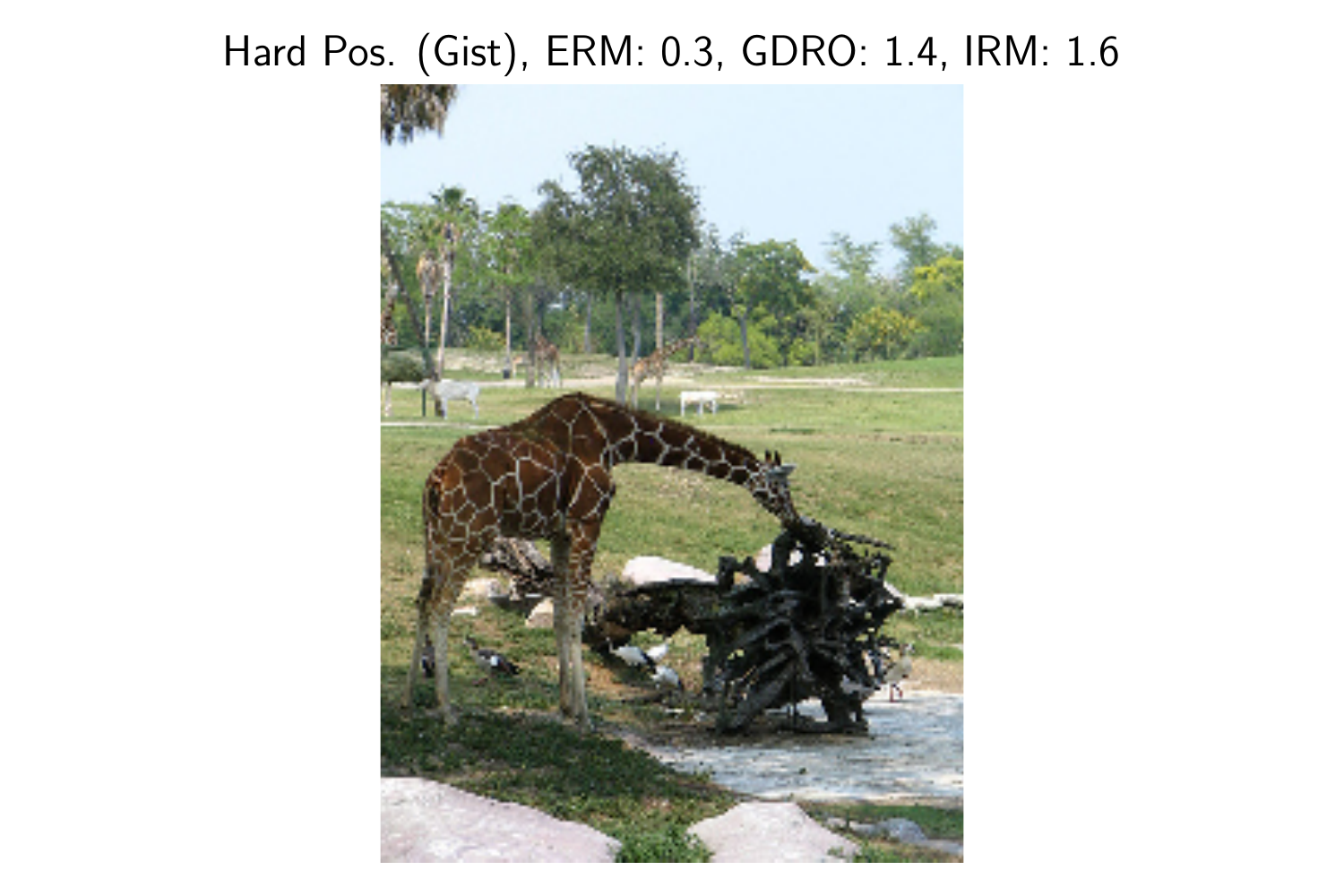}
% \caption{Second subfigure} \label{fig:b}
\end{subfigure}%\hspace*{\fill}
% \hspace*{-2cm}
% \begin{subfigure}{0.33\textwidth}
% \includegraphics[width=\linewidth]{figs/results/qual_examples/example_loss_diff_GDRO_over_ERM_car_Easy_Positive.pdf}
% % \caption{Second subfigure} \label{fig:b}
% \end{subfigure}
\caption{Examples of hard positives from \rococo-CE (L) and \rococo-Gist (R) for the \texttt{cow} task.
% est set examples where GDRO most improves over ERM (L) hard positives and (R) hard negatives
% from the \rococo-CE \texttt{car} category.
Titles show NLL for several methods on that image.
% From L to R: .
} \label{fig:results-qual-examples-critdiff}
\end{wrapfigure}

\paragraph{Contrasting \rococo-CE and \rococo-Gist.}
In Fig. \ref{fig:results-qual-examples-critdiff}, we use examples of hard positives from the \texttt{cow} task to contrast the CE and gist criteria.
On the left, we show a hard positive from \rococo-CE, with cows standing on pavement: this is a hard positive in \rococo-CE since there is no grass; GDRO and IRM outperform ERM on this image.
However, it is \textit{not} a hard positive in \rococo-Gist, since cows are the central focus of the image.
On the right, we show a hard positive from \rococo-Gist, where the image focuses on a giraffe, but there are several white cows in the background standing on a grassy field.
This is \textit{not} a hard positive in \rococo-CE, due to the large field, but it \textit{is} a hard positive in \rococo-Gist, since the giraffe is the focus of the image.
ERM outperforms GDRO and IRM on this image --- the environment-based objectives do not encourage as strong performance where the context (grass) and object (cow) align.

\section{Discussion} \label{sec:conclusion}

\paragraph{Limitations \& Impacts.}
The idea of computationally identifying OOC examples is necessarily limited: by their nature, OOC examples are exceptions to rules, and there will always be OOC examples which can only be discovered through qualitative examination.
Further, the task chosen for our benchmark was chosen for simplicity of analysis in a research context rather than real world significance --- outputting segmentations or many-way classification may be more applicable to most applications.
We hope that the impact of our work is to enable better evaluation of model performance on atypical or under-represented examples, both through usage of these challenge sets and ones inspired by the ideas presented here.
However, the usage of benchmarks can have downsides where standardized performance metrics are prioritized over fundamental advances in modelling which are not captured in those metrics.
In various applications of interest, results may need to be reported across a number of different metrics since this gives a clearer picture of model performance --- AUC may not always be a (or the most) relevant metric.

\paragraph{Conclusion \& Looking Forward.}

In this work, we study OOC evaluation as its own field, drawing attention to the range of evaluation schemes which can be used and the ways in which they may differ.
We demonstrate that using auxiliary structure in the data can be a useful means of defining context, and that this structure can be used in rich ways to identify various faces of the OOC problem.
Through this exploration, we find that methods which take advantage of auxiliary information may be more generously evaluated by OOC benchmarks which more are cleanly defined by that information.

In closing, we would like to highlight the idea systematizing OOC evaluation, whether this is through automatic discovery of OOC examples through annotations or some other means.
This not only enables the scalable creation of challenge sets, but allows for faster generation and exploration of new hypotheses about the type of context shifts that models struggle on, possibly analogous to automatic test generation in the debugging literature \citep{nebut2006automatic,cohen1996combinatorial}.
We hope that methods of this type can be applied elsewhere to better understand the challenges of OOC prediction.

\begin{ack}
Thanks to Marc-Etienne Brunet, Eleni Triantafilliou and Elliot Creager for their helpful thoughts on the manuscript, as well as to four anonymous reviewers for useful feedback.
David Madras was supported by an NSERC Alexander Graham Bell Canada Graduate Scholarship-Doctoral (CGS-D).
Resources used in preparing this research were provided, in part, by the
Province of Ontario, the Government of Canada through CIFAR, and companies
sponsoring the Vector Institute (\url{www.vectorinstitute.ai/\#partners}).
\end{ack}

% \section*{References}

\bibliography{bibliography}
\bibliographystyle{abbrvnat}

\clearpage
\appendix

\section{Other Methods} \label{app:other-methods}

% \paragraph{Other Methods}
We detail a few other methods which we explored but do not include for analysis in Sec. \ref{sec:experiments}:
\begin{itemize}
    \item Logistic regression on pre-trained Imagenet features: We found this performed noticeably worse than ERM across the board, with a slight improvement in loss on hard positives (but worse than most other methods).
    \item ERM trained from scratch: We found this to perform slightly worse across the board than ERM from a pre-trained model.
    \item ERM with learning rate decay: We explored several learning rate decays and found, with tuning, minor improvements in overall performance from ERM.
    Due to the level of tuning required, we chose not to include these results and just used a constant learning rate.
    \item ERM with data augmentation: We explored using data augmentations such as random affine transformations, cropping, and horizontal flips, but found they did not improve performance in ERM.
    % \item Undersampling of common environments \citep{tahir2012inverse}: We found this did not improve performance.
    \item Adaptive parameters \citep{perez2018film,triantafillou2021learning}: We explored the possibility of learning affine transformations after each ResNet block.
    With tuning, we found this improved performance on hard examples over ERM but not close to the level of the other robust methods shown.
    \item Auxiliary prediction: We explored adding a second readout head to the final layer and predicting the highest $\alpha$-context variable as an auxiliary loss.
    We found this to perform nearly identically to ERM in all metrics, and occasionally slightly worse.
\end{itemize}
We include this list for completeness: this is not to say that no such approach can be useful for this problem, but at the time of writing we did not find compelling enough results from these methods to yield interesting or informative comparative analysis in Sec. \ref{sec:experiments}.

\section{Data} \label{app:data}

\subsection{Licences}
\begin{itemize}
    \item The images in the COCO dataset \citep{lin2014microsoft} are provided under the Flickr terms of use.
    \item The annotations in the COCO dataset \citep{lin2014microsoft} are provided under a Creative Commons Attribution 4.0 License.
    \item The annotations in the COCO-Stuff dataset \citep{caesar2018coco} are provided under a Creative Commons Attribution 4.0 License.
\end{itemize}

\subsection{Dataset Details}
We use the COCO-Stuff dataset \citep{caesar2018coco} (\url{https://github.com/nightrome/cocostuff}), which contains images and annotaitons from the COCO dataset \citep{lin2014microsoft}, as well as adding additional annotations.
Since COCO is a competitive benchmark, the test set is not publicly available, so we merge the provided training/validation sets and create our own new training/validation/test split using 70/10/20 percentages.
This yields training/validation/test set sizes of 86302/12328/24657.

\subsection{Creation of the {\rococo \ } Suite}

When choosing the tasks, we filtered out a couple of tasks which seemed subjectively too similar to ones already in the benchmark.
Specifically, we removed \texttt{sheep}, \texttt{handbag}, \texttt{bottle}, and \texttt{wine-glass} due to the fact that other similar tasks had higher average difference in average NLL on all examples and average NLL on hard examples (specifically, \texttt{cow}, \texttt{backpack}, \texttt{cup}, and \texttt{cup} respectively).
We also filtered out those tasks without at least 50 hard positives and hard negatives.

% \begin{wrapfigure}{R}{0.4\textwidth}
% \centering
%   \includegraphics[scale=0.4]{figs/benchmark/num_tasks_by_num_confs.pdf}
%     \caption{The number of tasks in the COCO-Stuff dataset which have context variables identifiable at each level of $\alpha$.
%     There are 171 total tasks.
%     For instance, this figure shows that there are around 70 tasks with at least one context cue at $\alpha \geq 0.05$, and around 30 tasks with at least five context cues at $\alpha \geq 0$.
%     }
%     \label{fig:benchmark-num-tasks-num-confs}
% \end{wrapfigure}

% Alternatively, in Fig [fig], we can see that about 50\% of test images qualify as hard positives in some category, and about 85\% of test images qualify as hard negatives in some category.

% \begin{figure}
%     \centering
%     \includegraphics[scale=0.5]{figs/benchmark/num_tasks_by_num_confs.pdf}
%     \caption{The number of tasks in the COCO-Stuff dataset which have context variables identifiable at each level of $\alpha$.
%     There are 171 total tasks.
%     For instance, this figure shows that there are around 70 tasks with at least one context cue at $\alpha \geq 0.05$, and around 30 tasks with at least five context cues at $\alpha \geq 0$.
%     }
%     \label{fig:benchmark-num-tasks-num-confs}
% \end{figure}

\begin{figure}
\centering
\begin{minipage}{.48\textwidth}
  \centering
  \includegraphics[scale=0.4]{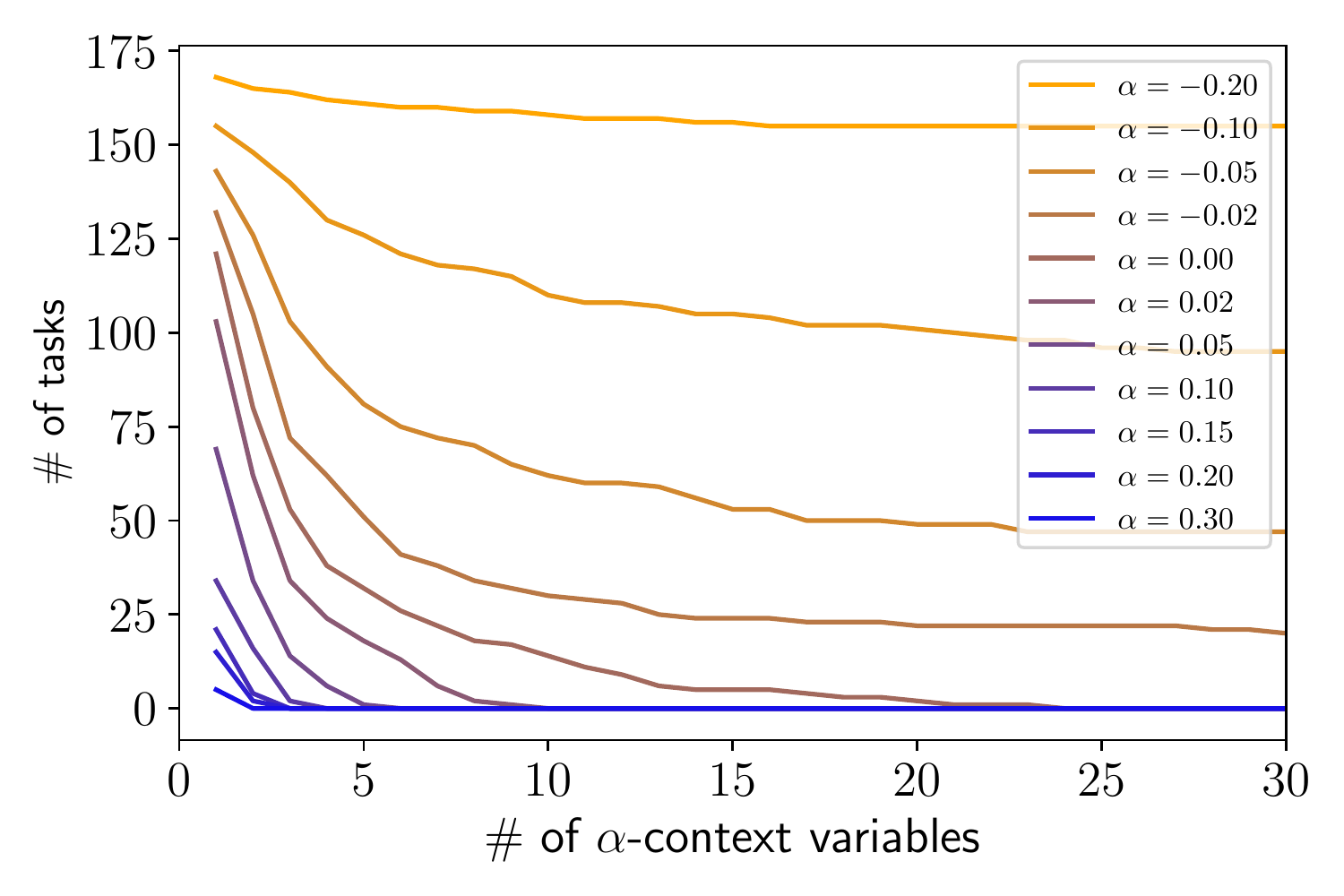}
    \caption{The number of tasks in the COCO-Stuff dataset which have context variables identifiable at each level of $\alpha$.
    There are 171 total tasks.
    For instance, this figure shows that there are around 70 tasks with at least one context cue at $\alpha \geq 0.05$, and around 30 tasks with at least five context cues at $\alpha \geq 0$.
    }
    \label{fig:benchmark-num-tasks-num-confs}
\end{minipage}\hspace*{\fill}
\begin{minipage}{.48\textwidth}
  \centering
    \includegraphics[scale=0.4]{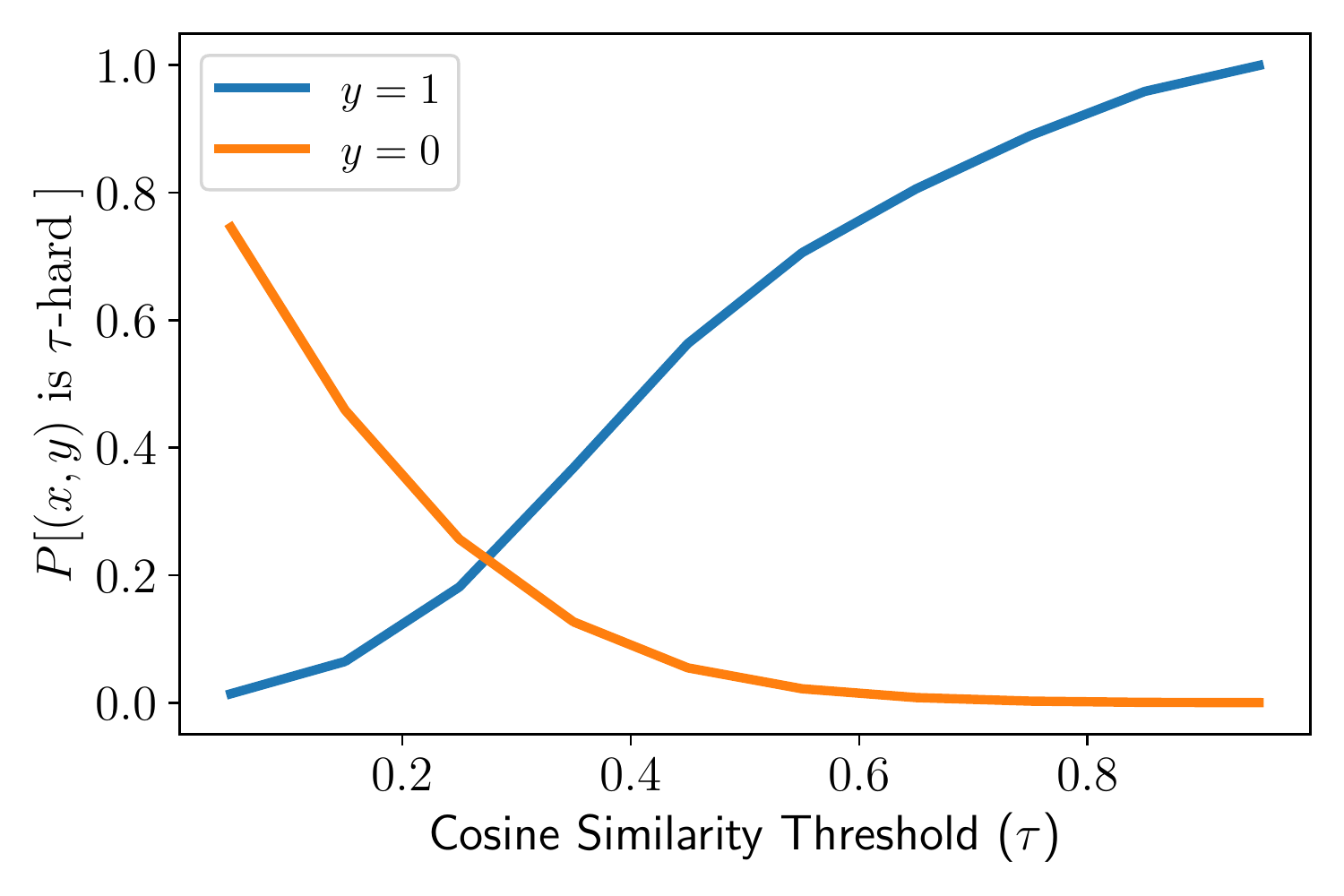}
    \caption{Using the gist criterion, the number of $\tau$-hard positive/negative examples in the COCO-Stuff test set as a proportion of all positive/negative examples, where $\tau$ thresholds the cosine similarity to the average caption embedding of that category in the training set.
    }
    \label{fig:benchmark-hard-pos-neg-examples-caption}

\end{minipage}
\end{figure}

Fig \ref{fig:benchmark-num-tasks-num-confs} shows that, under the CE criterion, most of the 171 object classes in COCO-Stuff have context cues for $\alpha \geq 0$.
Note that a 0-context cue is fairly strong: this means that on average, in images where $Y$ is present, $C$ takes up at least as much area as $Y$ does.
This indicates that there may be a large number of OOC examples hidden in many of these tasks.
% For instance, over two-thirds of tasks have at least one 0-context cue, and over 1/3 have at least one 0.05-context cue.
We find that the contexts returned using this method for large enough $\alpha$ are intuitive.
Some examples of (label, 0.05-context) pairs are: (\texttt{car}, \texttt{road}), (\texttt{bowl}, \texttt{dining\_table}), (\texttt{cow}, \texttt{grass}).
Additionally, many tasks have more than one context variable: at $\alpha=0$ about 50 tasks have 3, and at $\alpha=0.05$ about 15 tasks have 3.
Some examples of labels with multiple 0.05-context variables are: (\texttt{surfboard}, [\texttt{sea}, \texttt{sky-other}]),
(\texttt{tennis-racket}, [\texttt{person}, \texttt{playing-field}]),
(\texttt{traffic-light}, [\texttt{building-other}, \texttt{road}, \texttt{sky-other}, \texttt{tree}]).
All of these groupings are intuitive, providing some evidence that this method of isolating context variables may be a useful one.

In Fig. \ref{fig:benchmark-hard-pos-neg-examples-caption} we see, across all values of $\tau$, how many $\tau$-hard examples there are in the test set by the gist criterion, across all 171 tasks in COCO-Stuff.
As expected, the number of hard positives increases with $\tau$, and the number of hard negatives decreased.

\begin{figure}[t!] % "[t!]" placement specifier just for this example
\begin{subfigure}{0.48\textwidth}
\includegraphics[width=\linewidth]{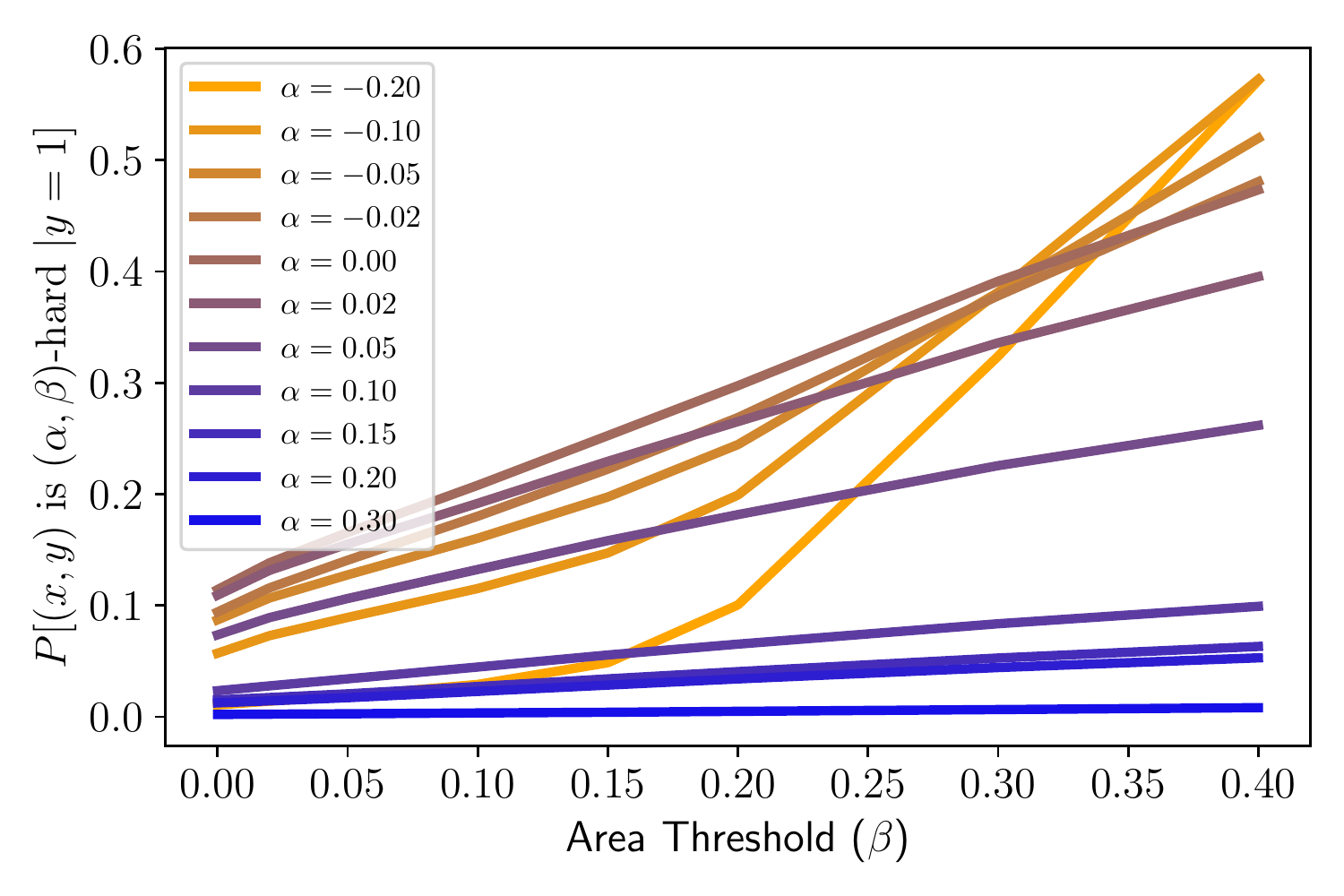}
% \caption{First subfigure} \label{fig:a}
\end{subfigure}\hspace*{\fill}
\begin{subfigure}{0.48\textwidth}
\includegraphics[width=\linewidth]{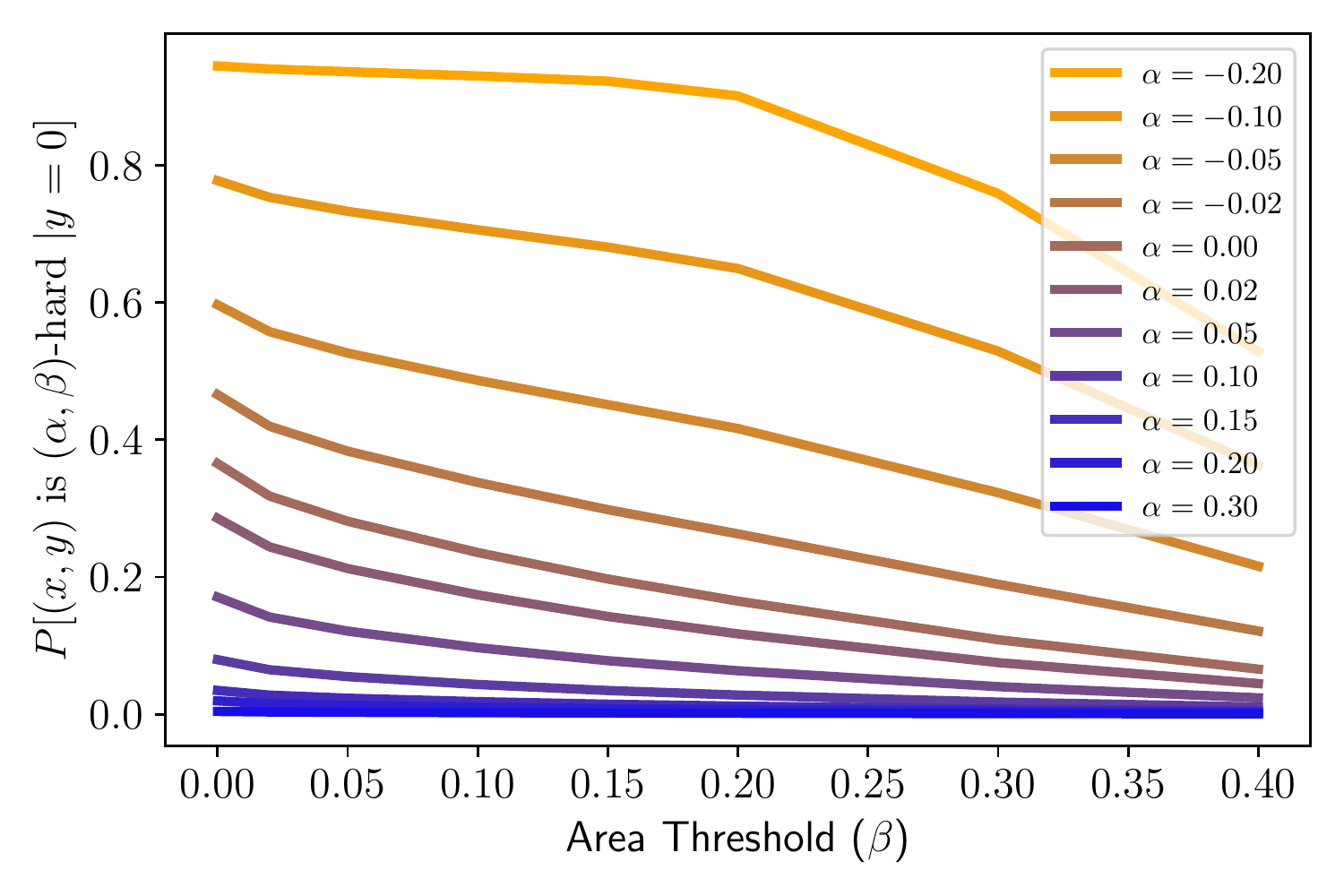}
% \caption{Second subfigure} \label{fig:b}
\end{subfigure}
\caption{Across a range of $\alpha, \beta$, the percentage of hard positive/negative examples (L/R) in the test set which are $(\alpha, \beta)$-hard.
} \label{fig:benchmark-hard-pos-neg-examples-area}
\end{figure}

Fig \ref{fig:benchmark-hard-pos-neg-examples-area} shows that hard positive/negative examples are numerous in the data.
For instance, we see that at the (0.05, 0.1) level, about 10\% of all possible (object, category) pairs are hard positives, and a similar number for hard negatives.

\begin{figure}[t!] % "[t!]" placement specifier just for this example
\begin{subfigure}{0.48\textwidth}
\includegraphics[width=\linewidth]{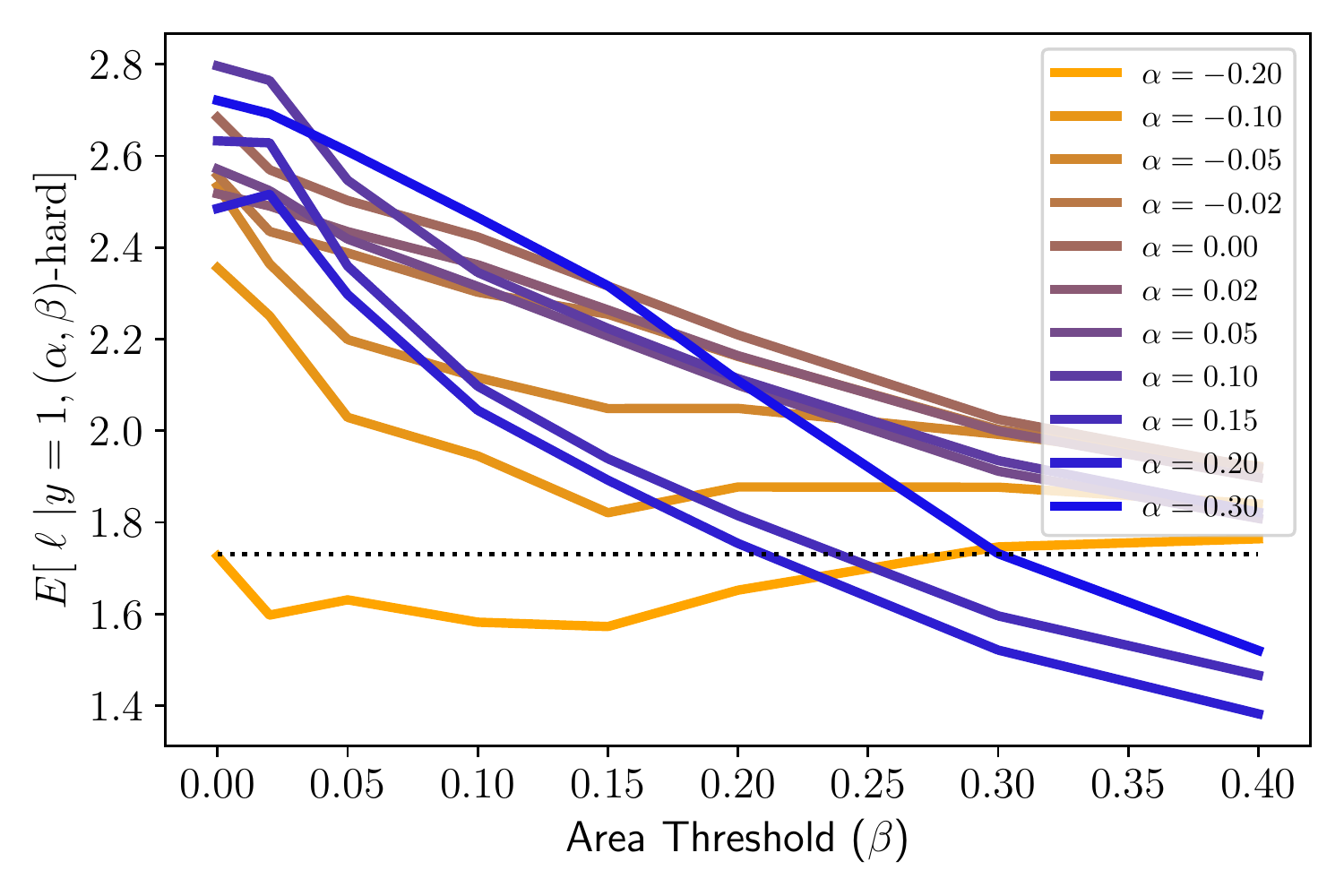}
% \caption{First subfigure} \label{fig:a}
\end{subfigure}\hspace*{\fill}
\begin{subfigure}{0.48\textwidth}
\includegraphics[width=\linewidth]{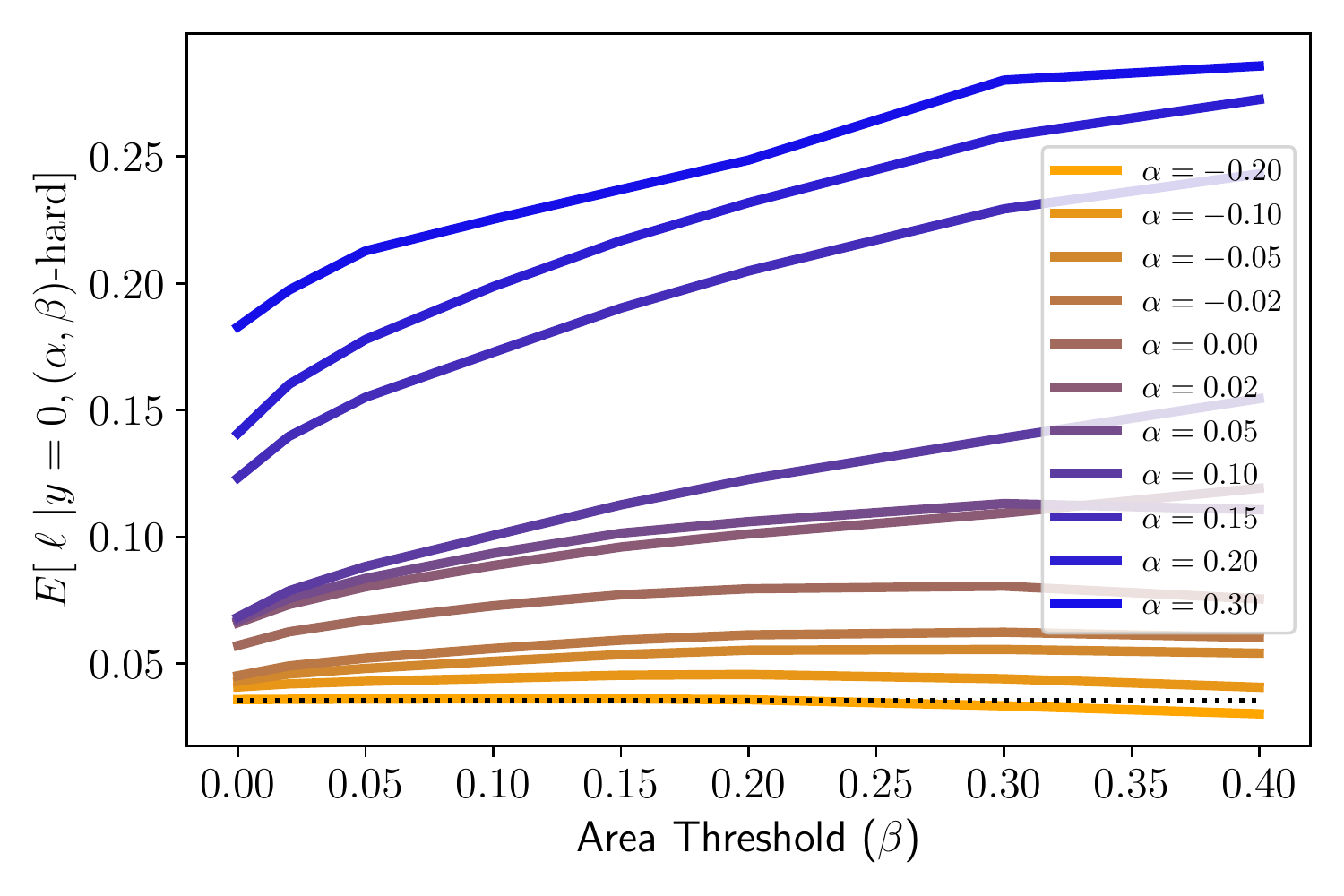}
% \caption{Second subfigure} \label{fig:b}
\end{subfigure}
\caption{Across a range of $\alpha, \beta$, the average test set loss of positive/negative examples (L/R) in the test set which are $(\alpha, \beta)$-hard.
} \label{fig:benchmark-hard-pos-neg-examples-area-loss}
\end{figure}

Fig. \ref{fig:benchmark-hard-pos-neg-examples-area-loss} shows the average loss incurred by $(\alpha, \beta)$-hard positive and negative examples under the CE criterion as $\alpha, \beta$ vary.

\begin{figure}
    \centering
    \includegraphics[scale=0.5]{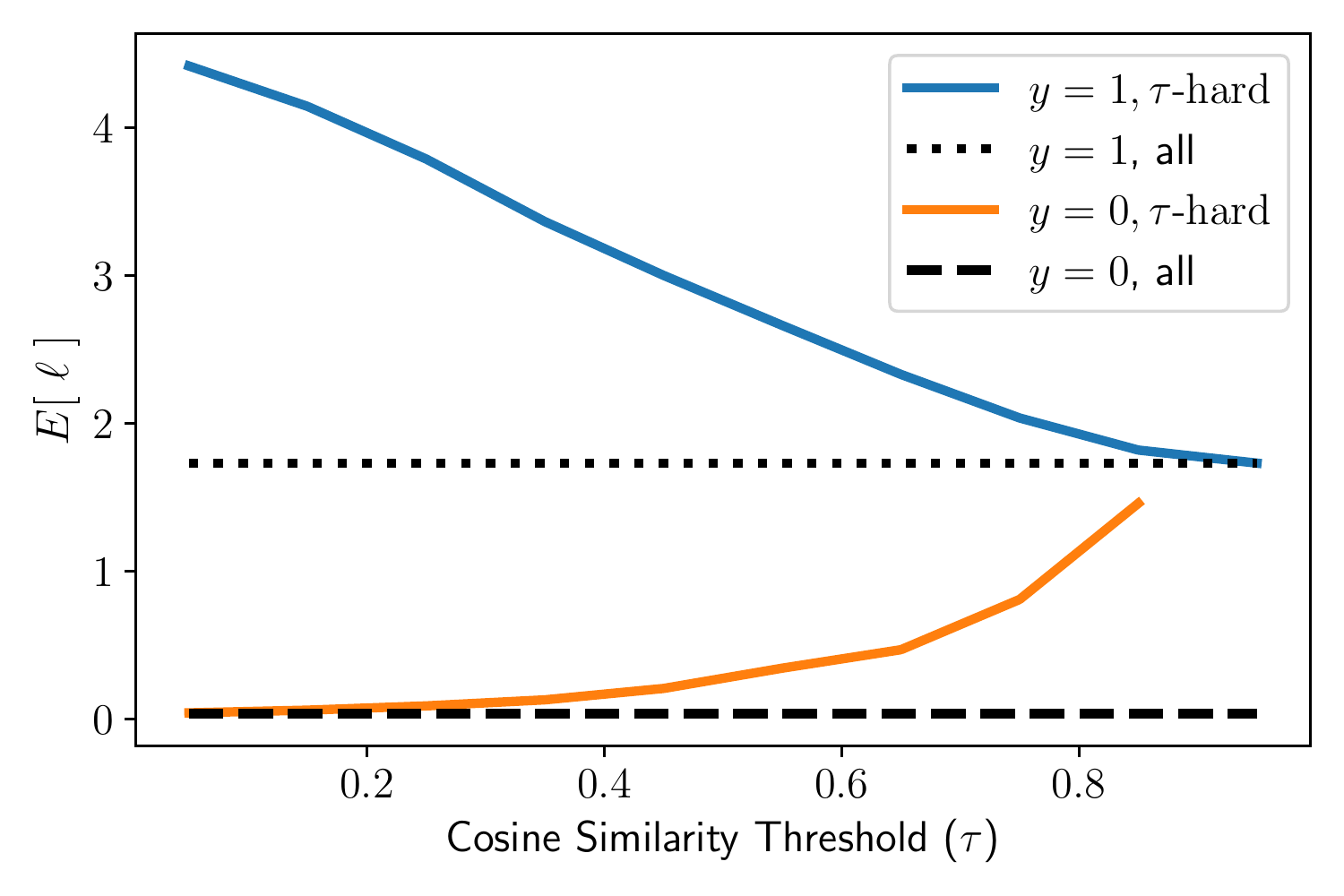}
    \caption{Caption}
    \label{fig:benchmark-hard-losses-by-caption}
\end{figure}

Fig. \ref{fig:benchmark-hard-losses-by-caption} shows the average loss incurred by $(\tau)$-hard positive and negative examples under the gist criterion as $\tau$ varies.

\begin{table}
\centering
\begin{minipage}{.48\textwidth}
    \centering
    \begin{tabular}{lrr}
\hline
 Task         &   Hard Positives &   Hard Negatives \\
\hline
 car          &             0.48 &             0.48 \\
 bowl         &             0.64 &             0.57 \\
 boat         &             0.73 &             0.43 \\
 fire-hydrant &             0.74 &             0.35 \\
 airplane     &             0.81 &             0.19 \\
 cow          &             0.73 &             0.24 \\
 backpack     &             0.39 &             0.3  \\
 cup          &             0.46 &             0.23 \\
 surfboard    &             0.66 &             0.21 \\
 tie          &             0.26 &             0.23 \\
 sports-ball  &             0.47 &             0.12 \\
 kite         &             0.57 &             0.19 \\
\hline
\end{tabular}

    \caption{Values for $\tau$ used for the gist criterion throughout the paper (test set).}
    \label{tab:tau-values-test}
\end{minipage}\hspace*{\fill}
\begin{minipage}{.48\textwidth}
    \centering
    \begin{tabular}{lrr}
\hline
 Task         &   Hard Positives &   Hard Negatives \\
\hline
 car          &             0.47 &             0.5  \\
 bowl         &             0.65 &             0.57 \\
 boat         &             0.72 &             0.42 \\
 fire-hydrant &             0.76 &             0.35 \\
 airplane     &             0.81 &             0.19 \\
 cow          &             0.71 &             0.23 \\
 backpack     &             0.39 &             0.31 \\
 cup          &             0.46 &             0.23 \\
 surfboard    &             0.68 &             0.21 \\
 tie          &             0.25 &             0.23 \\
 sports-ball  &             0.44 &             0.12 \\
 kite         &             0.58 &             0.19 \\
\hline
\end{tabular}
    \caption{Values for $\tau$ used for the gist criterion throughout the paper (validation set).}
    \label{tab:tau-values-valid}
\end{minipage}
\end{table}

In the paper, we choose $\alpha, \beta = 0.05, 0.1$ for the CE criterion, and then set $\tau$ in each task such that the number of hard positive and negative examples in each task is the same.
Tables \ref{tab:tau-values-test} and \ref{tab:tau-values-valid} show these values of $\tau$ for each for the 12 {\rococo\ } tasks.
\section{Experimental Details} \label{app:experiments}

\subsection{Licences}
The Resnet-50 is provided under an Apache 2.0 licence.

\subsection{General Hyperparameters}
We train (where not otherwise indicated) using SGD with learning rate of 1e-4, momentum of 0.9, L2 weight regularization of 1e-4, and batch size 32 (the largest we could fit on the available GPUs).
% , finetuning a pretrained ResNet-50.
For all methods, we use early stopping on the validation loss with a patience of 3 epochs.
All images were resized to $321 \times 321$.

\paragraph{Training Parameters for ERM in Sec. \ref{sec:benchmark}.}
For the experiments which refer to all 171 tasks (rather than just the 12 we focus on in {\rococo \ }, we use an Adam optimizer with 1e-3 learning rate and batch size 16.
All other training parameters are the same as noted elsewhere.

\paragraph{Sentence Embeddings.}
To computer sentence embeddings, we use \texttt{stsb-distilroberta-base-v2} from the \texttt{sentence\_transformers} package (\url{https://www.sbert.net/}).

\subsection{Compute}

We used GPUs on internal clusters, utilizing Nvidia Titan XP, T4, and P100 GPUs.
The total number of jobs run to produce our main results, i.e. Table \ref{tab:auc-by-task} and all associated figures, is 
% \begin{equation*}
\begin{align}
    &\textrm{num\_tasks} \times \textrm{num\_seeds} \times \textrm{num\_methods} \times \textrm{num\_hyperparameters\_per\_method}\\
    = & 12 \times 3 \times 5 \times 3.2 = 576
\end{align}
% \end{equation*}
where we calculate \textrm{num\_hyperparameters\_per\_method} from Table \ref{tab:hyperparameters}.
The job runtime lengths varied widely due to the use of early stopping.

\subsection{Method-Specific Hyperparameters}
Each of the methods we run other than ERM has a hyperparameter which trades off better overall performance and OOC performance.
For each method, we swept over 4 values of these parameters (3 for CVaR), running 3 seeds at each value.
The values of these hyperparameters are shown in Table \ref{tab:hyperparameters}.
They were chosen through some manual experimentation to get a sense of the useful ranges for each.

\begin{table}[]
    \centering
    
\begin{tabular}{ccc}
\hline
\textbf{Method} & \textbf{Hyperparameter Name} & \textbf{Values}        \\ \hline
GDRO            & $K$                          & {[}5, 30, 60, 240{]}   \\
IRM             & $\lambda$                    & {[}0.1, 1, 3, 10{]}    \\
CVaR            & $p$                          & {[}0.05, 0.1, 0.15{]}  \\
Focal           & $\gamma$                     & {[}0.2, 0.5, 0.7, 1{]} \\
Label Reweighting           & $\alpha$                     & {[}0.2, 0.5, 0.7, 1{]} \\
Label Undersampling           & $\alpha$                     & {[}0.2, 0.5, 0.7, 1{]} \\
Environment Reweighting           & $\alpha$                     & {[}0.5, 1, 1.5, 2{]} \\
Environment Undersampling           & $\alpha$                     & {[}0.2, 0.5, 0.7, 1{]} \\
\hline
\end{tabular}

    \caption{Hyperparameter lists for each method (ERM has no method-specific hyperparameters).}
    \label{tab:hyperparameters}
\end{table}

\subsection{Detailed Results}

For each metric of AUC, NLL, Error, and ECE, each of the 12 tasks in {\rococo \ }, and both \rococo-CE and \rococo-Gist, we show results below with standard deviation across 3 seeds.
There are the same results as Figures \ref{fig:results-auc-hard-easy}, \ref{fig:results-err-hard-posneg-easy}, \ref{fig:results-nll-hard-posneg-easy}, are \ref{fig:results-ece-hard-easy}, but shown in more detail.

\begin{table}
\resizebox{0.5\columnwidth}{!}{%
\centering
\begin{tabular}{llllllllll}
\hline
 Task         & ERM                    & CVaR          & Focal         & Reweight               & US                     & Reweight (Envs)        & US (Envs)              & GDRO                   & IRM                    \\
\hline
 car          & 0.769 (0.002)          & 0.787 (0.007) & 0.759 (0.005) & 0.773 (0.001)          & 0.773 (0.003)          & \textbf{0.886 (0.002)} & 0.846 (0.013)          & \textbf{0.891 (0.006)} & 0.868 (0.004)          \\
 bowl         & 0.749 (0.003)          & 0.781 (0.013) & 0.734 (0.004) & 0.751 (0.004)          & 0.759 (0.007)          & \textbf{0.864 (0.003)} & 0.814 (0.006)          & \textbf{0.865 (0.011)} & 0.828 (0.023)          \\
 boat         & 0.869 (0.002)          & 0.888 (0.018) & 0.823 (0.005) & 0.866 (0.004)          & 0.877 (0.006)          & \textbf{0.954 (0.002)} & 0.923 (0.002)          & \textbf{0.945 (0.012)} & 0.925 (0.004)          \\
 fire-hydrant & 0.913 (0.004)          & 0.933 (0.003) & 0.908 (0.003) & 0.927 (0.001)          & 0.913 (0.001)          & 0.933 (0.002)          & 0.920 (0.002)          & \textbf{0.946 (0.003)} & \textbf{0.942 (0.004)} \\
 airplane     & 0.986 (0.002)          & 0.984 (0.002) & 0.983 (0.001) & 0.986 (0.000)          & 0.985 (0.002)          & \textbf{0.991 (0.000)} & 0.986 (0.001)          & \textbf{0.991 (0.001)} & 0.986 (0.003)          \\
 cow          & 0.935 (0.003)          & 0.937 (0.003) & 0.932 (0.001) & 0.939 (0.002)          & 0.938 (0.003)          & \textbf{0.963 (0.001)} & 0.948 (0.002)          & \textbf{0.963 (0.004)} & 0.943 (0.001)          \\
 backpack     & 0.812 (0.004)          & 0.806 (0.010) & 0.809 (0.002) & 0.816 (0.003)          & 0.812 (0.008)          & 0.813 (0.002)          & 0.816 (0.014)          & \textbf{0.871 (0.008)} & 0.844 (0.018)          \\
\hline
\end{tabular}
% \caption{Foo}
}
\hfill
\resizebox{0.5\columnwidth}{!}{%
\centering
\begin{tabular}{llllllllll}
\hline
 Task         & ERM                    & CVaR                   & Focal                  & Reweight               & US                     & Reweight (Envs)        & US (Envs)              & GDRO                   & IRM                    \\
\hline
 car          & 0.766 (0.002)          & 0.785 (0.003)          & 0.763 (0.005)          & 0.768 (0.005)          & 0.773 (0.006)          & 0.845 (0.001)          & 0.823 (0.004)          & \textbf{0.863 (0.008)} & 0.832 (0.017)          \\
 bowl         & 0.642 (0.004)          & \textbf{0.686 (0.015)} & 0.635 (0.002)          & 0.656 (0.006)          & 0.678 (0.009)          & \textbf{0.692 (0.007)} & \textbf{0.698 (0.006)} & \textbf{0.704 (0.013)} & \textbf{0.670 (0.021)} \\
 boat         & 0.826 (0.002)          & 0.855 (0.019)          & 0.771 (0.008)          & 0.823 (0.004)          & 0.823 (0.002)          & \textbf{0.897 (0.002)} & 0.868 (0.003)          & 0.884 (0.008)          & 0.872 (0.002)          \\
 fire-hydrant & 0.840 (0.007)          & \textbf{0.865 (0.008)} & 0.822 (0.005)          & 0.849 (0.002)          & 0.841 (0.007)          & 0.834 (0.005)          & 0.836 (0.007)          & 0.823 (0.007)          & 0.845 (0.004)          \\
 airplane     & \textbf{0.977 (0.001)} & \textbf{0.979 (0.000)} & 0.975 (0.001)          & \textbf{0.978 (0.000)} & 0.976 (0.001)          & \textbf{0.978 (0.001)} & 0.977 (0.001)          & \textbf{0.981 (0.003)} & \textbf{0.977 (0.002)} \\
 cow          & 0.901 (0.004)          & \textbf{0.912 (0.001)} & 0.877 (0.002)          & 0.903 (0.002)          & 0.906 (0.004)          & 0.908 (0.002)          & \textbf{0.908 (0.003)} & \textbf{0.911 (0.006)} & 0.896 (0.008)          \\
 backpack     & 0.716 (0.004)          & 0.717 (0.013)          & 0.725 (0.002)          & 0.731 (0.003)          & \textbf{0.749 (0.004)} & 0.729 (0.004)          & 0.736 (0.003)          & \textbf{0.732 (0.012)} & \textbf{0.750 (0.010)} \\
\hline
\end{tabular}
% \caption{Bar}
}
\caption{AUC on hard test examples for all 12 \rococo \ stress tests, after hyperparameter selection. 
Bold numbers have overlapping standard deviations.
L: \rococo-CE.
R: \rococo-Gist.
}
\label{tab:hard-auc-by-task-std}
\end{table}

\begin{table}
\resizebox{0.5\columnwidth}{!}{%
\centering
\begin{tabular}{llllllllll}
\hline
 Task         & ERM                    & CVaR                   & Focal                  & Reweight               & US                     & Reweight (Envs)        & US (Envs)              & GDRO                   & IRM                    \\
\hline
 car          & \textbf{0.988 (0.000)} & 0.987 (0.000)          & 0.986 (0.000)          & \textbf{0.988 (0.000)} & 0.987 (0.000)          & 0.939 (0.001)          & 0.961 (0.004)          & 0.912 (0.014)          & 0.959 (0.003)          \\
 bowl         & \textbf{0.983 (0.000)} & \textbf{0.981 (0.002)} & 0.982 (0.000)          & \textbf{0.983 (0.000)} & 0.980 (0.002)          & 0.913 (0.004)          & 0.968 (0.004)          & 0.881 (0.037)          & 0.943 (0.007)          \\
 boat         & 0.995 (0.000)          & \textbf{0.996 (0.001)} & \textbf{0.997 (0.000)} & 0.996 (0.000)          & 0.995 (0.000)          & 0.961 (0.003)          & 0.982 (0.001)          & 0.961 (0.011)          & 0.984 (0.005)          \\
 fire-hydrant & 0.979 (0.001)          & 0.982 (0.003)          & \textbf{0.985 (0.001)} & \textbf{0.985 (0.001)} & 0.978 (0.001)          & 0.965 (0.002)          & 0.977 (0.004)          & 0.968 (0.004)          & 0.981 (0.001)          \\
 airplane     & 0.997 (0.001)          & \textbf{0.999 (0.001)} & \textbf{0.998 (0.000)} & 0.997 (0.000)          & 0.996 (0.000)          & 0.995 (0.000)          & 0.996 (0.000)          & \textbf{0.997 (0.002)} & \textbf{0.998 (0.002)} \\
 cow          & 0.995 (0.001)          & 0.993 (0.002)          & \textbf{0.997 (0.000)} & 0.997 (0.000)          & 0.995 (0.001)          & 0.989 (0.001)          & 0.995 (0.001)          & 0.981 (0.007)          & 0.995 (0.001)          \\
 backpack     & 0.932 (0.001)          & 0.910 (0.005)          & \textbf{0.937 (0.001)} & \textbf{0.937 (0.001)} & 0.931 (0.003)          & \textbf{0.935 (0.002)} & \textbf{0.932 (0.006)} & 0.877 (0.011)          & \textbf{0.928 (0.008)} \\
\hline
\end{tabular}
% \caption{Foo}
}
\hfill
\resizebox{0.5\columnwidth}{!}{%
\centering
\begin{tabular}{llllllllll}
\hline
 Task         & ERM                    & CVaR                   & Focal                  & Reweight               & US                     & Reweight (Envs)        & US (Envs)              & GDRO                   & IRM                    \\
\hline
 car          & \textbf{0.990 (0.000)} & 0.989 (0.000)          & 0.988 (0.000)          & 0.988 (0.000)          & 0.989 (0.001)          & 0.956 (0.001)          & 0.976 (0.001)          & 0.890 (0.011)          & 0.968 (0.004)          \\
 bowl         & \textbf{0.991 (0.000)} & \textbf{0.990 (0.001)} & 0.989 (0.000)          & \textbf{0.990 (0.000)} & 0.986 (0.001)          & 0.953 (0.003)          & 0.970 (0.002)          & 0.931 (0.023)          & 0.981 (0.001)          \\
 boat         & \textbf{0.999 (0.000)} & \textbf{0.999 (0.000)} & 0.999 (0.000)          & \textbf{0.999 (0.000)} & 0.998 (0.000)          & 0.996 (0.000)          & 0.998 (0.000)          & 0.995 (0.001)          & 0.997 (0.001)          \\
 fire-hydrant & 0.999 (0.000)          & \textbf{0.998 (0.001)} & 0.999 (0.000)          & 0.999 (0.000)          & 0.997 (0.000)          & 0.999 (0.000)          & 0.998 (0.000)          & \textbf{0.999 (0.001)} & \textbf{0.999 (0.000)} \\
 airplane     & 1.000 (0.000)          & \textbf{1.000 (0.000)} & 1.000 (0.000)          & \textbf{1.000 (0.000)} & \textbf{1.000 (0.000)} & 1.000 (0.000)          & \textbf{1.000 (0.000)} & \textbf{1.000 (0.000)} & 1.000 (0.000)          \\
 cow          & 1.000 (0.000)          & 0.998 (0.001)          & \textbf{1.000 (0.000)} & 1.000 (0.000)          & 0.999 (0.000)          & 0.999 (0.000)          & 0.998 (0.000)          & 0.998 (0.001)          & 0.999 (0.000)          \\
 backpack     & \textbf{0.972 (0.001)} & 0.952 (0.006)          & \textbf{0.972 (0.001)} & \textbf{0.972 (0.000)} & 0.961 (0.003)          & \textbf{0.973 (0.001)} & 0.966 (0.001)          & 0.958 (0.002)          & \textbf{0.968 (0.005)} \\
\hline
\end{tabular}
% \caption{Bar}
}
\caption{AUC on easy test examples for all 12 \rococo \ stress tests, after hyperparameter selection. 
Bold numbers have overlapping standard deviations.
L: \rococo-CE.
R: \rococo-Gist.
}
\label{tab:easy-auc-by-task-std}
\end{table}

\begin{table}
\resizebox{0.5\columnwidth}{!}{%
\centering
\begin{tabular}{llllllllll}
\hline
 Task         & ERM           & CVaR          & Focal                  & Reweight               & US            & Reweight (Envs)        & US (Envs)     & GDRO                   & IRM                    \\
\hline
 car          & 0.435 (0.012) & 0.476 (0.030) & 0.197 (0.008)          & 0.301 (0.003)          & 0.357 (0.017) & 0.253 (0.013)          & 0.365 (0.020) & \textbf{0.136 (0.011)} & 0.253 (0.041)          \\
 bowl         & 0.581 (0.032) & 0.627 (0.054) & \textbf{0.162 (0.007)} & 0.336 (0.014)          & 0.406 (0.021) & 0.314 (0.005)          & 0.369 (0.015) & 0.239 (0.007)          & 0.286 (0.023)          \\
 boat         & 0.353 (0.009) & 0.362 (0.028) & \textbf{0.120 (0.009)} & 0.183 (0.002)          & 0.307 (0.002) & 0.199 (0.002)          & 0.332 (0.022) & \textbf{0.145 (0.020)} & 0.225 (0.055)          \\
 fire-hydrant & 0.460 (0.011) & 0.406 (0.010) & \textbf{0.101 (0.011)} & 0.217 (0.004)          & 0.418 (0.020) & 0.383 (0.005)          & 0.397 (0.007) & 0.268 (0.026)          & 0.268 (0.070)          \\
 airplane     & 0.137 (0.008) & 0.125 (0.013) & \textbf{0.054 (0.000)} & \textbf{0.054 (0.003)} & 0.112 (0.005) & 0.103 (0.002)          & 0.114 (0.004) & \textbf{0.056 (0.007)} & 0.090 (0.013)          \\
 cow          & 0.357 (0.009) & 0.445 (0.027) & \textbf{0.115 (0.010)} & 0.192 (0.002)          & 0.340 (0.013) & 0.259 (0.005)          & 0.314 (0.019) & 0.187 (0.016)          & 0.249 (0.015)          \\
 backpack     & 0.822 (0.003) & 0.883 (0.039) & \textbf{0.239 (0.005)} & \textbf{0.237 (0.007)} & 0.588 (0.042) & 0.599 (0.014)          & 0.638 (0.011) & \textbf{0.238 (0.014)} & \textbf{0.271 (0.082)} \\
\hline
\end{tabular}
% \caption{Foo}
}
\hfill
\resizebox{0.5\columnwidth}{!}{%
\centering
\begin{tabular}{llllllllll}
\hline
 Task         & ERM           & CVaR          & Focal                  & Reweight               & US            & Reweight (Envs)        & US (Envs)     & GDRO                   & IRM                    \\
\hline
 car          & 0.459 (0.013) & 0.495 (0.029) & \textbf{0.207 (0.009)} & 0.323 (0.006)          & 0.380 (0.014) & 0.279 (0.015)          & 0.386 (0.024) & \textbf{0.251 (0.044)} & 0.278 (0.057)          \\
 bowl         & 0.616 (0.031) & 0.670 (0.059) & \textbf{0.170 (0.008)} & 0.370 (0.017)          & 0.443 (0.021) & 0.405 (0.022)          & 0.416 (0.016) & 0.300 (0.006)          & 0.276 (0.027)          \\
 boat         & 0.419 (0.015) & 0.458 (0.040) & \textbf{0.129 (0.009)} & 0.211 (0.000)          & 0.357 (0.006) & 0.283 (0.009)          & 0.400 (0.021) & 0.211 (0.040)          & 0.361 (0.135)          \\
 fire-hydrant & 0.612 (0.015) & 0.543 (0.002) & \textbf{0.150 (0.010)} & 0.307 (0.011)          & 0.550 (0.028) & 0.527 (0.002)          & 0.524 (0.009) & 0.399 (0.024)          & 0.376 (0.083)          \\
 airplane     & 0.200 (0.013) & 0.186 (0.021) & \textbf{0.087 (0.004)} & \textbf{0.087 (0.002)} & 0.177 (0.005) & 0.168 (0.004)          & 0.177 (0.002) & 0.098 (0.002)          & 0.116 (0.011)          \\
 cow          & 0.459 (0.011) & 0.526 (0.043) & \textbf{0.144 (0.018)} & 0.249 (0.006)          & 0.426 (0.009) & 0.361 (0.004)          & 0.431 (0.033) & 0.268 (0.022)          & 0.320 (0.033)          \\
 backpack     & 0.881 (0.008) & 0.919 (0.044) & \textbf{0.318 (0.002)} & \textbf{0.316 (0.006)} & 0.666 (0.035) & 0.686 (0.005)          & 0.715 (0.008) & 0.340 (0.006)          & \textbf{0.337 (0.073)} \\
\hline
\end{tabular}
% \caption{Bar}
}
\caption{Classification error on hard positive test examples for all 12 \rococo \ stress tests, after hyperparameter selection. 
Bold numbers have overlapping standard deviations.
L: \rococo-CE.
R: \rococo-Gist.
}
\label{tab:hard-pos-err-by-task-std}
\end{table}

\begin{table}
\resizebox{0.5\columnwidth}{!}{%
\centering
\begin{tabular}{llllllllll}
\hline
 Task         & ERM                    & CVaR                   & Focal         & Reweight      & US                     & Reweight (Envs)        & US (Envs)              & GDRO          & IRM                    \\
\hline
 car          & 0.118 (0.009)          & \textbf{0.084 (0.019)} & 0.554 (0.017) & 0.269 (0.006) & 0.195 (0.027)          & 0.132 (0.009)          & 0.122 (0.013)          & 0.176 (0.029) & 0.158 (0.053)          \\
 bowl         & \textbf{0.087 (0.016)} & \textbf{0.113 (0.043)} & 0.628 (0.024) & 0.274 (0.017) & 0.200 (0.024)          & 0.142 (0.010)          & 0.163 (0.021)          & 0.220 (0.030) & 0.191 (0.027)          \\
 boat         & \textbf{0.052 (0.004)} & \textbf{0.037 (0.022)} & 0.622 (0.023) & 0.266 (0.028) & 0.088 (0.008)          & \textbf{0.064 (0.006)} & \textbf{0.037 (0.009)} & 0.130 (0.044) & \textbf{0.113 (0.090)} \\
 fire-hydrant & \textbf{0.004 (0.001)} & \textbf{0.003 (0.001)} & 0.330 (0.006) & 0.119 (0.007) & 0.009 (0.003)          & 0.008 (0.001)          & 0.009 (0.003)          & 0.041 (0.003) & 0.117 (0.078)          \\
 airplane     & \textbf{0.002 (0.000)} & \textbf{0.002 (0.001)} & 0.039 (0.003) & 0.038 (0.004) & \textbf{0.003 (0.000)} & 0.004 (0.000)          & \textbf{0.003 (0.001)} & 0.012 (0.002) & 0.010 (0.002)          \\
 cow          & \textbf{0.011 (0.001)} & \textbf{0.010 (0.001)} & 0.175 (0.003) & 0.068 (0.002) & \textbf{0.010 (0.001)} & 0.014 (0.001)          & \textbf{0.010 (0.002)} & 0.051 (0.012) & 0.045 (0.007)          \\
 backpack     & \textbf{0.018 (0.002)} & \textbf{0.022 (0.019)} & 0.307 (0.004) & 0.274 (0.002) & 0.069 (0.012)          & 0.075 (0.005)          & 0.051 (0.005)          & 0.225 (0.009) & 0.266 (0.035)          \\
\hline
\end{tabular}
% \caption{Foo}
}
\hfill
\resizebox{0.5\columnwidth}{!}{%
\centering
\begin{tabular}{llllllllll}
\hline
 Task         & ERM                    & CVaR                   & Focal         & Reweight      & US            & Reweight (Envs)   & US (Envs)              & GDRO          & IRM           \\
\hline
 car          & 0.115 (0.008)          & \textbf{0.086 (0.019)} & 0.521 (0.016) & 0.262 (0.009) & 0.187 (0.029) & 0.156 (0.014)     & 0.125 (0.013)          & 0.188 (0.078) & 0.218 (0.031) \\
 bowl         & \textbf{0.141 (0.024)} & \textbf{0.147 (0.046)} & 0.785 (0.031) & 0.403 (0.014) & 0.305 (0.032) & 0.359 (0.021)     & 0.292 (0.036)          & 0.471 (0.039) & 0.561 (0.039) \\
 boat         & \textbf{0.067 (0.007)} & \textbf{0.045 (0.021)} & 0.672 (0.021) & 0.343 (0.026) & 0.114 (0.008) & 0.116 (0.006)     & \textbf{0.069 (0.004)} & 0.210 (0.045) & 0.186 (0.087) \\
 fire-hydrant & 0.013 (0.002)          & \textbf{0.004 (0.001)} & 0.427 (0.009) & 0.193 (0.008) & 0.024 (0.005) & 0.019 (0.002)     & 0.026 (0.007)          & 0.127 (0.026) & 0.212 (0.116) \\
 airplane     & \textbf{0.002 (0.000)} & \textbf{0.003 (0.001)} & 0.048 (0.004) & 0.047 (0.004) & 0.003 (0.000) & 0.006 (0.000)     & 0.003 (0.000)          & 0.029 (0.002) & 0.078 (0.025) \\
 cow          & 0.019 (0.000)          & \textbf{0.014 (0.001)} & 0.291 (0.005) & 0.118 (0.003) & 0.025 (0.001) & 0.034 (0.003)     & 0.021 (0.004)          & 0.105 (0.018) & 0.146 (0.037) \\
 backpack     & \textbf{0.027 (0.003)} & \textbf{0.032 (0.019)} & 0.387 (0.006) & 0.338 (0.004) & 0.101 (0.015) & 0.104 (0.006)     & 0.076 (0.005)          & 0.323 (0.015) & 0.332 (0.032) \\
\hline
\end{tabular}
% \caption{Bar}
}
\caption{Classification error on hard negative test examples for all 12 \rococo \ stress tests, after hyperparameter selection. 
Bold numbers have overlapping standard deviations.
L: \rococo-CE.
R: \rococo-Gist.
}
\label{tab:hard-neg-err-by-task-std}
\end{table}

\begin{table}
\resizebox{0.5\columnwidth}{!}{%
\centering
\begin{tabular}{llllllllll}
\hline
 Task         & ERM                    & CVaR                   & Focal         & Reweight      & US                     & Reweight (Envs)        & US (Envs)              & GDRO          & IRM           \\
\hline
 car          & 0.020 (0.001)          & \textbf{0.019 (0.001)} & 0.066 (0.002) & 0.031 (0.001) & 0.026 (0.002)          & 0.071 (0.004)          & 0.030 (0.002)          & 0.465 (0.104) & 0.068 (0.013) \\
 bowl         & \textbf{0.018 (0.001)} & \textbf{0.019 (0.001)} & 0.087 (0.006) & 0.034 (0.001) & 0.028 (0.001)          & 0.070 (0.001)          & 0.036 (0.002)          & 0.107 (0.020) & 0.077 (0.005) \\
 boat         & 0.004 (0.000)          & \textbf{0.004 (0.000)} & 0.039 (0.002) & 0.016 (0.001) & 0.006 (0.000)          & 0.058 (0.002)          & 0.006 (0.001)          & 0.089 (0.012) & 0.092 (0.059) \\
 fire-hydrant & 0.003 (0.000)          & \textbf{0.002 (0.000)} & 0.058 (0.002) & 0.020 (0.001) & 0.003 (0.000)          & 0.005 (0.000)          & 0.004 (0.001)          & 0.027 (0.001) & 0.040 (0.029) \\
 airplane     & \textbf{0.001 (0.000)} & \textbf{0.001 (0.000)} & 0.008 (0.000) & 0.007 (0.001) & \textbf{0.001 (0.000)} & 0.002 (0.000)          & \textbf{0.001 (0.000)} & 0.031 (0.009) & 0.003 (0.001) \\
 cow          & 0.003 (0.000)          & \textbf{0.003 (0.000)} & 0.023 (0.000) & 0.009 (0.000) & 0.004 (0.000)          & 0.006 (0.000)          & 0.004 (0.000)          & 0.026 (0.003) & 0.009 (0.001) \\
 backpack     & \textbf{0.028 (0.000)} & 0.038 (0.005)          & 0.124 (0.002) & 0.109 (0.002) & 0.045 (0.003)          & 0.038 (0.002)          & 0.037 (0.000)          & 0.135 (0.013) & 0.146 (0.024) \\
\hline
\end{tabular}
% \caption{Foo}
}
\hfill
\resizebox{0.5\columnwidth}{!}{%
\centering
\begin{tabular}{llllllllll}
\hline
 Task         & ERM                    & CVaR                   & Focal                  & Reweight               & US                     & Reweight (Envs)        & US (Envs)              & GDRO          & IRM           \\
\hline
 car          & \textbf{0.019 (0.001)} & \textbf{0.017 (0.001)} & 0.067 (0.003)          & 0.030 (0.000)          & 0.025 (0.001)          & 0.068 (0.004)          & 0.029 (0.002)          & 0.131 (0.025) & 0.109 (0.034) \\
 bowl         & \textbf{0.014 (0.000)} & 0.016 (0.001)          & 0.080 (0.006)          & 0.026 (0.002)          & 0.022 (0.001)          & 0.041 (0.003)          & 0.028 (0.002)          & 0.092 (0.019) & 0.238 (0.040) \\
 boat         & 0.003 (0.000)          & \textbf{0.002 (0.000)} & 0.037 (0.001)          & 0.013 (0.001)          & 0.005 (0.000)          & 0.055 (0.002)          & 0.003 (0.000)          & 0.085 (0.012) & 0.194 (0.109) \\
 fire-hydrant & \textbf{0.001 (0.000)} & \textbf{0.001 (0.000)} & 0.051 (0.002)          & 0.014 (0.001)          & 0.001 (0.000)          & 0.003 (0.000)          & 0.002 (0.000)          & 0.020 (0.002) & 0.033 (0.027) \\
 airplane     & 0.001 (0.000)          & \textbf{0.000 (0.000)} & 0.006 (0.000)          & 0.005 (0.001)          & \textbf{0.000 (0.000)} & 0.001 (0.000)          & \textbf{0.001 (0.000)} & 0.008 (0.001) & 0.013 (0.004) \\
 cow          & \textbf{0.002 (0.000)} & 0.002 (0.000)          & 0.010 (0.000)          & 0.004 (0.000)          & \textbf{0.002 (0.000)} & 0.003 (0.000)          & 0.002 (0.000)          & 0.020 (0.003) & 0.014 (0.003) \\
 backpack     & \textbf{0.023 (0.000)} & 0.034 (0.005)          & 0.099 (0.003)          & 0.083 (0.001)          & 0.031 (0.001)          & 0.025 (0.000)          & 0.026 (0.001)          & 0.095 (0.004) & 0.120 (0.021) \\
\hline
\end{tabular}
% \caption{Bar}
}
\caption{Classification error on easy test examples for all 12 \rococo \ stress tests, after hyperparameter selection. 
Bold numbers have overlapping standard deviations.
L: \rococo-CE.
R: \rococo-Gist.
}
\label{tab:easy-err-by-task-std}
\end{table}

\begin{table}
\resizebox{0.5\columnwidth}{!}{%
\centering
\begin{tabular}{llllllllll}
\hline
 Task         & ERM           & CVaR          & Focal                  & Reweight               & US            & Reweight (Envs)        & US (Envs)     & GDRO                   & IRM                    \\
\hline
 car          & 1.072 (0.041) & 0.669 (0.064) & 0.419 (0.009)          & 0.673 (0.009)          & 0.899 (0.113) & 0.462 (0.016)          & 0.862 (0.038) & \textbf{0.309 (0.012)} & 0.461 (0.068)          \\
 bowl         & 1.385 (0.088) & 0.702 (0.005) & \textbf{0.415 (0.007)} & 0.720 (0.032)          & 0.940 (0.138) & 0.580 (0.015)          & 0.827 (0.120) & 0.480 (0.011)          & 0.529 (0.030)          \\
 boat         & 1.171 (0.033) & 0.817 (0.049) & \textbf{0.297 (0.006)} & 0.454 (0.012)          & 0.964 (0.019) & 0.395 (0.008)          & 1.038 (0.065) & \textbf{0.285 (0.021)} & 0.413 (0.060)          \\
 fire-hydrant & 1.639 (0.040) & 0.996 (0.069) & \textbf{0.316 (0.006)} & 0.429 (0.005)          & 1.488 (0.083) & 1.090 (0.010)          & 1.453 (0.218) & 0.472 (0.020)          & 0.437 (0.076)          \\
 airplane     & 0.485 (0.025) & 0.420 (0.069) & \textbf{0.130 (0.002)} & 0.144 (0.002)          & 0.469 (0.013) & 0.318 (0.003)          & 0.463 (0.033) & \textbf{0.150 (0.032)} & 0.272 (0.050)          \\
 cow          & 1.272 (0.037) & 1.102 (0.064) & \textbf{0.290 (0.006)} & 0.497 (0.008)          & 1.219 (0.108) & 0.793 (0.014)          & 1.154 (0.096) & 0.410 (0.036)          & 0.584 (0.027)          \\
 backpack     & 2.035 (0.049) & 0.719 (0.003) & \textbf{0.506 (0.006)} & \textbf{0.492 (0.010)} & 1.519 (0.284) & 1.225 (0.043)          & 1.453 (0.124) & \textbf{0.511 (0.021)} & \textbf{0.523 (0.094)} \\
\hline
\end{tabular}
% \caption{Foo}
}
\hfill
\resizebox{0.5\columnwidth}{!}{%
\centering
\begin{tabular}{llllllllll}
\hline
 Task         & ERM           & CVaR          & Focal                  & Reweight               & US            & Reweight (Envs)        & US (Envs)     & GDRO                   & IRM                    \\
\hline
 car          & 1.126 (0.045) & 0.690 (0.068) & \textbf{0.431 (0.009)} & 0.708 (0.011)          & 0.941 (0.116) & 0.511 (0.017)          & 0.916 (0.040) & \textbf{0.456 (0.039)} & \textbf{0.487 (0.083)} \\
 bowl         & 1.486 (0.089) & 0.705 (0.005) & \textbf{0.435 (0.007)} & 0.769 (0.035)          & 1.002 (0.147) & 0.727 (0.036)          & 0.927 (0.136) & 0.533 (0.010)          & 0.675 (0.010)          \\
 boat         & 1.386 (0.043) & 1.004 (0.070) & \textbf{0.343 (0.004)} & 0.517 (0.019)          & 1.134 (0.025) & 0.608 (0.025)          & 1.274 (0.082) & 0.416 (0.049)          & 0.616 (0.133)          \\
 fire-hydrant & 2.300 (0.060) & 1.382 (0.085) & \textbf{0.404 (0.005)} & 0.646 (0.008)          & 2.066 (0.101) & 1.595 (0.004)          & 2.035 (0.288) & 0.707 (0.034)          & 0.617 (0.100)          \\
 airplane     & 0.771 (0.038) & 0.628 (0.079) & \textbf{0.190 (0.004)} & 0.210 (0.003)          & 0.763 (0.023) & 0.532 (0.006)          & 0.740 (0.046) & 0.247 (0.016)          & 0.269 (0.035)          \\
 cow          & 1.634 (0.037) & 1.339 (0.093) & \textbf{0.348 (0.009)} & 0.623 (0.011)          & 1.559 (0.062) & 1.137 (0.021)          & 1.517 (0.135) & 0.576 (0.071)          & 0.669 (0.034)          \\
 backpack     & 2.363 (0.060) & 0.725 (0.005) & \textbf{0.594 (0.003)} & 0.631 (0.009)          & 1.809 (0.313) & 1.458 (0.044)          & 1.740 (0.159) & 0.635 (0.006)          & \textbf{0.603 (0.077)} \\
\hline
\end{tabular}
% \caption{Bar}
}
\caption{NLL on hard positive test examples for all 12 \rococo \ stress tests, after hyperparameter selection. 
Bold numbers have overlapping standard deviations.
L: \rococo-CE.
R: \rococo-Gist.
}
\label{tab:hard-pos-NLL-by-task-std}
\end{table}

\begin{table}
\resizebox{0.5\columnwidth}{!}{%
\centering
\begin{tabular}{llllllllll}
\hline
 Task         & ERM                    & CVaR                   & Focal         & Reweight      & US                     & Reweight (Envs)   & US (Envs)              & GDRO          & IRM           \\
\hline
 car          & \textbf{0.330 (0.019)} & 0.444 (0.061)          & 0.819 (0.008) & 0.643 (0.014) & 0.491 (0.050)          & 0.397 (0.016)     & \textbf{0.341 (0.029)} & 0.487 (0.038) & 0.471 (0.106) \\
 bowl         & \textbf{0.241 (0.033)} & 0.665 (0.005)          & 0.890 (0.024) & 0.629 (0.031) & 0.491 (0.030)          & 0.377 (0.015)     & 0.415 (0.031)          & 0.550 (0.039) & 0.510 (0.033) \\
 boat         & 0.165 (0.007)          & \textbf{0.158 (0.056)} & 0.903 (0.017) & 0.682 (0.044) & 0.246 (0.016)          & 0.169 (0.013)     & \textbf{0.123 (0.020)} & 0.340 (0.089) & 0.409 (0.117) \\
 fire-hydrant & \textbf{0.023 (0.003)} & 0.050 (0.013)          & 0.637 (0.007) & 0.290 (0.010) & 0.034 (0.006)          & 0.040 (0.002)     & \textbf{0.034 (0.010)} & 0.178 (0.023) & 0.343 (0.108) \\
 airplane     & \textbf{0.009 (0.001)} & \textbf{0.013 (0.003)} & 0.165 (0.011) & 0.138 (0.012) & 0.011 (0.000)          & 0.017 (0.001)     & \textbf{0.010 (0.001)} & 0.047 (0.012) & 0.038 (0.008) \\
 cow          & \textbf{0.034 (0.001)} & 0.054 (0.011)          & 0.453 (0.005) & 0.201 (0.005) & \textbf{0.034 (0.002)} & 0.042 (0.002)     & \textbf{0.036 (0.007)} & 0.156 (0.033) & 0.131 (0.013) \\
 backpack     & \textbf{0.077 (0.005)} & 0.644 (0.007)          & 0.597 (0.004) & 0.617 (0.008) & 0.187 (0.029)          & 0.206 (0.010)     & 0.153 (0.018)          & 0.457 (0.005) & 0.544 (0.045) \\
\hline
\end{tabular}
% \caption{Foo}
}
\hfill
\resizebox{0.5\columnwidth}{!}{%
\centering
\begin{tabular}{llllllllll}
\hline
 Task         & ERM                    & CVaR                   & Focal         & Reweight      & US            & Reweight (Envs)   & US (Envs)              & GDRO          & IRM           \\
\hline
 car          & \textbf{0.321 (0.016)} & 0.439 (0.060)          & 0.784 (0.007) & 0.619 (0.021) & 0.472 (0.054) & 0.475 (0.018)     & \textbf{0.360 (0.031)} & 0.539 (0.075) & 0.575 (0.049) \\
 bowl         & \textbf{0.361 (0.050)} & 0.671 (0.004)          & 1.026 (0.026) & 0.876 (0.033) & 0.701 (0.036) & 0.819 (0.043)     & 0.722 (0.036)          & 0.786 (0.016) & 0.701 (0.005) \\
 boat         & \textbf{0.200 (0.010)} & \textbf{0.182 (0.056)} & 0.971 (0.015) & 0.838 (0.051) & 0.310 (0.013) & 0.299 (0.017)     & \textbf{0.197 (0.016)} & 0.499 (0.107) & 0.468 (0.097) \\
 fire-hydrant & \textbf{0.044 (0.004)} & 0.064 (0.011)          & 0.721 (0.009) & 0.438 (0.012) & 0.065 (0.011) & 0.092 (0.006)     & 0.075 (0.019)          & 0.369 (0.038) & 0.465 (0.121) \\
 airplane     & \textbf{0.011 (0.001)} & \textbf{0.016 (0.004)} & 0.177 (0.011) & 0.153 (0.013) & 0.013 (0.001) & 0.023 (0.001)     & \textbf{0.012 (0.001)} & 0.098 (0.011) & 0.230 (0.039) \\
 cow          & \textbf{0.056 (0.001)} & \textbf{0.064 (0.009)} & 0.618 (0.009) & 0.331 (0.005) & 0.074 (0.005) & 0.102 (0.004)     & \textbf{0.066 (0.014)} & 0.309 (0.046) & 0.349 (0.082) \\
 backpack     & \textbf{0.103 (0.005)} & 0.655 (0.004)          & 0.663 (0.005) & 0.754 (0.007) & 0.259 (0.034) & 0.261 (0.013)     & 0.207 (0.014)          & 0.602 (0.019) & 0.628 (0.048) \\
\hline
\end{tabular}
% \caption{Bar}
}
\caption{NLL on hard negative test examples for all 12 \rococo \ stress tests, after hyperparameter selection. 
Bold numbers have overlapping standard deviations.
L: \rococo-CE.
R: \rococo-Gist.
}
\label{tab:hard-neg-NLL-by-task-std}
\end{table}

\begin{table}
\resizebox{0.5\columnwidth}{!}{%
\centering
\begin{tabular}{llllllllll}
\hline
 Task         & ERM                    & CVaR                   & Focal         & Reweight      & US                     & Reweight (Envs)   & US (Envs)              & GDRO          & IRM           \\
\hline
 car          & \textbf{0.059 (0.002)} & 0.286 (0.057)          & 0.305 (0.007) & 0.095 (0.001) & 0.075 (0.008)          & 0.249 (0.009)     & 0.092 (0.004)          & 0.782 (0.084) & 0.222 (0.028) \\
 bowl         & \textbf{0.051 (0.002)} & 0.600 (0.003)          & 0.320 (0.011) & 0.096 (0.004) & 0.079 (0.005)          & 0.224 (0.003)     & 0.108 (0.008)          & 0.352 (0.067) & 0.223 (0.011) \\
 boat         & \textbf{0.015 (0.000)} & 0.033 (0.003)          & 0.293 (0.006) & 0.057 (0.004) & 0.020 (0.001)          & 0.255 (0.006)     & 0.021 (0.002)          & 0.295 (0.009) & 0.306 (0.098) \\
 fire-hydrant & \textbf{0.012 (0.000)} & 0.023 (0.006)          & 0.289 (0.005) & 0.059 (0.001) & 0.013 (0.001)          & 0.028 (0.001)     & 0.016 (0.002)          & 0.140 (0.021) & 0.161 (0.062) \\
 airplane     & \textbf{0.006 (0.000)} & \textbf{0.007 (0.001)} & 0.056 (0.004) & 0.035 (0.003) & \textbf{0.006 (0.000)} & 0.010 (0.001)     & \textbf{0.006 (0.000)} & 0.101 (0.025) & 0.013 (0.003) \\
 cow          & \textbf{0.011 (0.000)} & 0.022 (0.005)          & 0.192 (0.003) & 0.035 (0.000) & \textbf{0.012 (0.001)} & 0.025 (0.000)     & 0.013 (0.001)          & 0.118 (0.020) & 0.038 (0.002) \\
 backpack     & \textbf{0.086 (0.001)} & 0.636 (0.007)          & 0.380 (0.002) & 0.281 (0.003) & 0.123 (0.003)          & 0.113 (0.005)     & 0.109 (0.004)          & 0.363 (0.032) & 0.359 (0.026) \\
\hline
\end{tabular}
% \caption{Foo}
}
\hfill
\resizebox{0.5\columnwidth}{!}{%
\centering
\begin{tabular}{llllllllll}
\hline
 Task         & ERM                    & CVaR                   & Focal         & Reweight      & US                     & Reweight (Envs)   & US (Envs)              & GDRO          & IRM           \\
\hline
 car          & \textbf{0.056 (0.002)} & 0.285 (0.057)          & 0.306 (0.007) & 0.093 (0.001) & 0.073 (0.008)          & 0.242 (0.009)     & 0.087 (0.004)          & 0.408 (0.074) & 0.313 (0.074) \\
 bowl         & \textbf{0.041 (0.001)} & 0.599 (0.003)          & 0.312 (0.011) & 0.083 (0.004) & 0.066 (0.005)          & 0.144 (0.008)     & 0.089 (0.009)          & 0.338 (0.068) & 0.495 (0.108) \\
 boat         & \textbf{0.010 (0.000)} & 0.029 (0.003)          & 0.290 (0.006) & 0.051 (0.004) & 0.016 (0.001)          & 0.248 (0.006)     & 0.012 (0.001)          & 0.288 (0.009) & 0.453 (0.143) \\
 fire-hydrant & \textbf{0.005 (0.000)} & 0.019 (0.006)          & 0.282 (0.005) & 0.047 (0.001) & 0.006 (0.001)          & 0.020 (0.001)     & 0.008 (0.002)          & 0.125 (0.021) & 0.151 (0.062) \\
 airplane     & \textbf{0.002 (0.000)} & 0.004 (0.001)          & 0.053 (0.004) & 0.032 (0.003) & \textbf{0.002 (0.000)} & 0.006 (0.001)     & \textbf{0.002 (0.000)} & 0.053 (0.008) & 0.067 (0.017) \\
 cow          & \textbf{0.005 (0.000)} & 0.019 (0.005)          & 0.174 (0.004) & 0.020 (0.001) & 0.006 (0.000)          & 0.016 (0.000)     & 0.007 (0.001)          & 0.101 (0.022) & 0.073 (0.026) \\
 backpack     & \textbf{0.065 (0.001)} & 0.632 (0.008)          & 0.398 (0.004) & 0.226 (0.003) & 0.087 (0.001)          & 0.085 (0.003)     & 0.079 (0.006)          & 0.306 (0.024) & 0.326 (0.020) \\
\hline
\end{tabular}
% \caption{Bar}
}
\caption{NLL on easy test examples for all 12 \rococo \ stress tests, after hyperparameter selection. 
Bold numbers have overlapping standard deviations.
L: \rococo-CE.
R: \rococo-Gist.
}
\label{tab:easy-NLL-by-task-std}
\end{table}

\begin{table}
\resizebox{0.5\columnwidth}{!}{%
\centering
\begin{tabular}{llllllllll}
\hline
 Task         & ERM                    & CVaR                   & Focal                  & Reweight               & US                     & Reweight (Envs)        & US (Envs)              & GDRO                   & IRM                    \\
\hline
 car          & 0.017 (0.001)          & \textbf{0.005 (0.002)} & \textbf{0.004 (0.000)} & 0.012 (0.000)          & 0.015 (0.002)          & \textbf{0.004 (0.000)} & 0.014 (0.001)          & 0.008 (0.001)          & \textbf{0.004 (0.001)} \\
 bowl         & 0.013 (0.001)          & 0.011 (0.000)          & 0.007 (0.001)          & 0.009 (0.000)          & 0.010 (0.002)          & \textbf{0.004 (0.000)} & 0.009 (0.002)          & 0.008 (0.002)          & \textbf{0.005 (0.001)} \\
 boat         & 0.004 (0.000)          & 0.002 (0.000)          & 0.007 (0.001)          & 0.004 (0.000)          & 0.004 (0.000)          & \textbf{0.001 (0.000)} & 0.004 (0.000)          & 0.002 (0.000)          & 0.005 (0.001)          \\
 fire-hydrant & 0.002 (0.000)          & \textbf{0.001 (0.000)} & 0.003 (0.000)          & \textbf{0.001 (0.000)} & 0.002 (0.000)          & \textbf{0.001 (0.000)} & 0.002 (0.000)          & 0.004 (0.001)          & 0.007 (0.002)          \\
 airplane     & 0.001 (0.000)          & \textbf{0.000 (0.000)} & 0.010 (0.001)          & 0.005 (0.001)          & 0.001 (0.000)          & \textbf{0.000 (0.000)} & 0.001 (0.000)          & 0.001 (0.001)          & 0.001 (0.000)          \\
 cow          & 0.002 (0.000)          & \textbf{0.002 (0.001)} & 0.009 (0.000)          & 0.001 (0.000)          & 0.002 (0.000)          & \textbf{0.001 (0.000)} & 0.002 (0.000)          & \textbf{0.001 (0.001)} & \textbf{0.001 (0.000)} \\
 backpack     & 0.006 (0.001)          & 0.108 (0.003)          & 0.009 (0.001)          & 0.022 (0.001)          & 0.008 (0.003)          & \textbf{0.002 (0.001)} & 0.004 (0.000)          & 0.007 (0.003)          & 0.011 (0.003)          \\
\hline
\end{tabular}
% \caption{Foo}
}
\hfill
\resizebox{0.5\columnwidth}{!}{%
\centering
\begin{tabular}{llllllllll}
\hline
 Task         & ERM                    & CVaR                   & Focal                  & Reweight      & US                     & Reweight (Envs)        & US (Envs)              & GDRO                   & IRM                    \\
\hline
 car          & 0.019 (0.001)          & \textbf{0.006 (0.002)} & \textbf{0.003 (0.000)} & 0.013 (0.001) & 0.016 (0.002)          & 0.005 (0.000)          & 0.015 (0.001)          & 0.006 (0.003)          & \textbf{0.003 (0.001)} \\
 bowl         & 0.015 (0.001)          & 0.008 (0.001)          & 0.014 (0.001)          & 0.014 (0.000) & 0.014 (0.001)          & 0.013 (0.000)          & 0.014 (0.002)          & \textbf{0.005 (0.001)} & \textbf{0.004 (0.001)} \\
 boat         & 0.005 (0.000)          & \textbf{0.003 (0.000)} & 0.009 (0.001)          & 0.006 (0.000) & 0.005 (0.000)          & \textbf{0.002 (0.000)} & 0.004 (0.000)          & \textbf{0.002 (0.001)} & \textbf{0.002 (0.000)} \\
 fire-hydrant & 0.003 (0.000)          & \textbf{0.001 (0.000)} & 0.004 (0.001)          & 0.004 (0.001) & 0.003 (0.000)          & 0.002 (0.000)          & 0.003 (0.000)          & 0.005 (0.002)          & 0.006 (0.002)          \\
 airplane     & 0.001 (0.000)          & 0.001 (0.000)          & 0.009 (0.001)          & 0.004 (0.001) & 0.001 (0.000)          & \textbf{0.000 (0.000)} & 0.001 (0.000)          & 0.003 (0.001)          & 0.008 (0.004)          \\
 cow          & 0.002 (0.000)          & \textbf{0.002 (0.000)} & 0.006 (0.000)          & 0.004 (0.000) & 0.003 (0.000)          & 0.002 (0.000)          & 0.003 (0.000)          & \textbf{0.003 (0.002)} & 0.003 (0.001)          \\
 backpack     & 0.008 (0.000)          & 0.106 (0.003)          & \textbf{0.005 (0.001)} & 0.038 (0.001) & 0.014 (0.002)          & \textbf{0.004 (0.000)} & 0.007 (0.002)          & 0.011 (0.003)          & \textbf{0.012 (0.009)} \\
\hline
\end{tabular}
% \caption{Bar}
}
\caption{ECE on hard test examples for all 12 \rococo \ stress tests, after hyperparameter selection. 
Bold numbers have overlapping standard deviations.
L: \rococo-CE.
R: \rococo-Gist.
}
\label{tab:hard-ece-by-task-std}
\end{table}

\begin{table}
\resizebox{0.5\columnwidth}{!}{%
\centering
\begin{tabular}{llllllllll}
\hline
 Task         & ERM                    & CVaR          & Focal         & Reweight               & US                     & Reweight (Envs)   & US (Envs)              & GDRO          & IRM           \\
\hline
 car          & \textbf{0.008 (0.001)} & 0.198 (0.039) & 0.147 (0.005) & 0.016 (0.002)          & \textbf{0.009 (0.006)} & 0.069 (0.004)     & \textbf{0.011 (0.008)} & 0.116 (0.082) & 0.054 (0.013) \\
 bowl         & \textbf{0.007 (0.001)} & 0.395 (0.001) & 0.130 (0.002) & 0.012 (0.001)          & \textbf{0.009 (0.005)} & 0.047 (0.003)     & \textbf{0.012 (0.008)} & 0.125 (0.029) & 0.057 (0.016) \\
 boat         & \textbf{0.002 (0.000)} & 0.022 (0.003) & 0.185 (0.003) & 0.014 (0.002)          & 0.003 (0.000)          & 0.119 (0.002)     & 0.004 (0.001)          & 0.093 (0.028) & 0.119 (0.007) \\
 fire-hydrant & \textbf{0.001 (0.000)} & 0.016 (0.005) & 0.151 (0.003) & 0.010 (0.001)          & 0.002 (0.000)          & 0.011 (0.000)     & \textbf{0.003 (0.002)} & 0.069 (0.015) & 0.069 (0.009) \\
 airplane     & \textbf{0.000 (0.000)} & 0.002 (0.000) & 0.032 (0.002) & 0.015 (0.002)          & \textbf{0.000 (0.000)} & 0.004 (0.000)     & \textbf{0.000 (0.000)} & 0.020 (0.004) & 0.004 (0.001) \\
 cow          & \textbf{0.000 (0.000)} & 0.013 (0.005) & 0.119 (0.003) & 0.007 (0.000)          & 0.001 (0.000)          & 0.007 (0.000)     & \textbf{0.001 (0.000)} & 0.044 (0.016) & 0.013 (0.001) \\
 backpack     & \textbf{0.003 (0.000)} & 0.308 (0.001) & 0.098 (0.002) & 0.010 (0.002)          & \textbf{0.005 (0.002)} & 0.016 (0.002)     & \textbf{0.008 (0.007)} & 0.065 (0.009) & 0.042 (0.026) \\
\hline
\end{tabular}
% \caption{Foo}
}
\hfill
\resizebox{0.5\columnwidth}{!}{%
\centering
\begin{tabular}{llllllllll}
\hline
 Task         & ERM                    & CVaR          & Focal         & Reweight      & US                     & Reweight (Envs)   & US (Envs)              & GDRO          & IRM           \\
\hline
 car          & \textbf{0.010 (0.001)} & 0.199 (0.039) & 0.147 (0.005) & 0.017 (0.002) & \textbf{0.010 (0.006)} & 0.070 (0.004)     & \textbf{0.012 (0.008)} & 0.125 (0.053) & 0.076 (0.026) \\
 bowl         & \textbf{0.009 (0.001)} & 0.398 (0.001) & 0.137 (0.002) & 0.017 (0.001) & \textbf{0.012 (0.005)} & 0.044 (0.004)     & \textbf{0.015 (0.011)} & 0.137 (0.030) & 0.113 (0.086) \\
 boat         & \textbf{0.003 (0.000)} & 0.023 (0.003) & 0.187 (0.003) & 0.017 (0.002) & 0.004 (0.000)          & 0.120 (0.002)     & 0.004 (0.001)          & 0.095 (0.028) & 0.079 (0.054) \\
 fire-hydrant & \textbf{0.002 (0.000)} & 0.016 (0.006) & 0.157 (0.004) & 0.014 (0.001) & \textbf{0.003 (0.000)} & 0.011 (0.000)     & \textbf{0.004 (0.001)} & 0.069 (0.015) & 0.072 (0.010) \\
 airplane     & \textbf{0.001 (0.000)} & 0.002 (0.000) & 0.034 (0.002) & 0.016 (0.002) & \textbf{0.001 (0.000)} & 0.004 (0.000)     & \textbf{0.001 (0.000)} & 0.028 (0.005) & 0.032 (0.014) \\
 cow          & \textbf{0.001 (0.000)} & 0.013 (0.005) & 0.126 (0.003) & 0.009 (0.000) & \textbf{0.001 (0.000)} & 0.008 (0.000)     & \textbf{0.001 (0.000)} & 0.046 (0.015) & 0.037 (0.018) \\
 backpack     & \textbf{0.007 (0.000)} & 0.310 (0.002) & 0.142 (0.002) & 0.026 (0.001) & \textbf{0.006 (0.003)} & 0.021 (0.002)     & \textbf{0.012 (0.007)} & 0.079 (0.013) & 0.059 (0.022) \\
\hline
\end{tabular}
% \caption{Bar}
}
\caption{ECE on easy test examples for all 12 \rococo \ stress tests, after hyperparameter selection. 
Bold numbers have overlapping standard deviations.
L: \rococo-CE.
R: \rococo-Gist.
}
\label{tab:easy-ece-by-task-std}
\end{table}

\clearpage
\section{Qualitative Analysis} \label{app:qual}

\subsection{Examples}

\begin{figure}[t!] % "[t!]" placement specifier just for this example
\begin{subfigure}{0.33\textwidth}
\includegraphics[width=\linewidth]{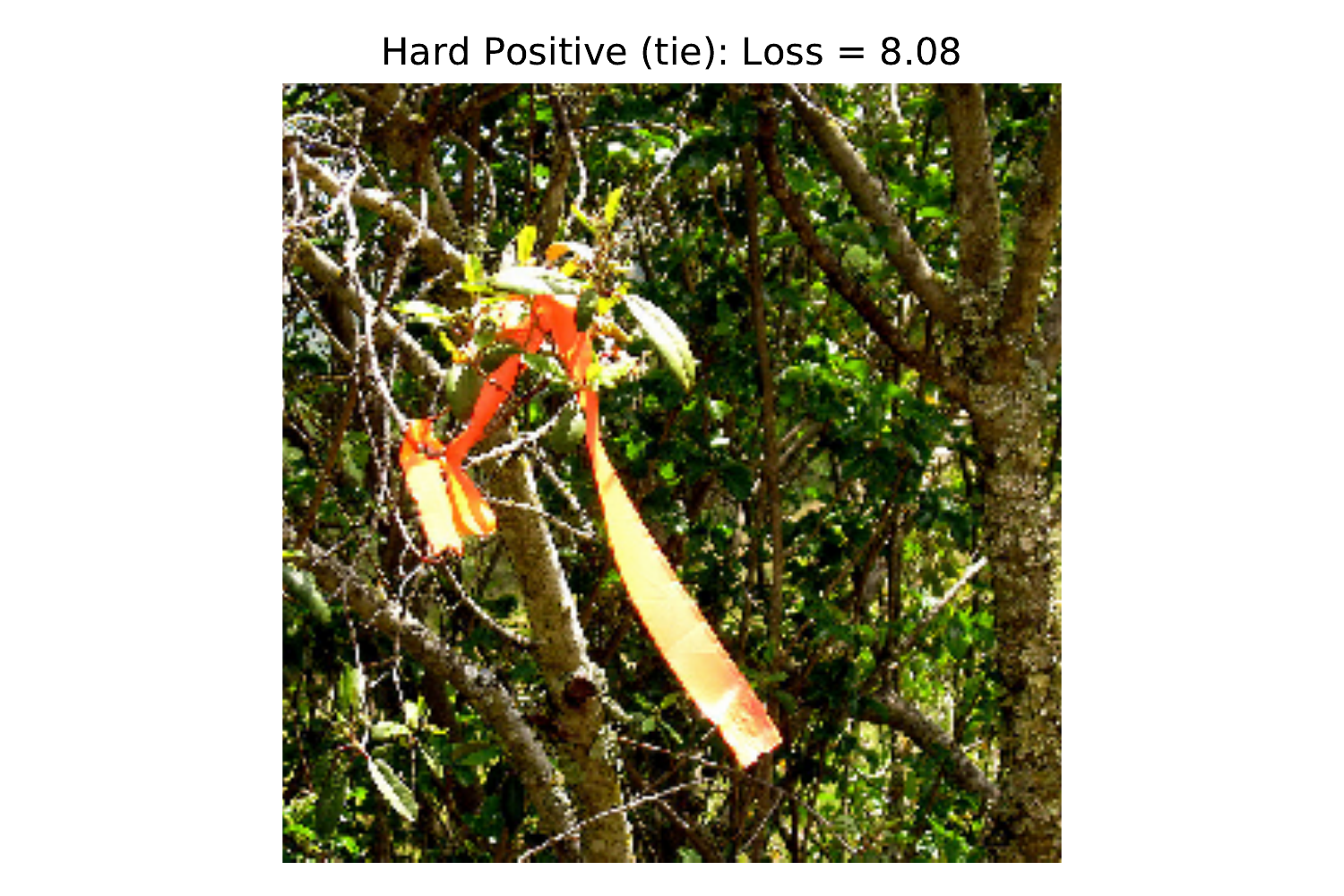}
\end{subfigure}\hspace*{\fill}
\begin{subfigure}{0.33\textwidth}
\includegraphics[width=\linewidth]{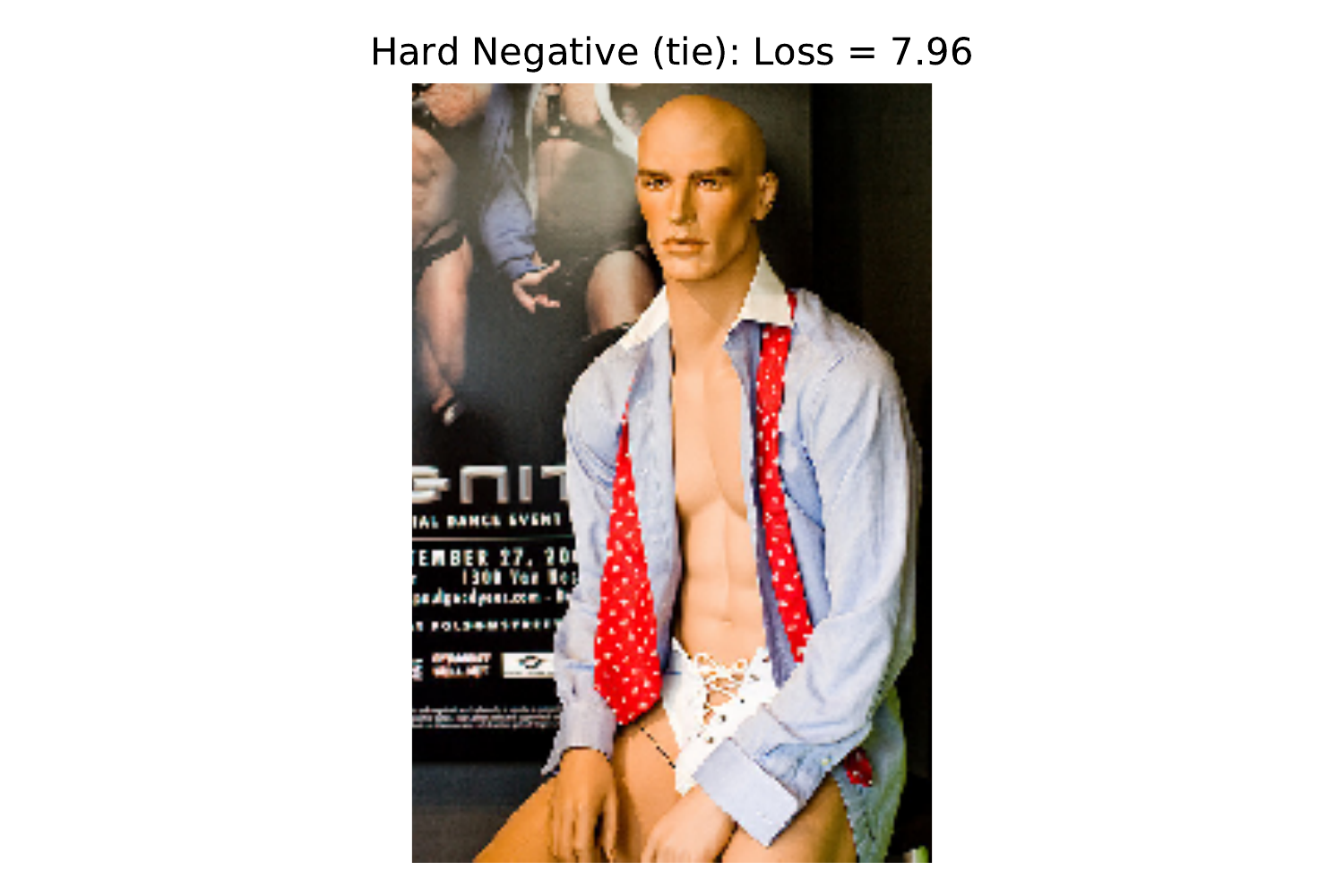}
\end{subfigure}\hspace*{\fill}
\begin{subfigure}{0.33\textwidth}
\includegraphics[width=\linewidth]{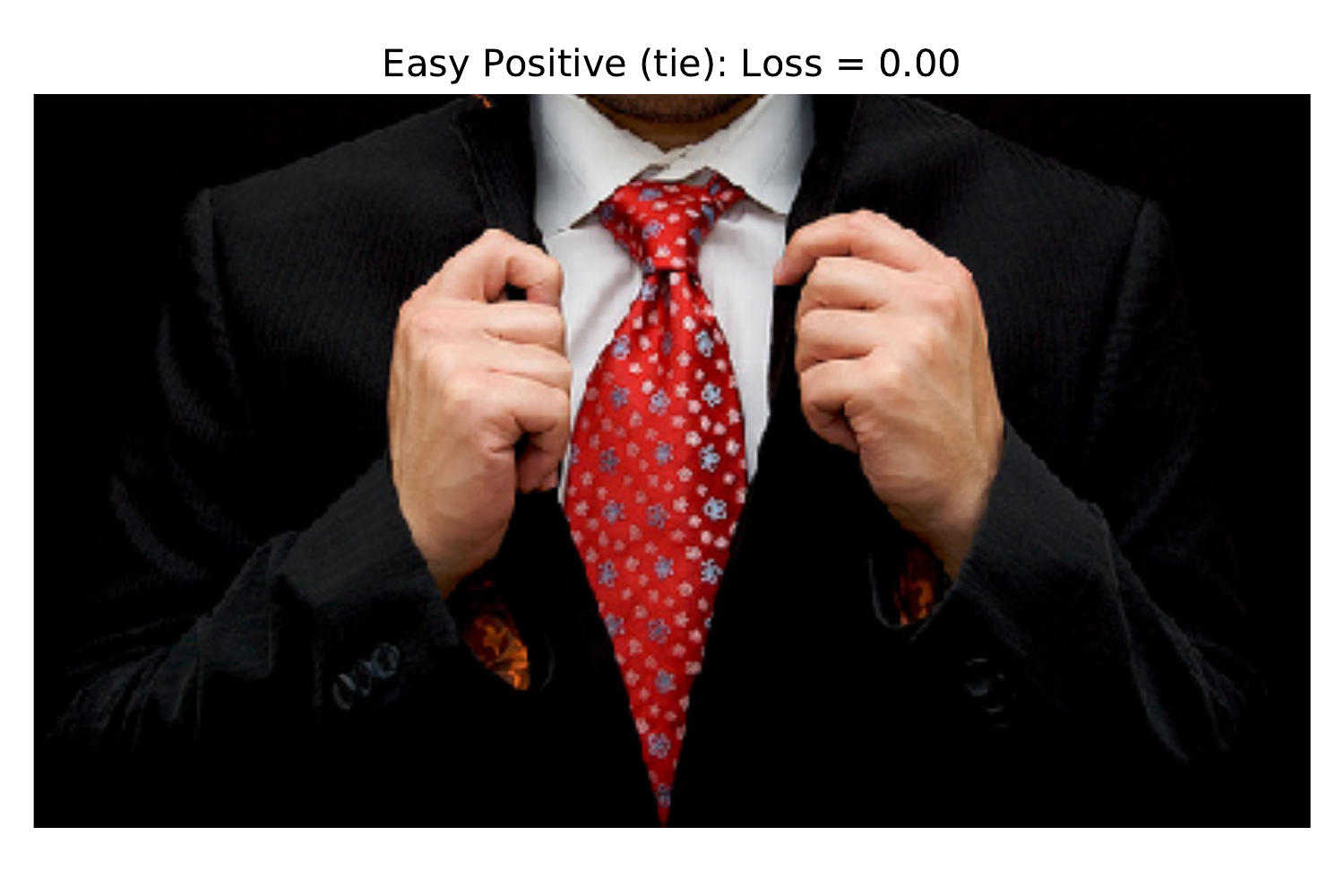}
\end{subfigure}
\caption{Examples from the \texttt{tie} task.
} 
\label{app:examples-qual-tie}
\end{figure}

Here we show examples from \rococo-CE: from left to right we show a hard positive, hard negative, and easy positive for each of the 12 tasks.
We show the loss for a sample ERM model in the title of the caption.
We choose these examples by finding the highest (lowest) loss ERM examples for each hard (easy) category --- additionally, for ease of viewing, for hard positives, we choose examples for which the target object's area is larger than average for positive examples of that class.
We note that label noise can be a problem in COCO, as in any dataset, and finding the highest loss examples can occasionally surface a mislabelled datapoint (see Fig. \ref{app:examples-qual-tie}, the Hard Negative in the \texttt{tie} class, where a tie is present but is not labelled).
Along with subjectivity in labels, mislabelling is an occasional hazard of any dataset requiring mass labelling --- fortunately, we did not find too many occurrences of mislabelled points when examining hard positives and negatives in \rococo.

\begin{figure}[t!] % "[t!]" placement specifier just for this example
\begin{subfigure}{0.33\textwidth}
\includegraphics[width=\linewidth]{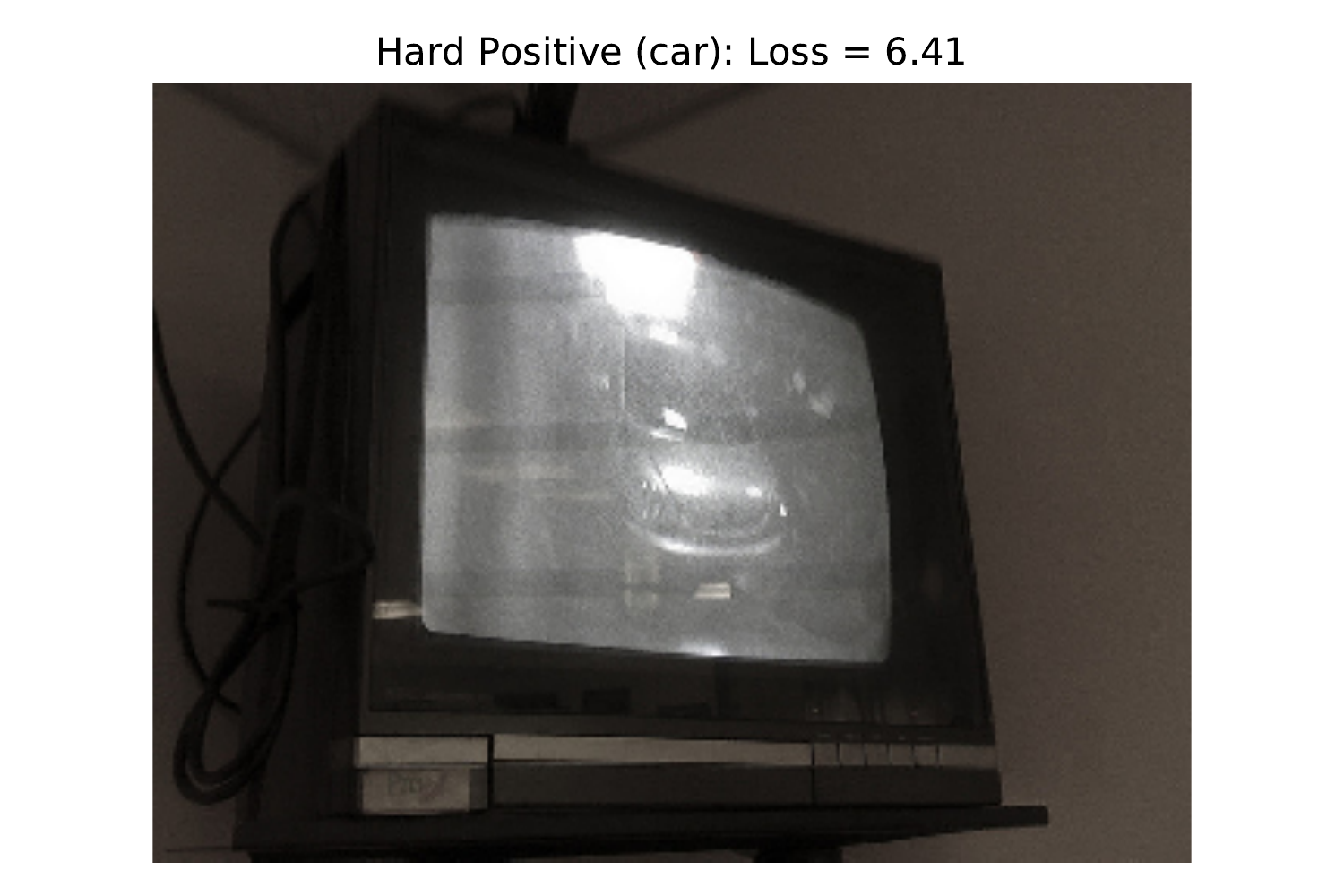}
\end{subfigure}\hspace*{\fill}
\begin{subfigure}{0.33\textwidth}
\includegraphics[width=\linewidth]{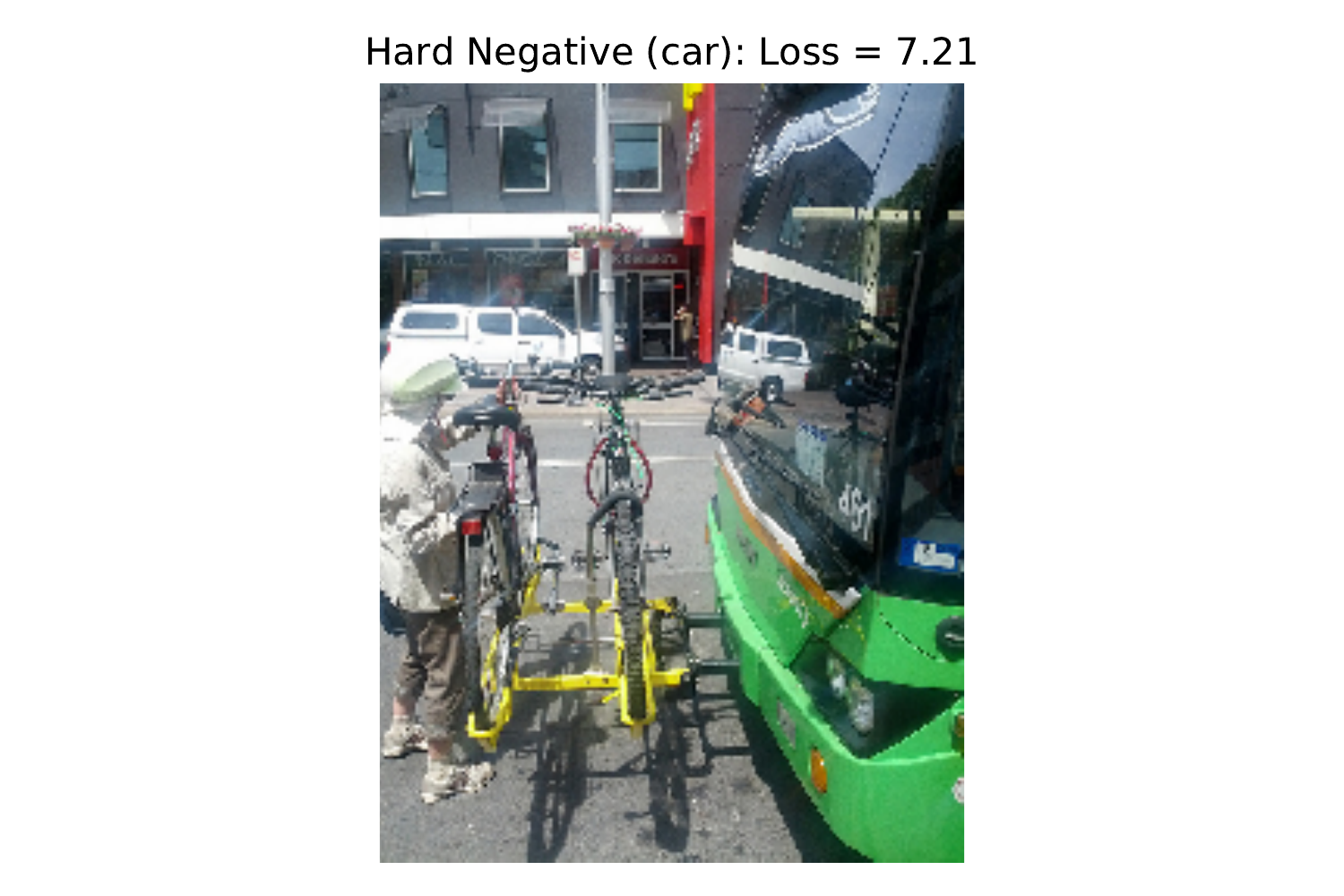}
\end{subfigure}\hspace*{\fill}
\begin{subfigure}{0.33\textwidth}
\includegraphics[width=\linewidth]{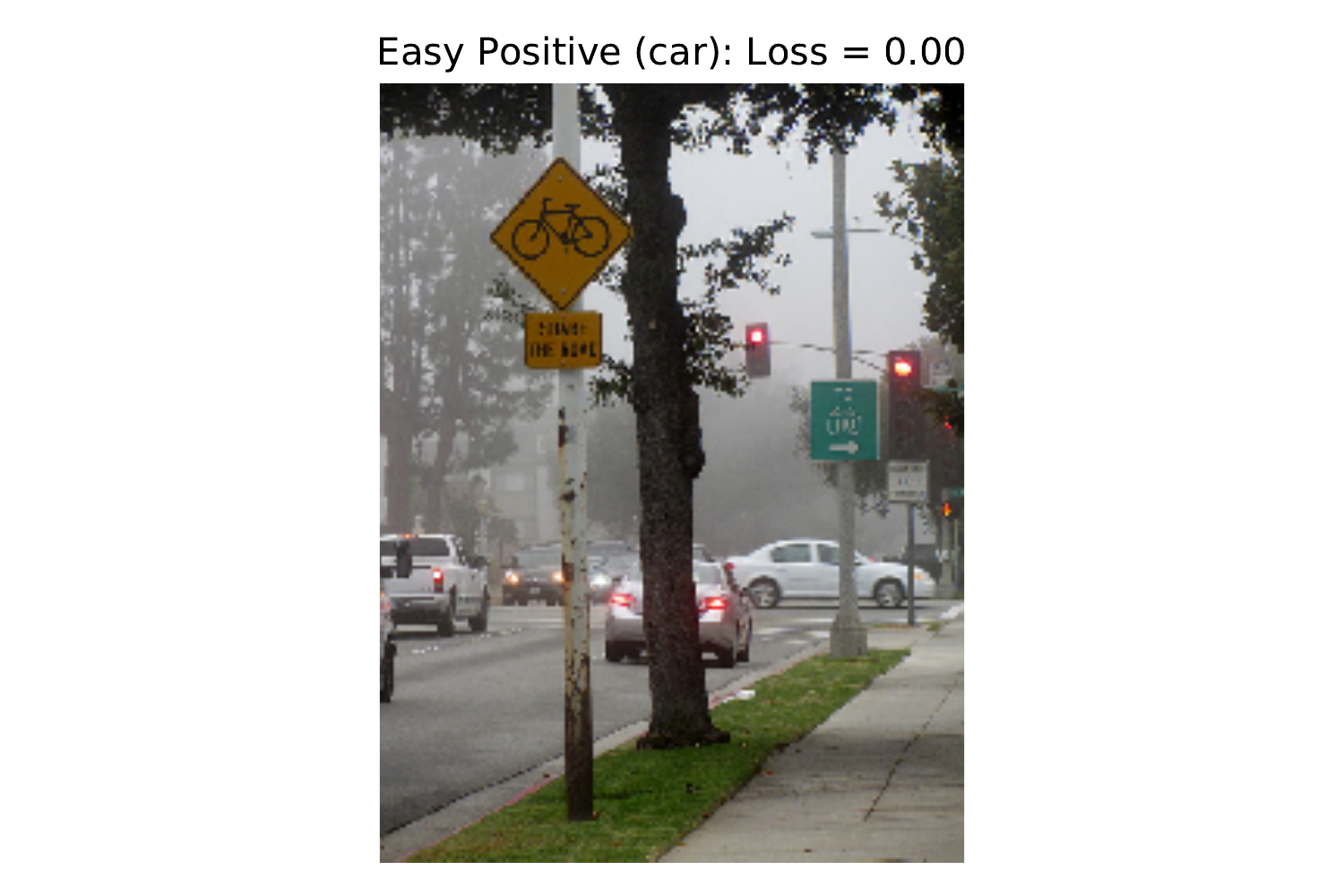}
\end{subfigure}
\caption{Examples from the \texttt{car} task.} 
\end{figure}

\begin{figure}[t!] % "[t!]" placement specifier just for this example
\begin{subfigure}{0.33\textwidth}
\includegraphics[width=\linewidth]{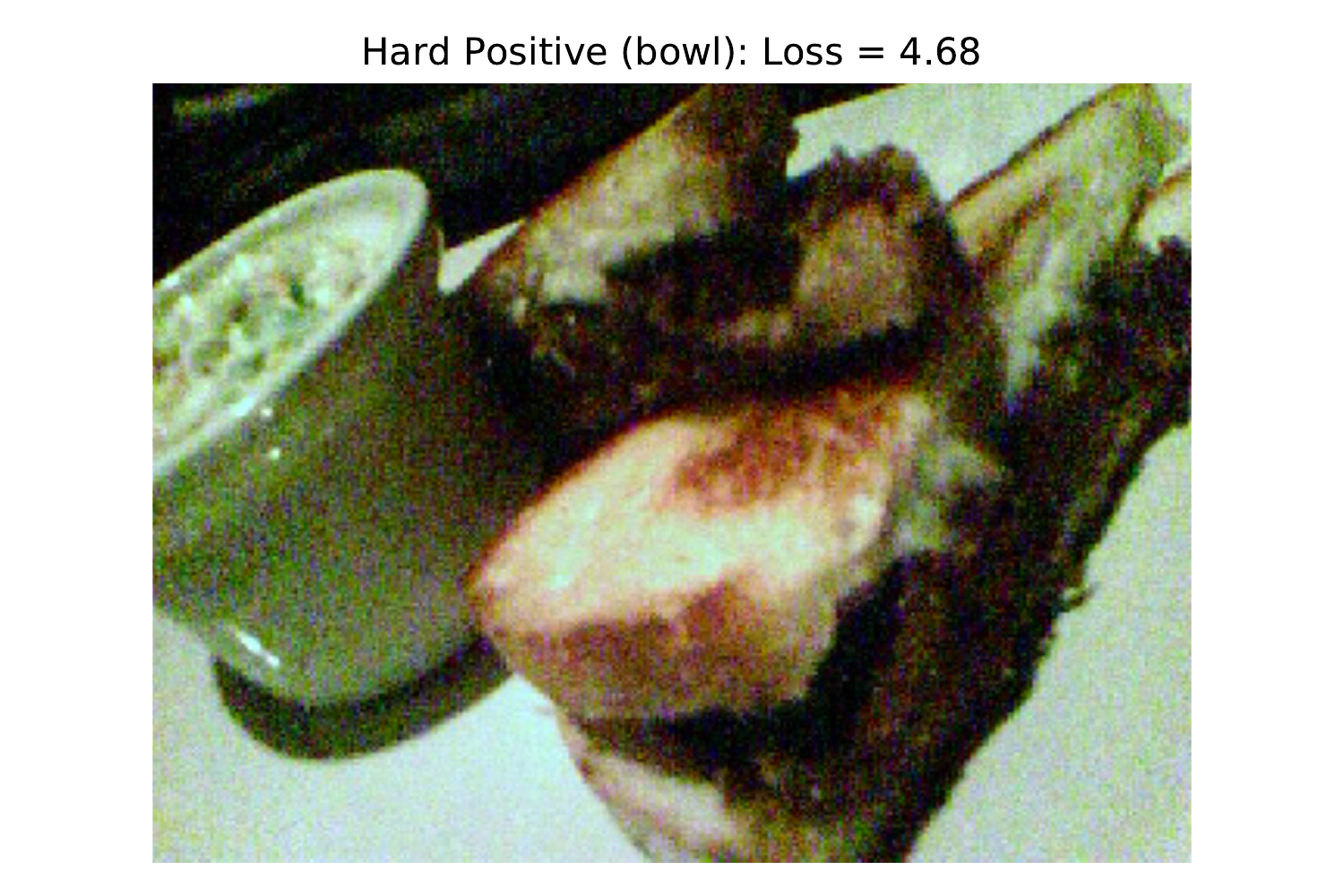}
\end{subfigure}\hspace*{\fill}
\begin{subfigure}{0.33\textwidth}
\includegraphics[width=\linewidth]{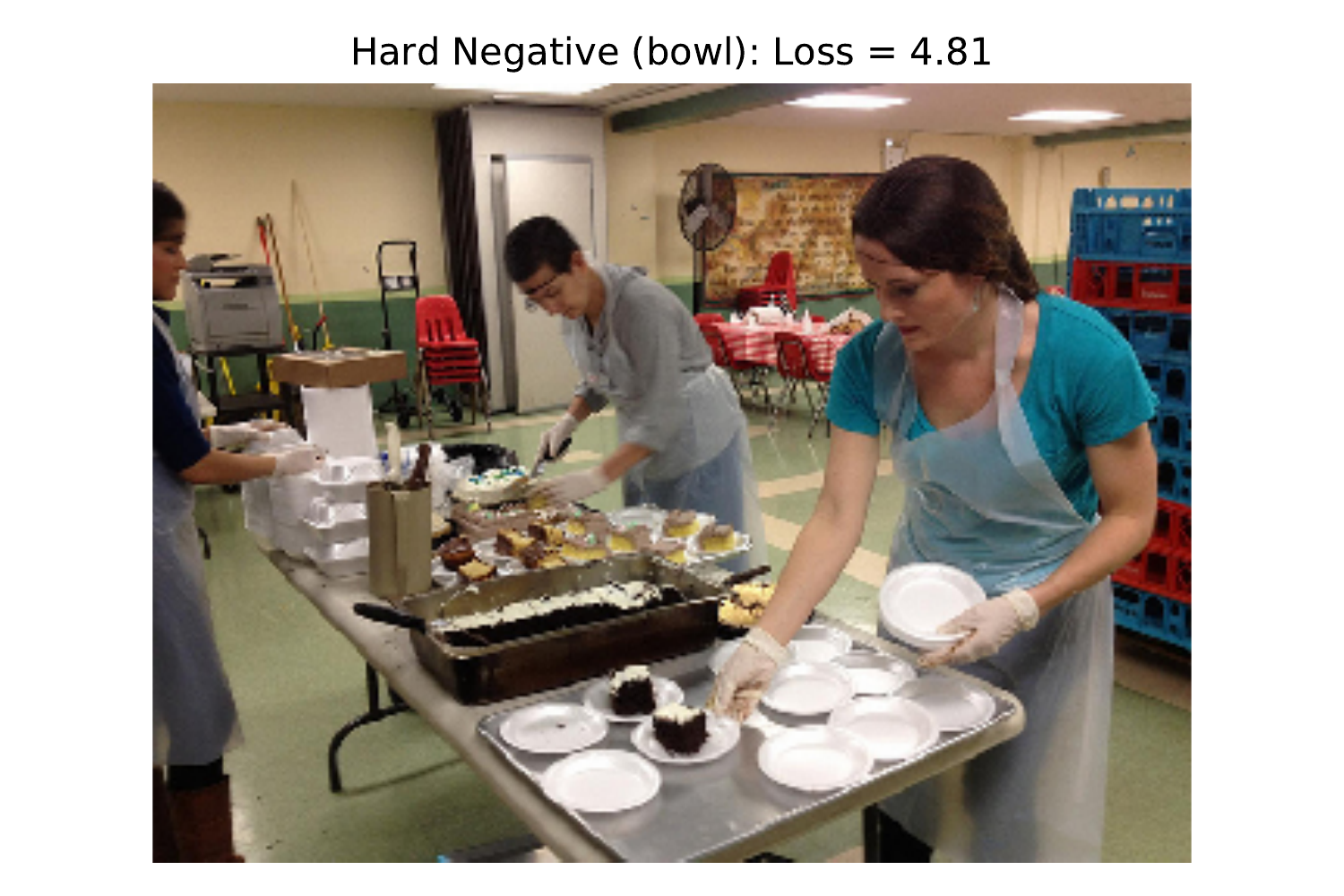}
\end{subfigure}\hspace*{\fill}
\begin{subfigure}{0.33\textwidth}
\includegraphics[width=\linewidth]{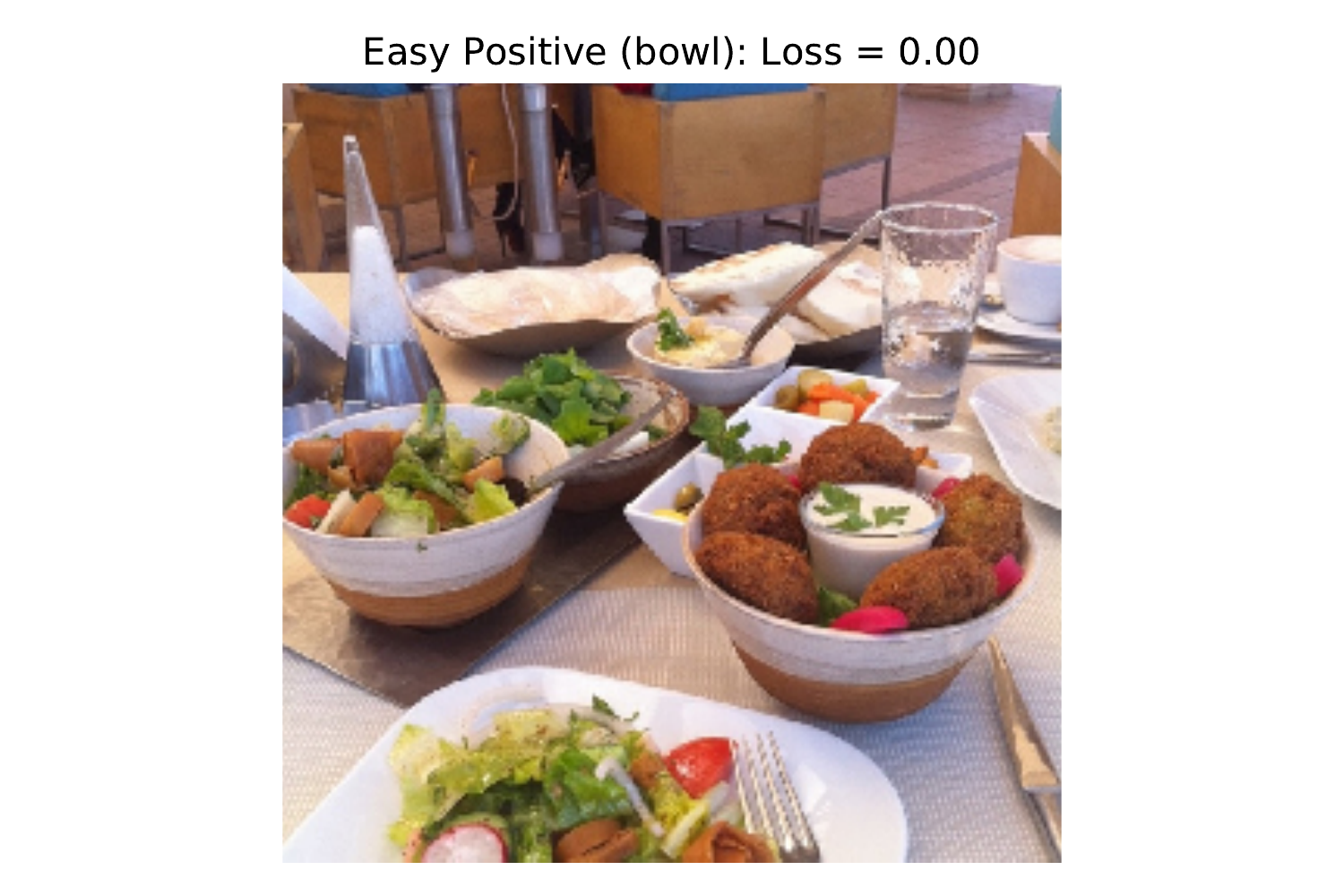}
\end{subfigure}
\caption{Examples from the \texttt{bowl} task.} 
\end{figure}

\begin{figure}[t!] % "[t!]" placement specifier just for this example
\begin{subfigure}{0.33\textwidth}
\includegraphics[width=\linewidth]{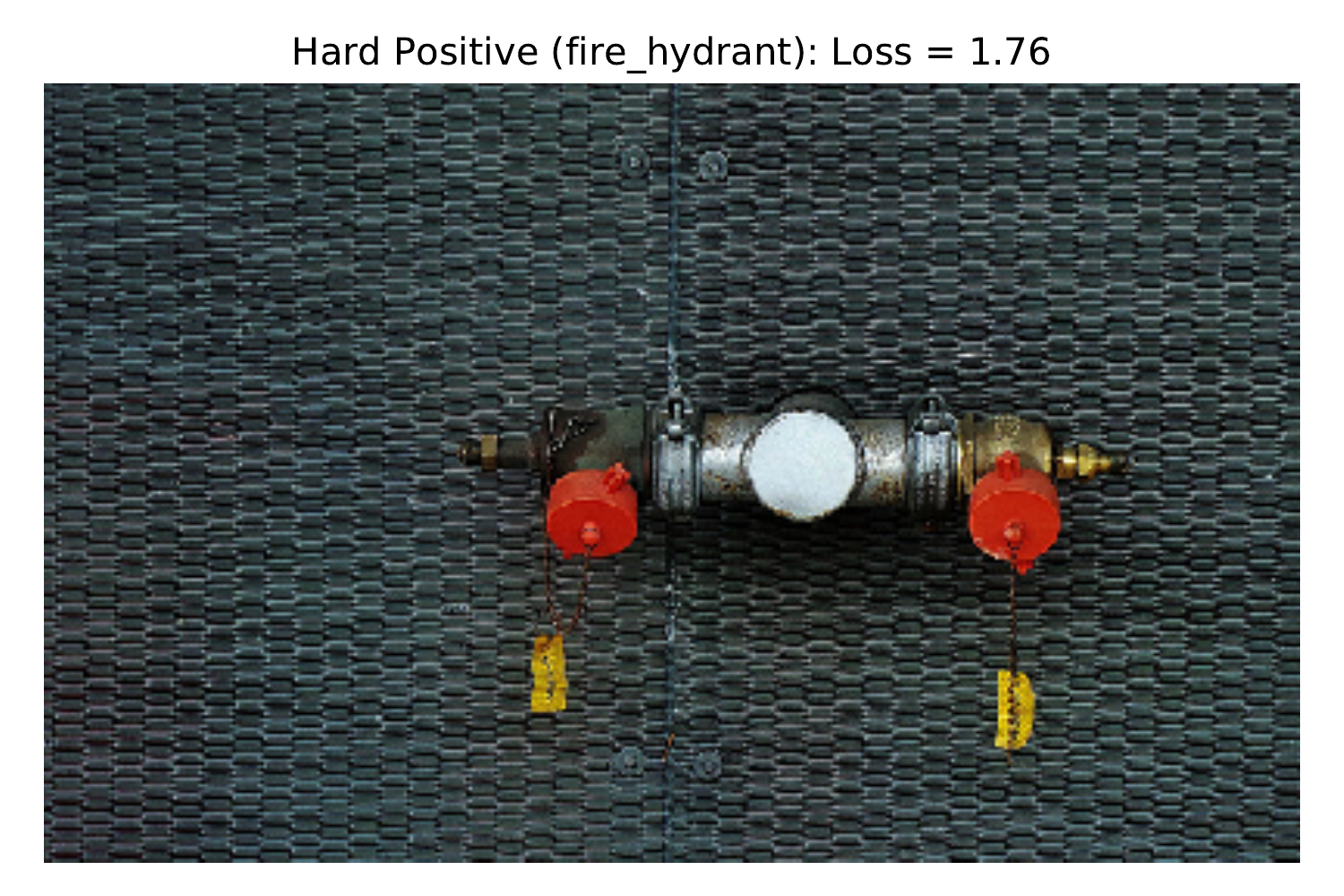}
\end{subfigure}\hspace*{\fill}
\begin{subfigure}{0.33\textwidth}
\includegraphics[width=\linewidth]{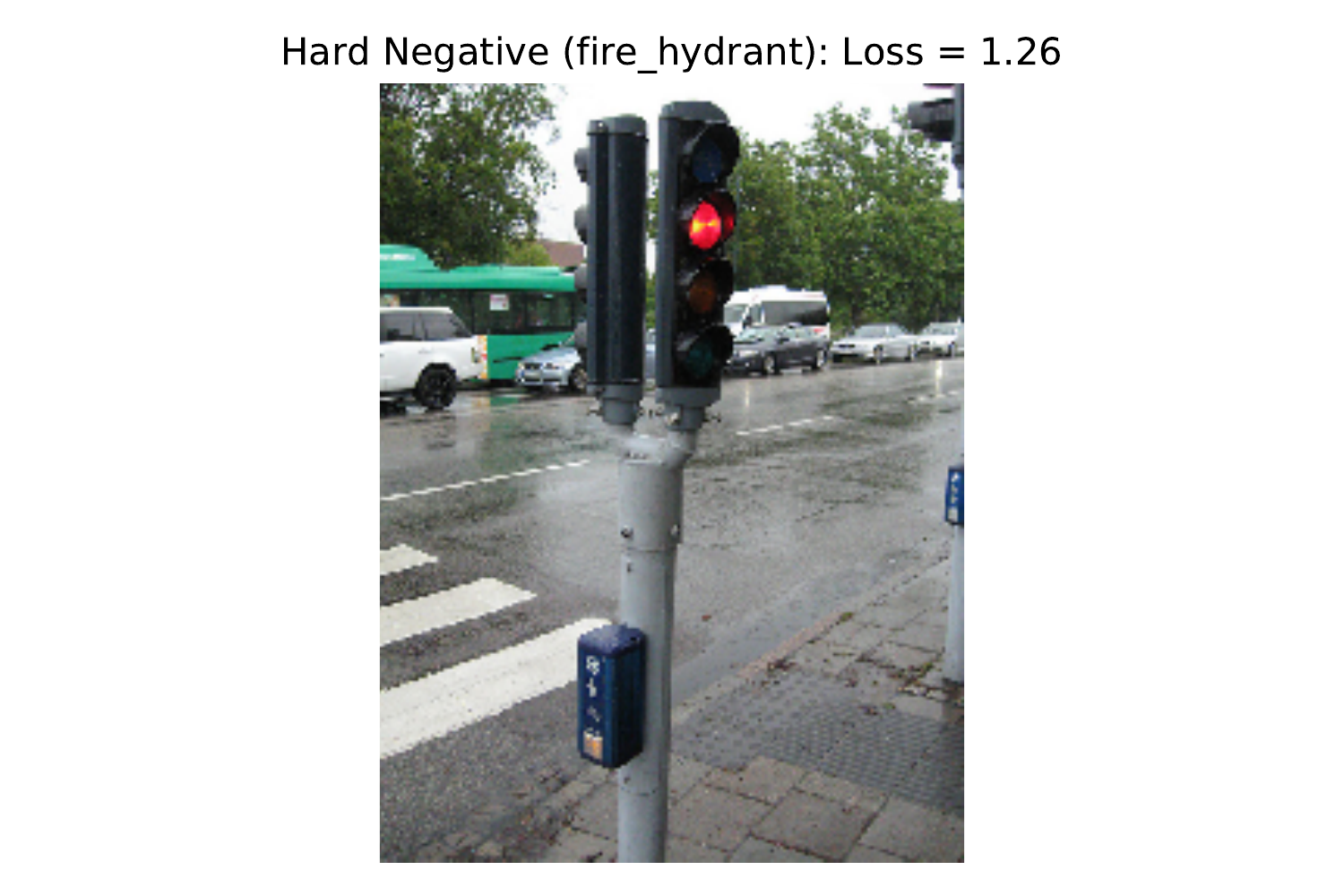}
\end{subfigure}\hspace*{\fill}
\begin{subfigure}{0.33\textwidth}
\includegraphics[width=\linewidth]{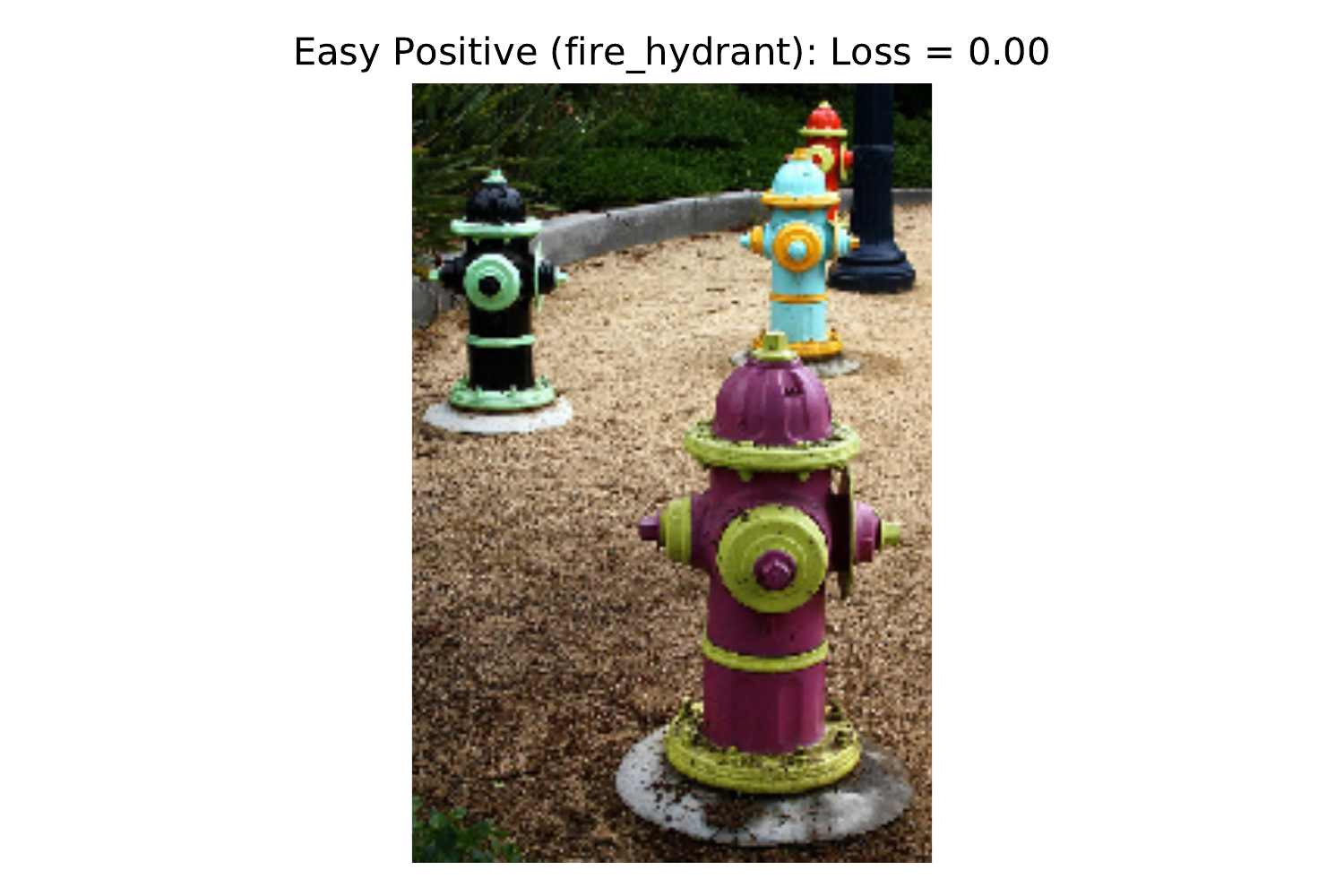}
\end{subfigure}
\caption{Examples from the \texttt{fire\_hydrant} task.} 
\end{figure}

\begin{figure}[t!] % "[t!]" placement specifier just for this example
\begin{subfigure}{0.33\textwidth}
\includegraphics[width=\linewidth]{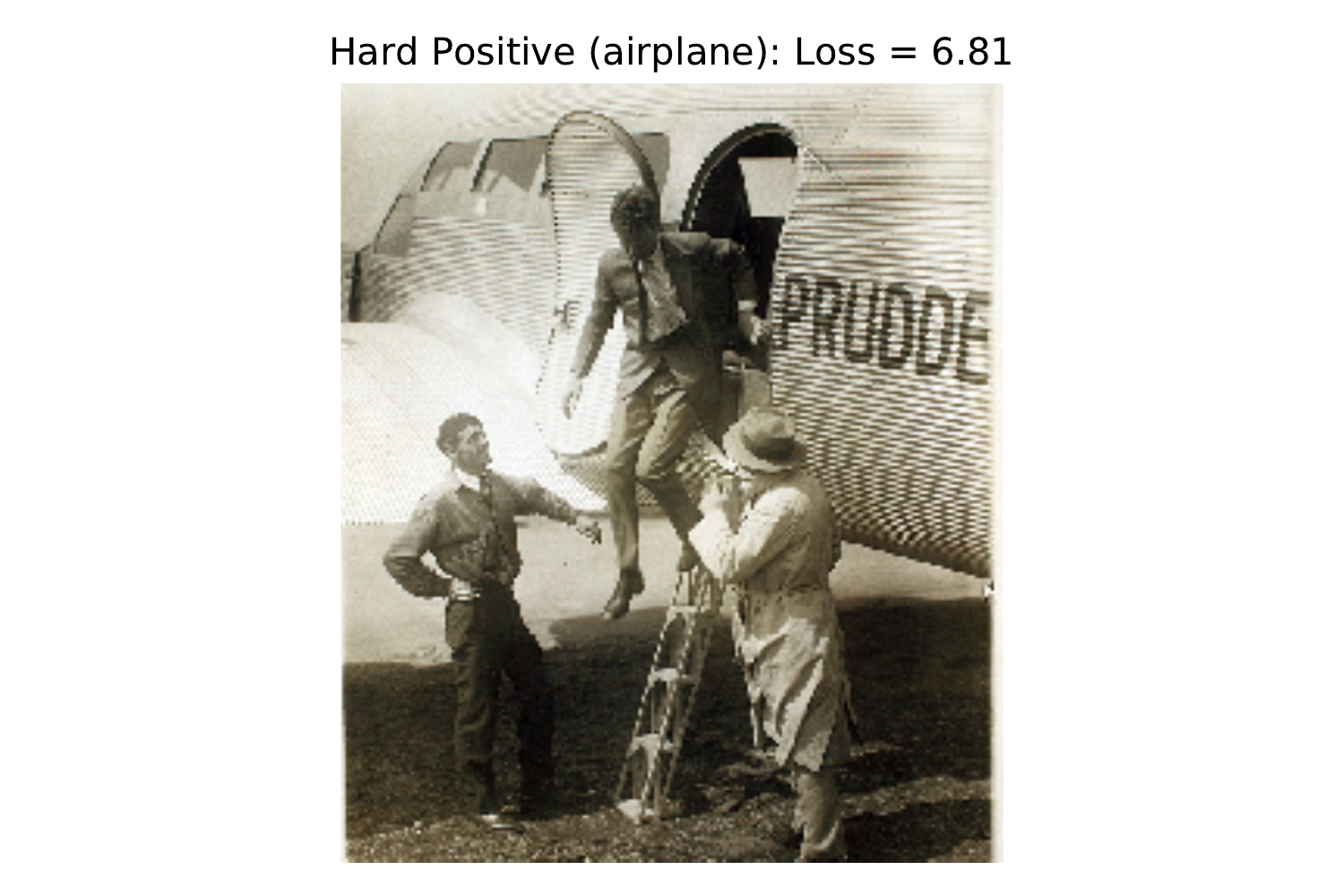}
\end{subfigure}\hspace*{\fill}
\begin{subfigure}{0.33\textwidth}
\includegraphics[width=\linewidth]{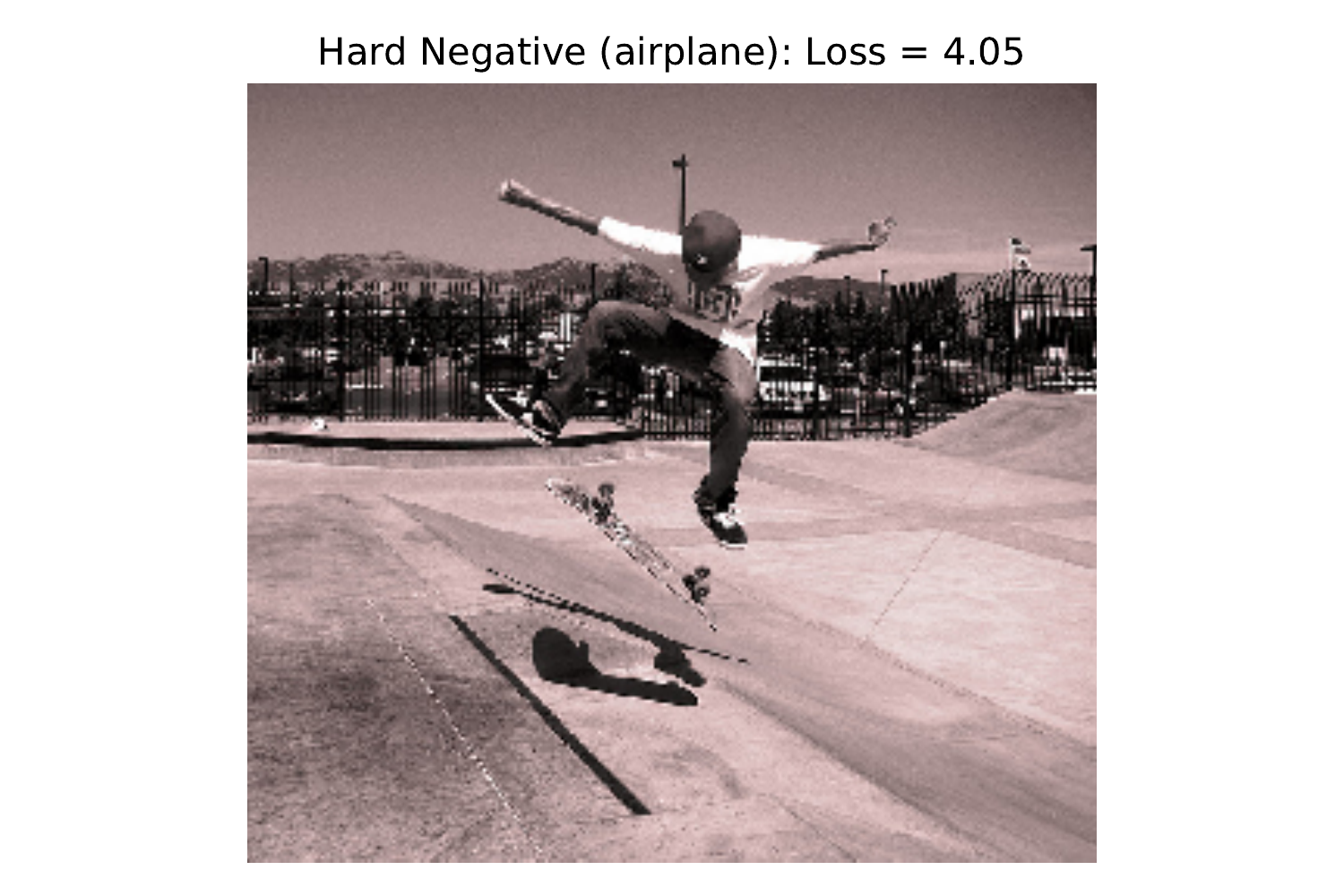}
\end{subfigure}\hspace*{\fill}
\begin{subfigure}{0.33\textwidth}
\includegraphics[width=\linewidth]{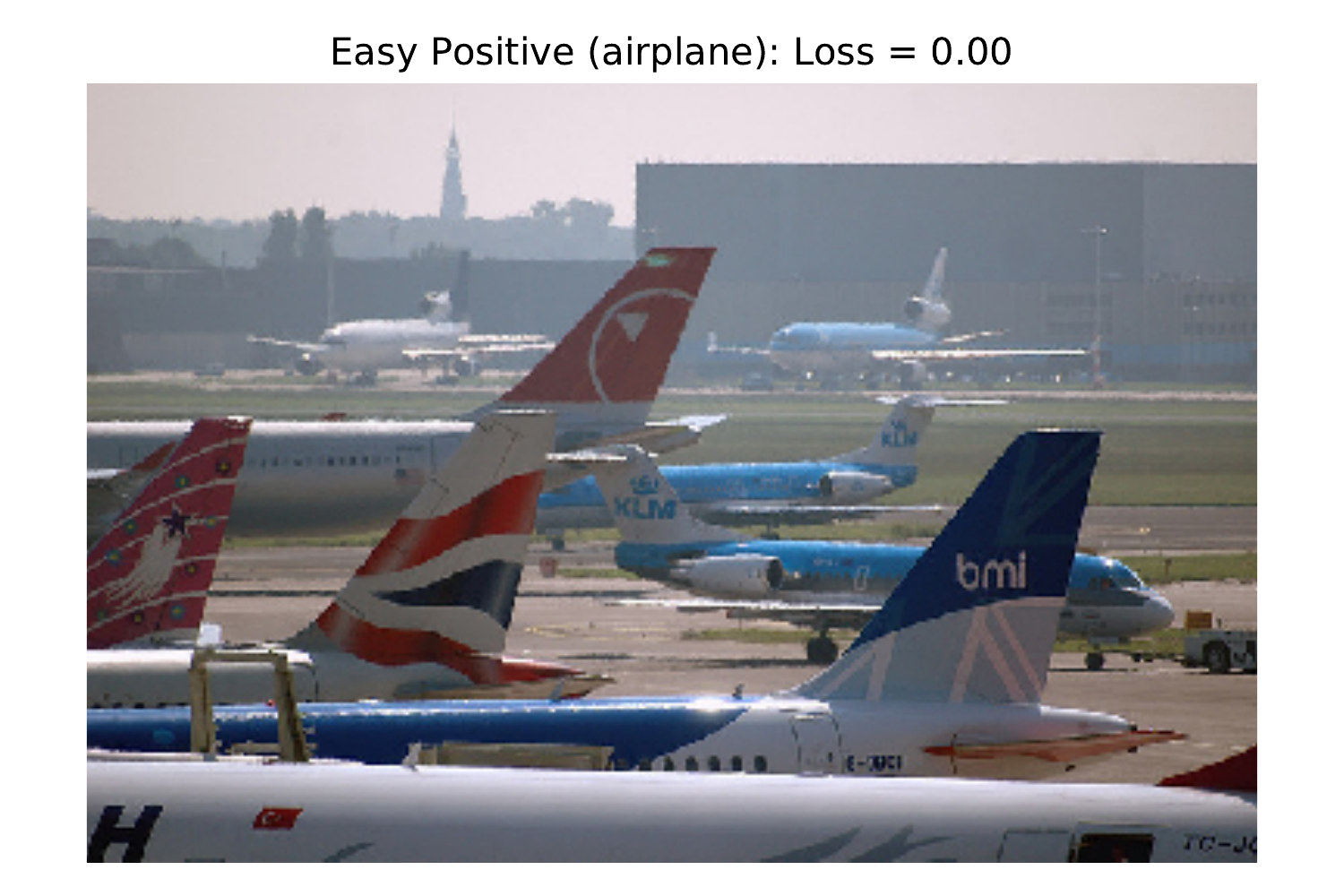}
\end{subfigure}
\caption{Examples from the \texttt{airplane} task.} 
\end{figure}

\begin{figure}[t!] % "[t!]" placement specifier just for this example
\begin{subfigure}{0.33\textwidth}
\includegraphics[width=\linewidth]{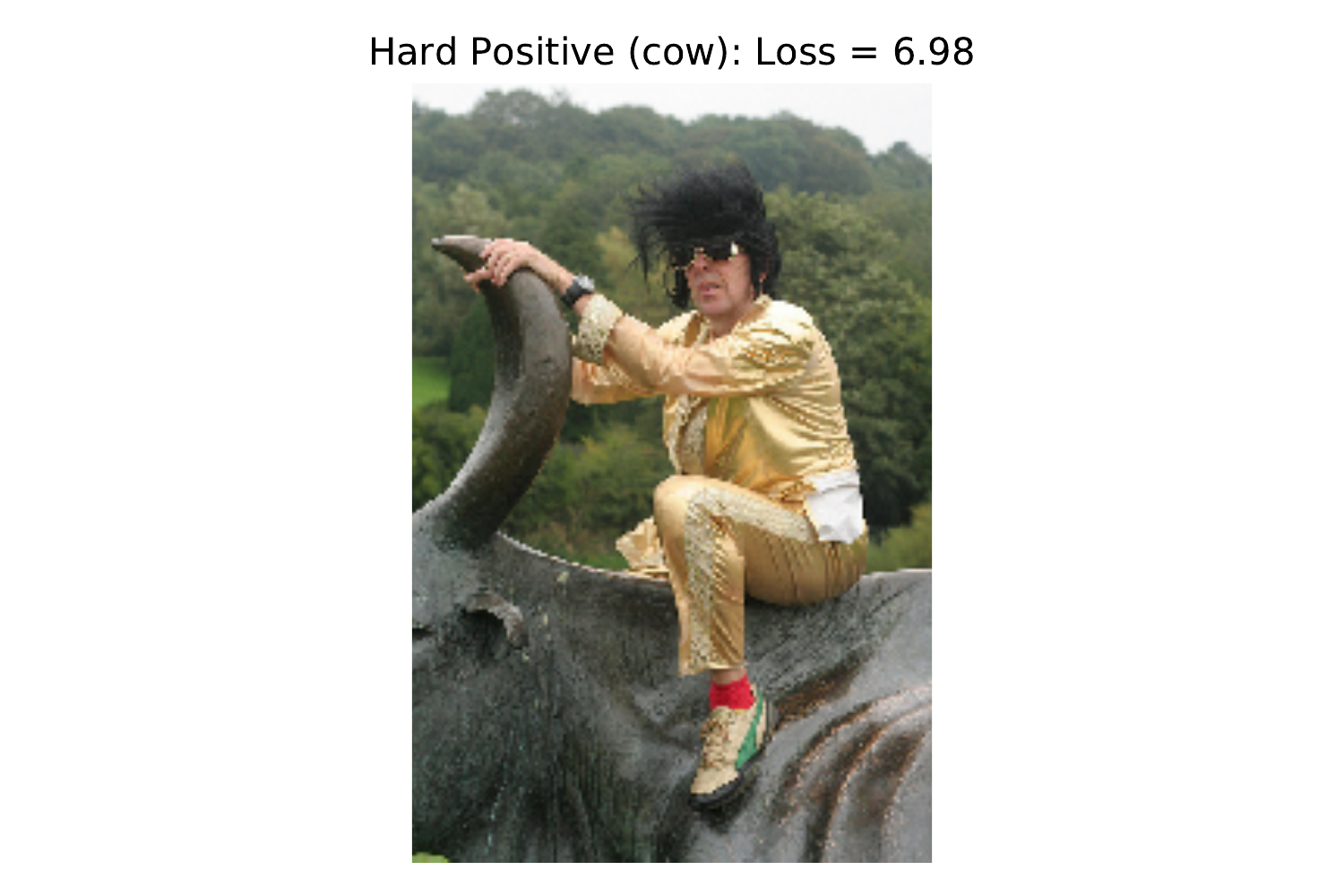}
\end{subfigure}\hspace*{\fill}
\begin{subfigure}{0.33\textwidth}
\includegraphics[width=\linewidth]{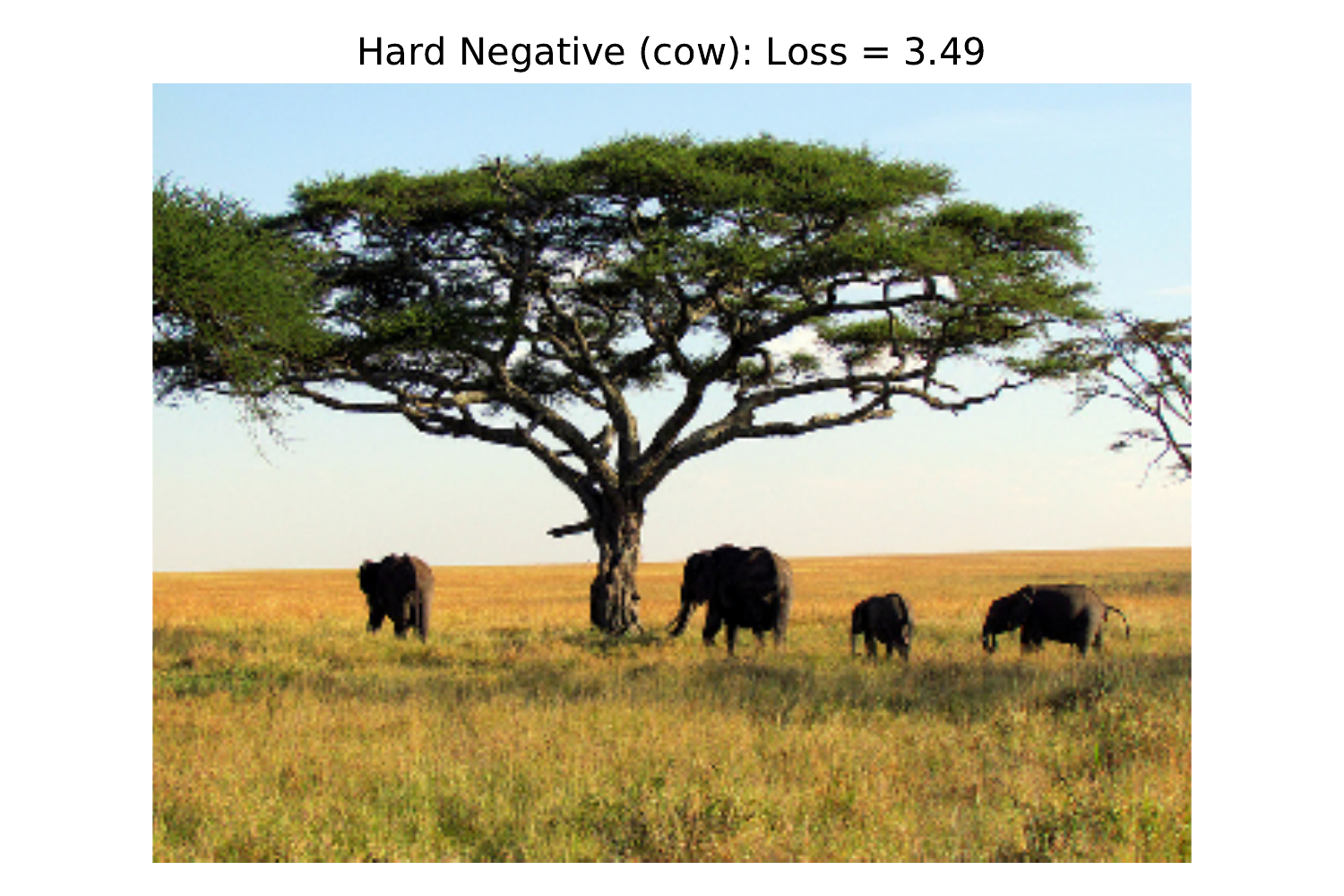}
\end{subfigure}\hspace*{\fill}
\begin{subfigure}{0.33\textwidth}
\includegraphics[width=\linewidth]{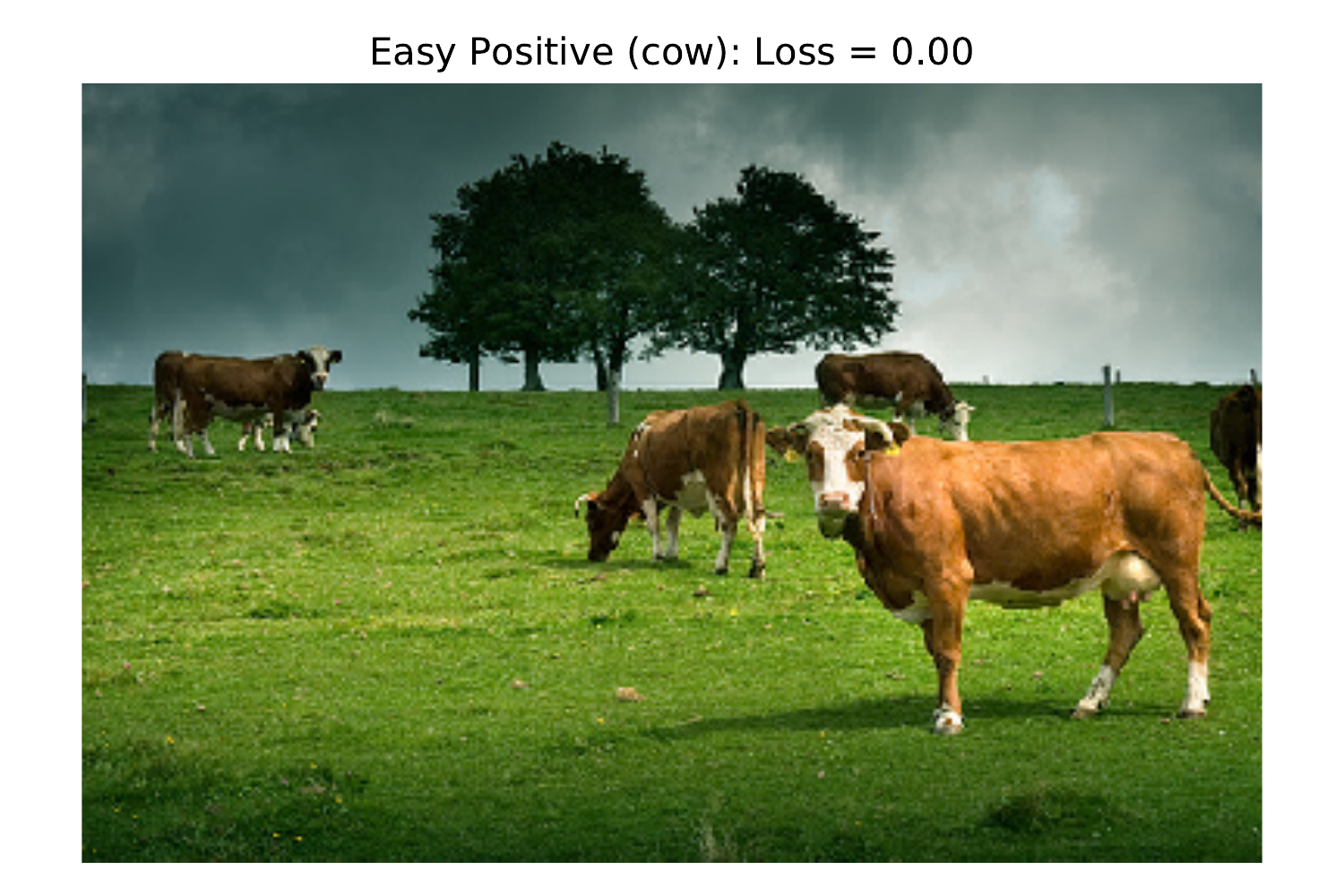}
\end{subfigure}
\caption{Examples from the \texttt{cow} task.} 
\end{figure}

\begin{figure}[t!] % "[t!]" placement specifier just for this example
\begin{subfigure}{0.33\textwidth}
\includegraphics[width=\linewidth]{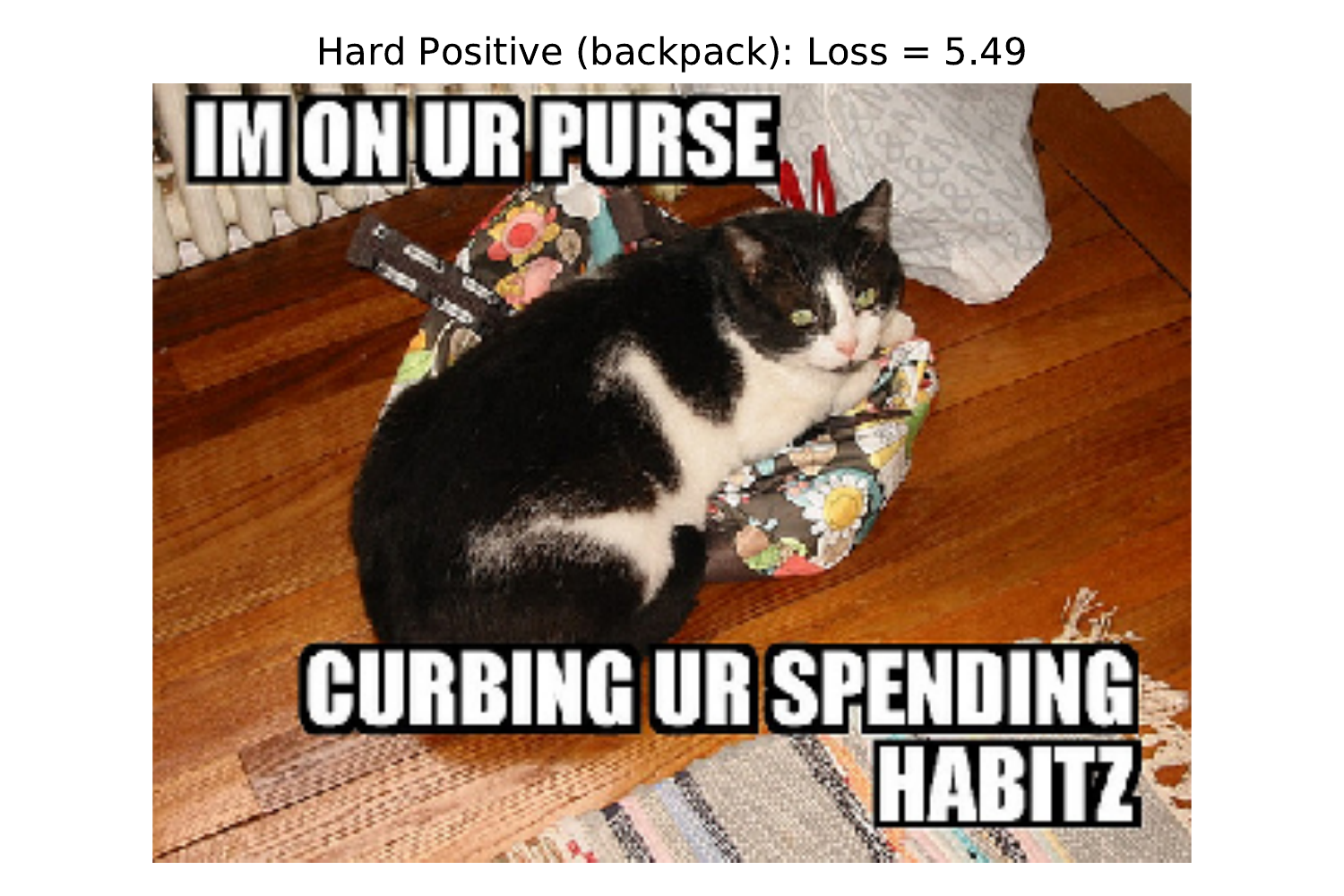}
\end{subfigure}\hspace*{\fill}
\begin{subfigure}{0.33\textwidth}
\includegraphics[width=\linewidth]{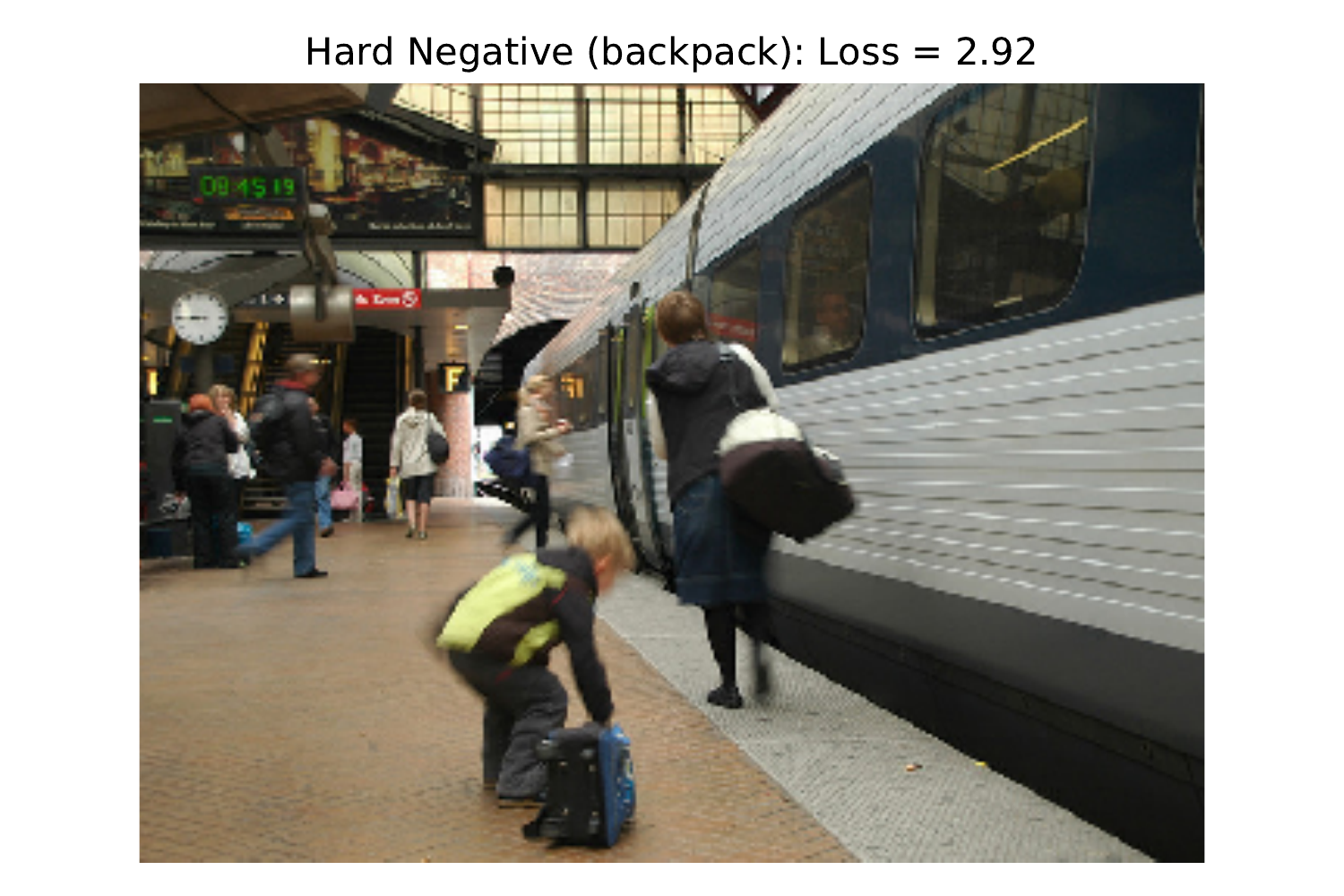}
\end{subfigure}\hspace*{\fill}
\begin{subfigure}{0.33\textwidth}
\includegraphics[width=\linewidth]{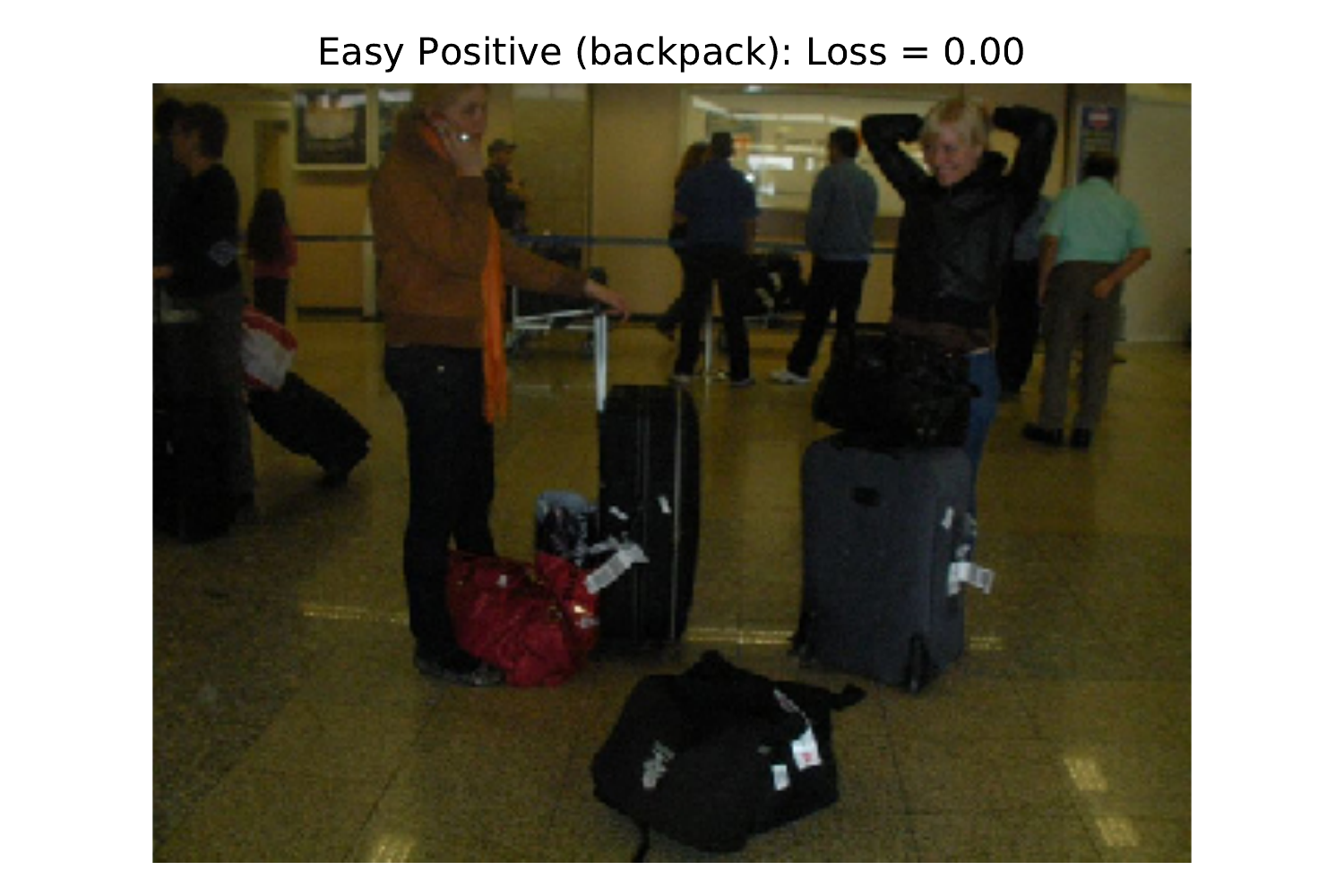}
\end{subfigure}
\caption{Examples from the \texttt{backpack} task.} 
\end{figure}

\begin{figure}[t!] % "[t!]" placement specifier just for this example
\begin{subfigure}{0.33\textwidth}
\includegraphics[width=\linewidth]{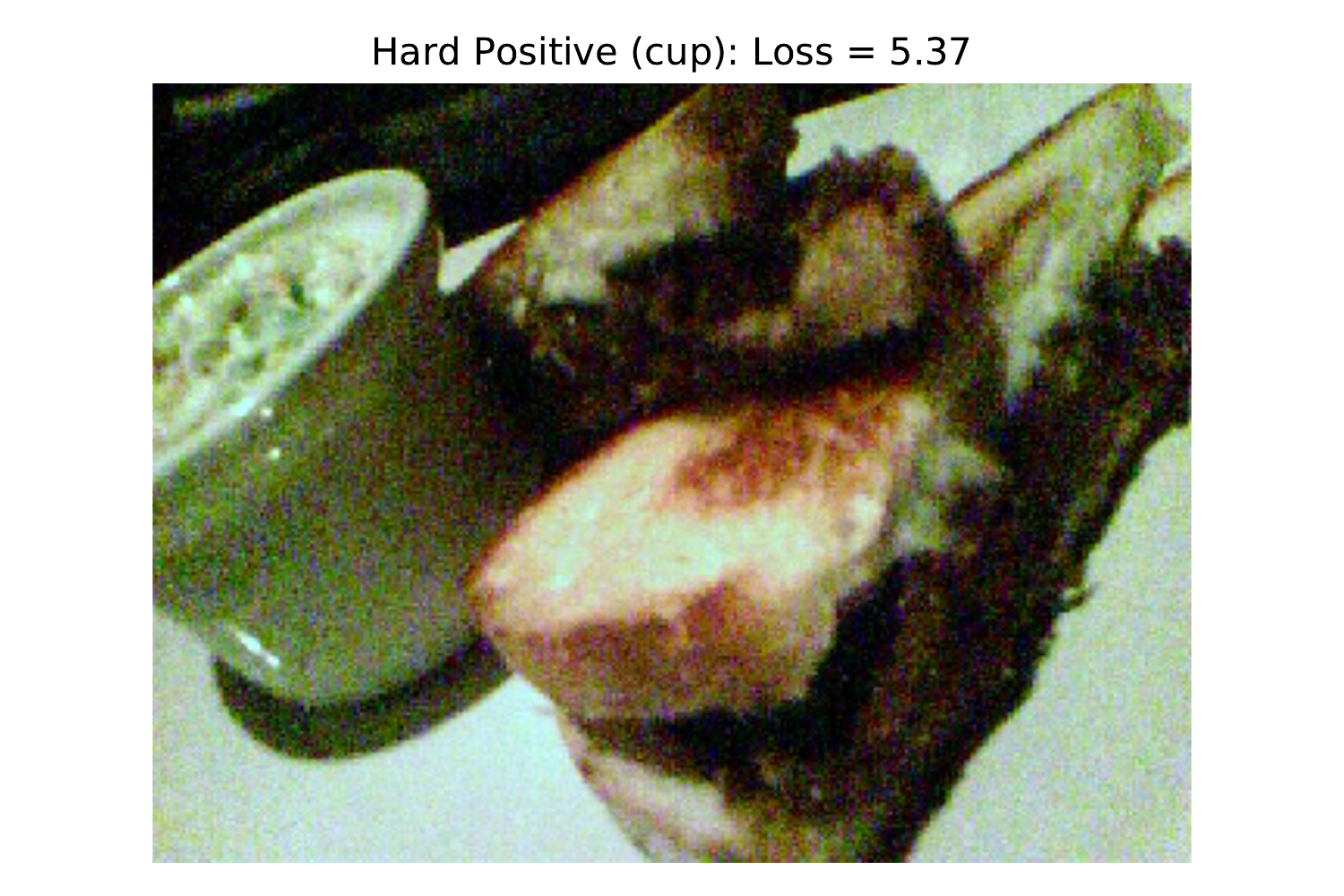}
\end{subfigure}\hspace*{\fill}
\begin{subfigure}{0.33\textwidth}
\includegraphics[width=\linewidth]{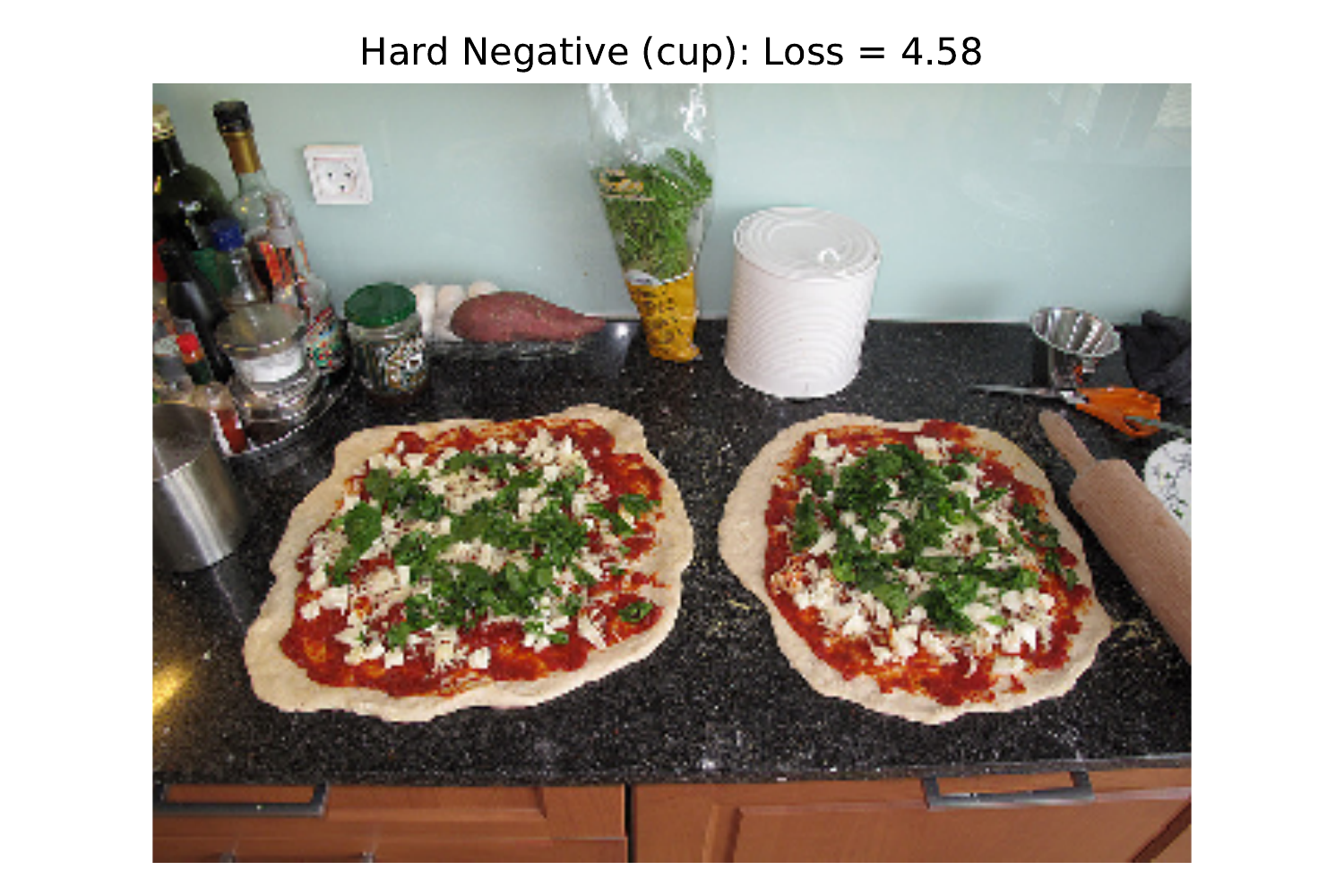}
\end{subfigure}\hspace*{\fill}
\begin{subfigure}{0.33\textwidth}
\includegraphics[width=\linewidth]{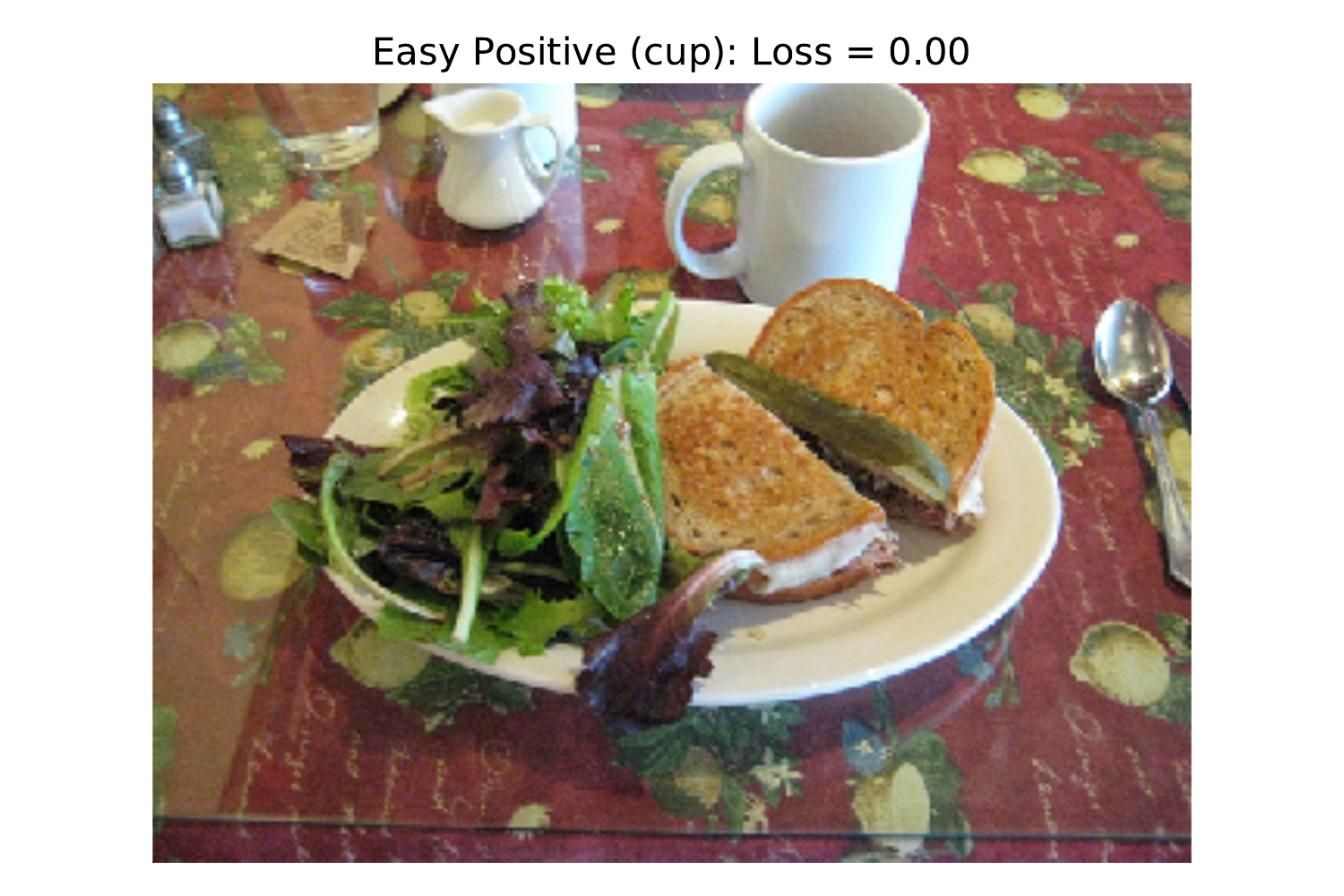}
\end{subfigure}
\caption{Examples from the \texttt{cup} task.} 
\end{figure}

\begin{figure}[t!] % "[t!]" placement specifier just for this example
\begin{subfigure}{0.33\textwidth}
\includegraphics[width=\linewidth]{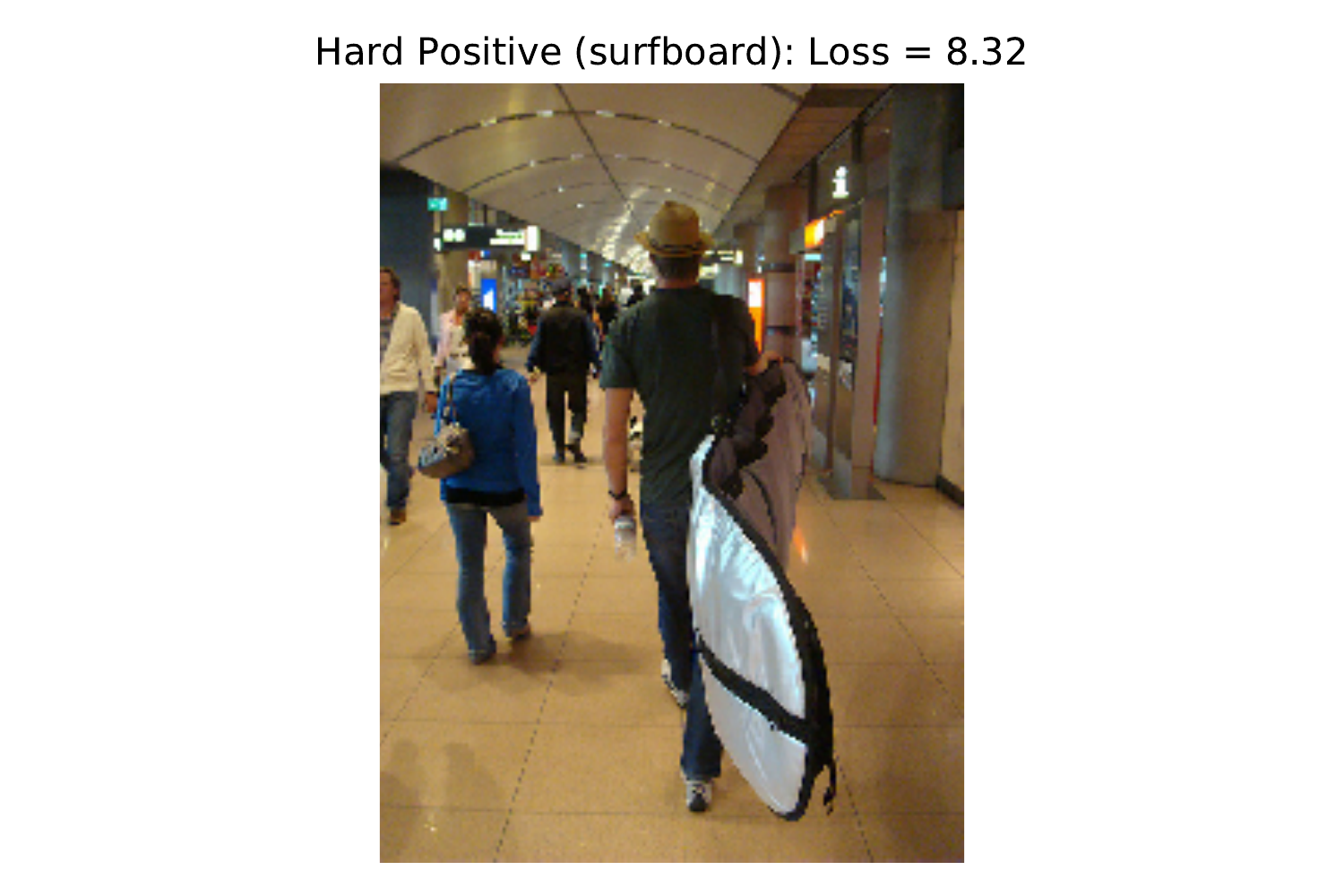}
\end{subfigure}\hspace*{\fill}
\begin{subfigure}{0.33\textwidth}
\includegraphics[width=\linewidth]{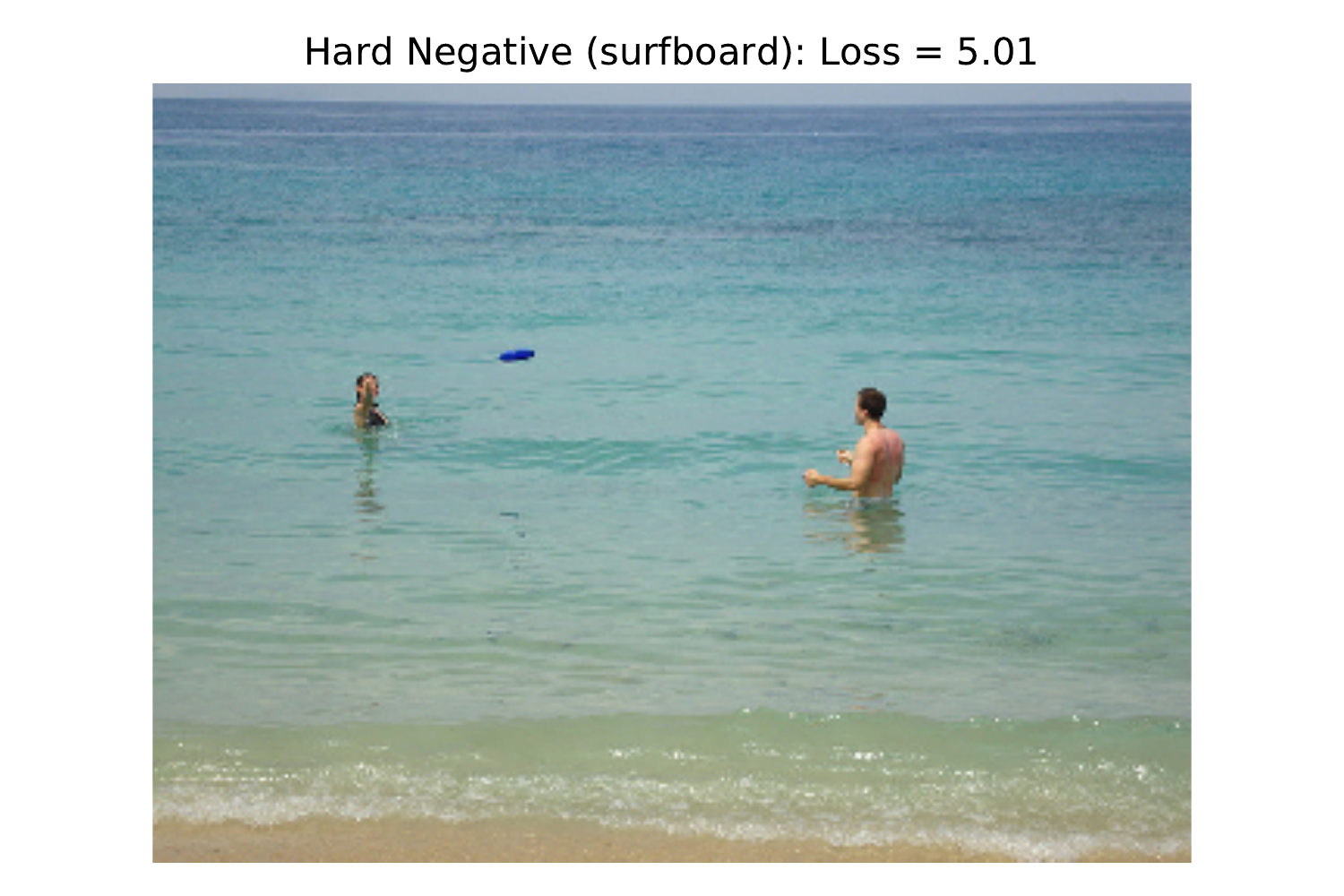}
\end{subfigure}\hspace*{\fill}
\begin{subfigure}{0.33\textwidth}
\includegraphics[width=\linewidth]{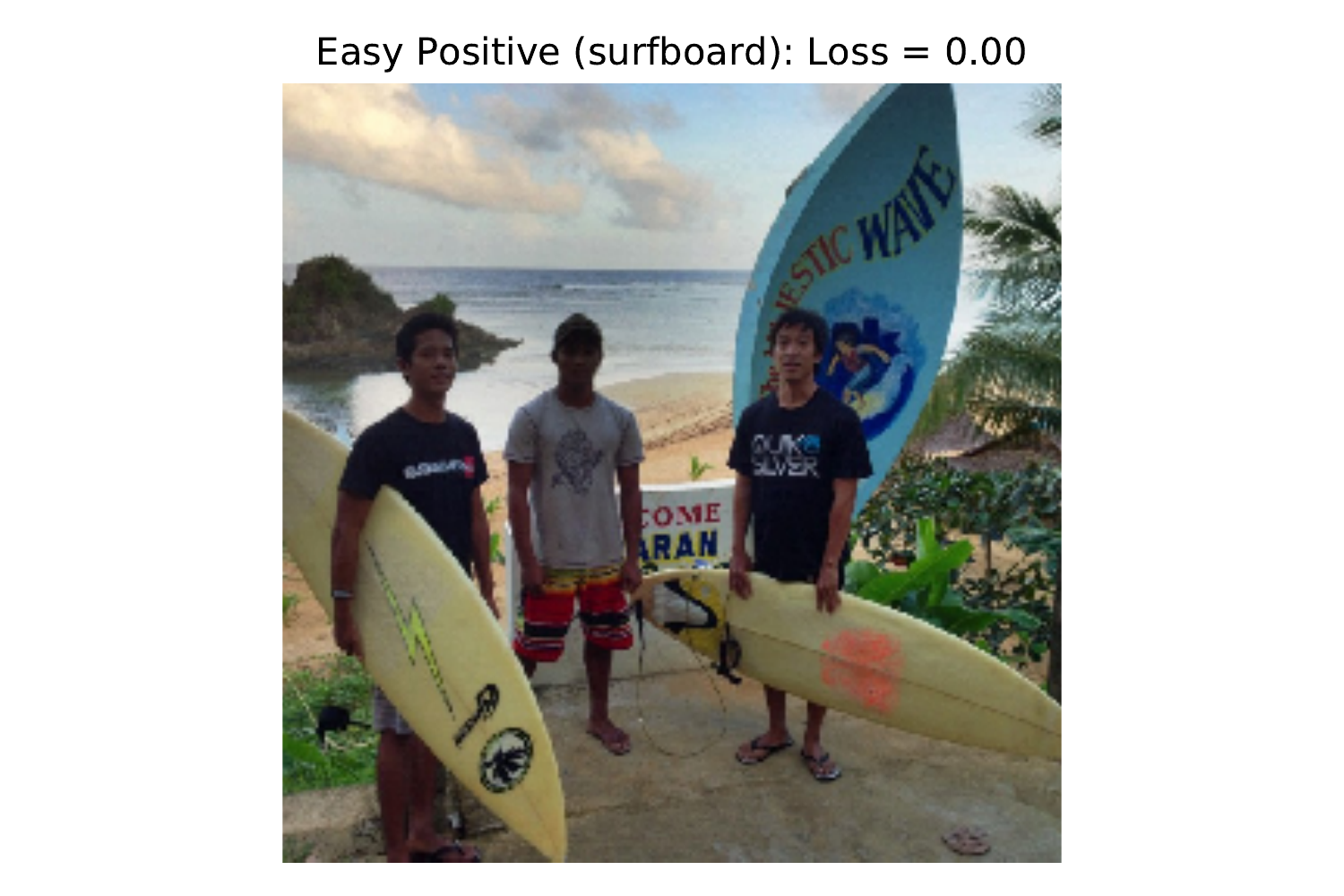}
\end{subfigure}
\caption{Examples from the \texttt{surfboard} task.} 
\end{figure}

\begin{figure}[t!] % "[t!]" placement specifier just for this example
\begin{subfigure}{0.33\textwidth}
\includegraphics[width=\linewidth]{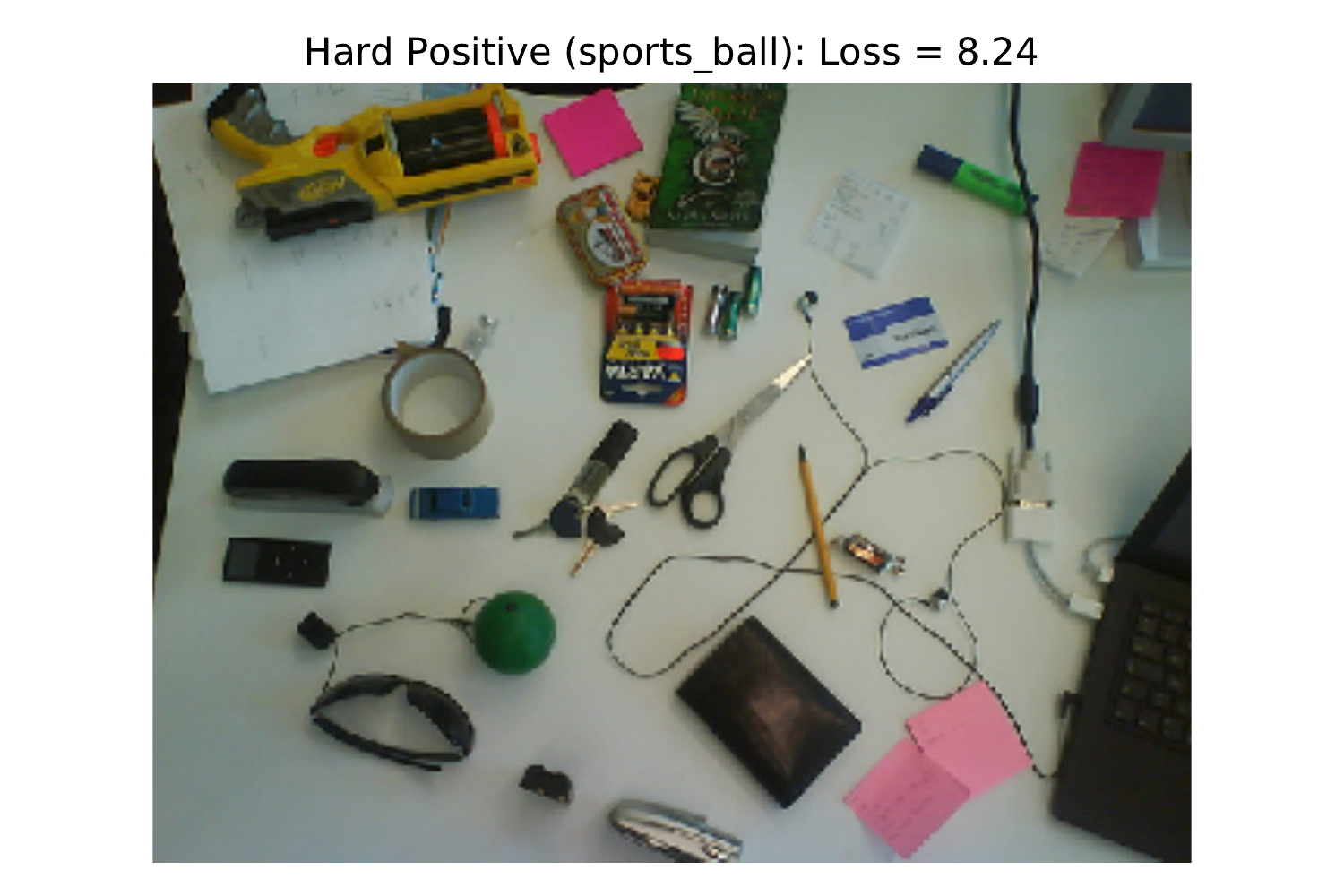}
\end{subfigure}\hspace*{\fill}
\begin{subfigure}{0.33\textwidth}
\includegraphics[width=\linewidth]{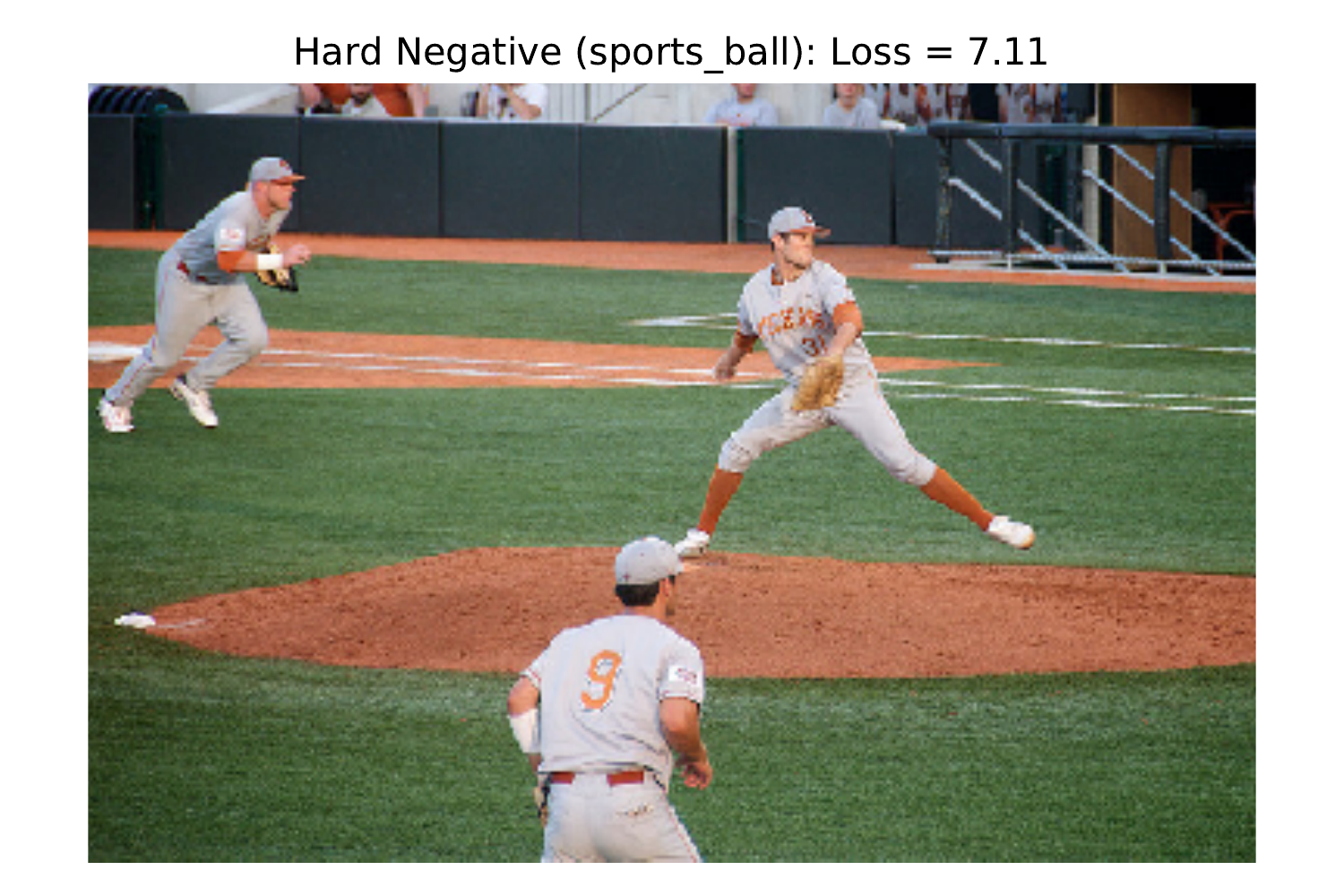}
\end{subfigure}\hspace*{\fill}
\begin{subfigure}{0.33\textwidth}
\includegraphics[width=\linewidth]{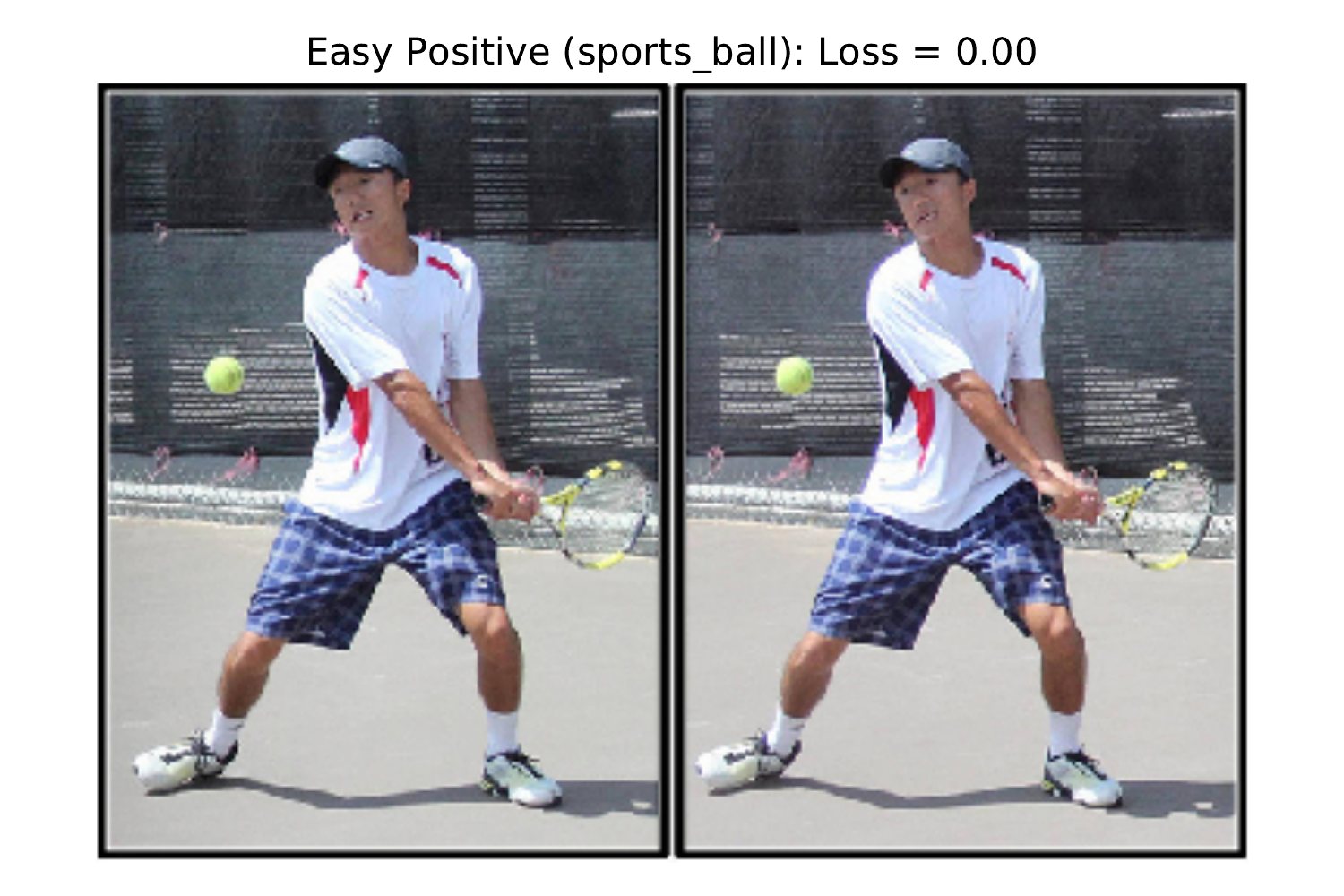}
\end{subfigure}
\caption{Examples from the \texttt{sports\_ball} task.} 
\end{figure}

\begin{figure}[t!] % "[t!]" placement specifier just for this example
\begin{subfigure}{0.33\textwidth}
\includegraphics[width=\linewidth]{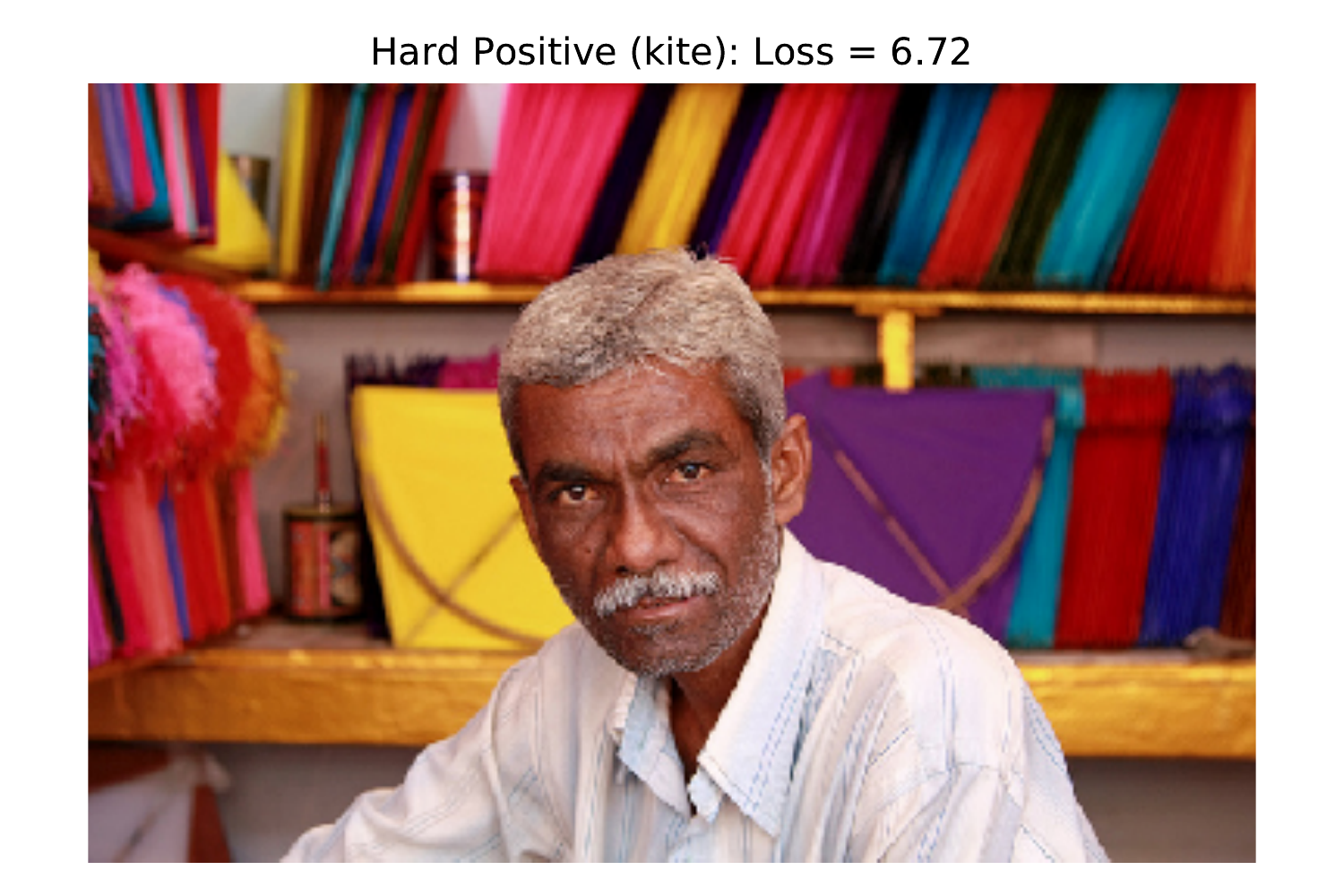}
\end{subfigure}\hspace*{\fill}
\begin{subfigure}{0.33\textwidth}
\includegraphics[width=\linewidth]{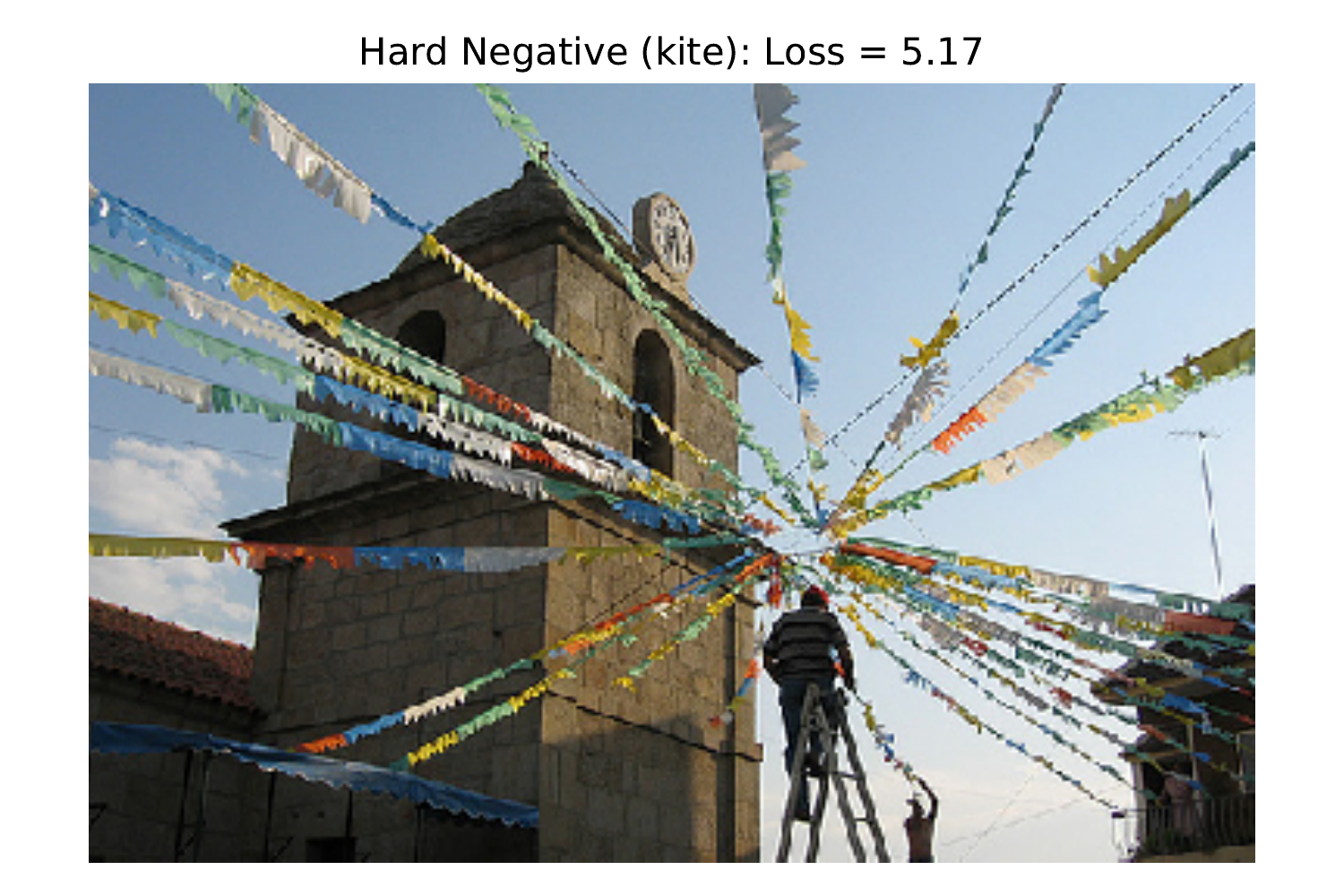}
\end{subfigure}\hspace*{\fill}
\begin{subfigure}{0.33\textwidth}
\includegraphics[width=\linewidth]{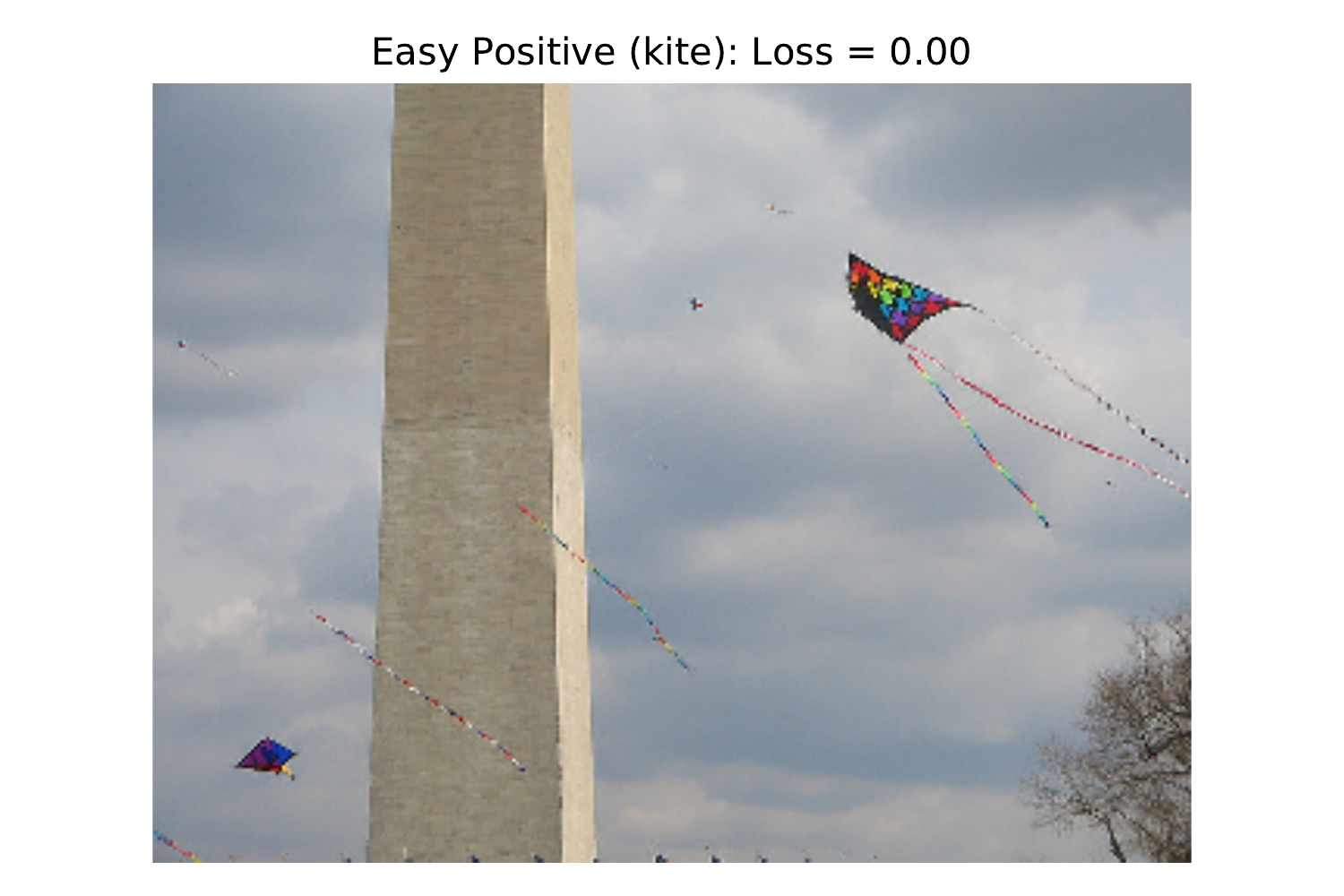}
\end{subfigure}
\caption{Examples from the \texttt{kite} task.} 
\end{figure}

% \vspace{20cm}
\subsection{Analyzing GDRO's Errors}
\begin{wrapfigure}{R}{0.6\linewidth} % "[t!]" placement specifier just for this example
% \begin{figure} % "[t!]" placement specifier just for this example
% \hspace*{-0.5cm}
% \begin{subfigure}{0.3\textwidth}
% \includegraphics[width=\linewidth]{figs/results/qual_examples/example_loss_diff_GDRO_over_ERM_car_Hard_Positive.pdf}
% % \caption{First subfigure} \label{fig:a}
% \end{subfigure}%\hspace*{\fill}
% \hspace*{-2cm}
% \begin{subfigure}{0.3\textwidth}
% \includegraphics[width=\linewidth]{figs/results/qual_examples/example_loss_diff_GDRO_over_ERM_car_Hard_Negative.pdf}
% % \caption{Second subfigure} \label{fig:b}
% \end{subfigure}%\hspace*{\fill}
% \hspace*{-2cm}
% \begin{subfigure}{0.33\textwidth}
% \includegraphics[width=\linewidth]{figs/results/qual_examples/example_loss_diff_GDRO_over_ERM_car_Easy_Positive.pdf}
% % \caption{Second subfigure} \label{fig:b}
% \end{subfigure}

% \medskip
\begin{subfigure}{0.3\textwidth}
% \hspace*{-1.5cm}
\includegraphics[width=\linewidth]{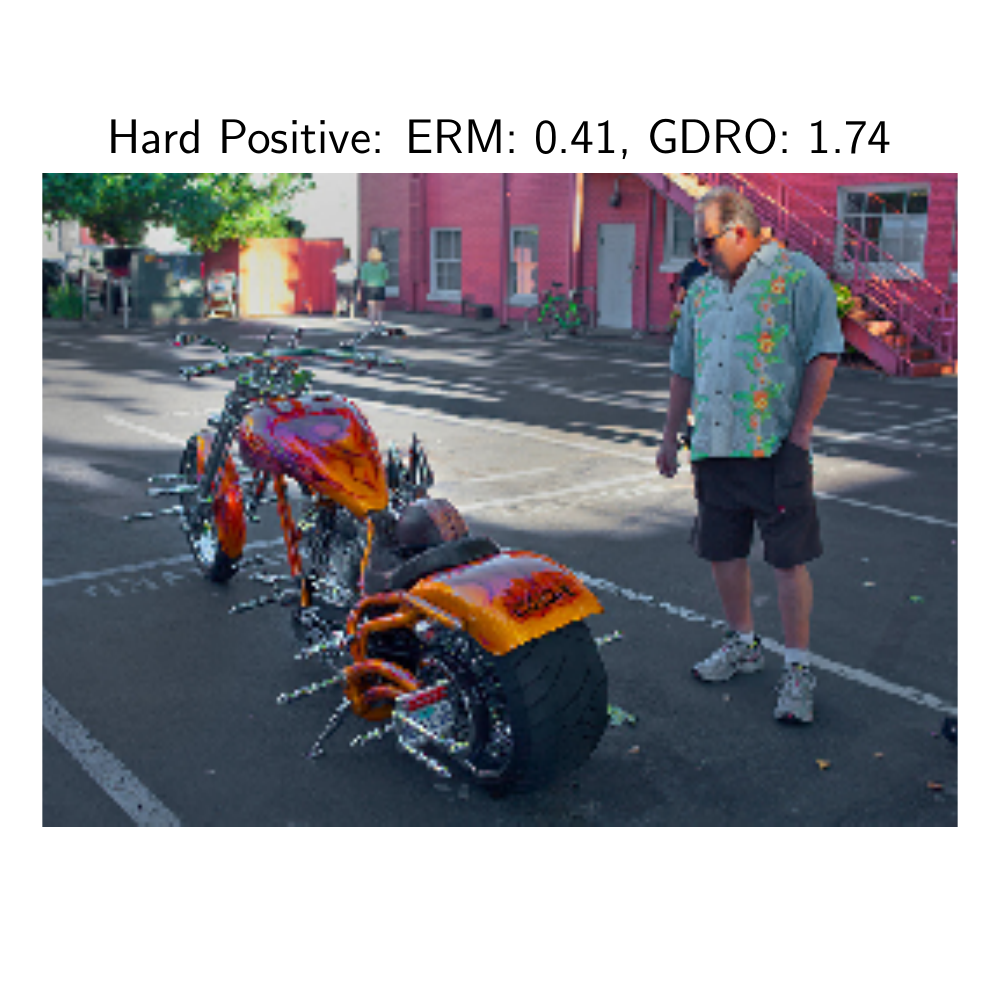}
% \caption{First subfigure} \label{fig:a}
\end{subfigure}\hspace*{\fill}
% \hspace*{-3cm}
\begin{subfigure}{0.3\textwidth}
\includegraphics[width=\linewidth]{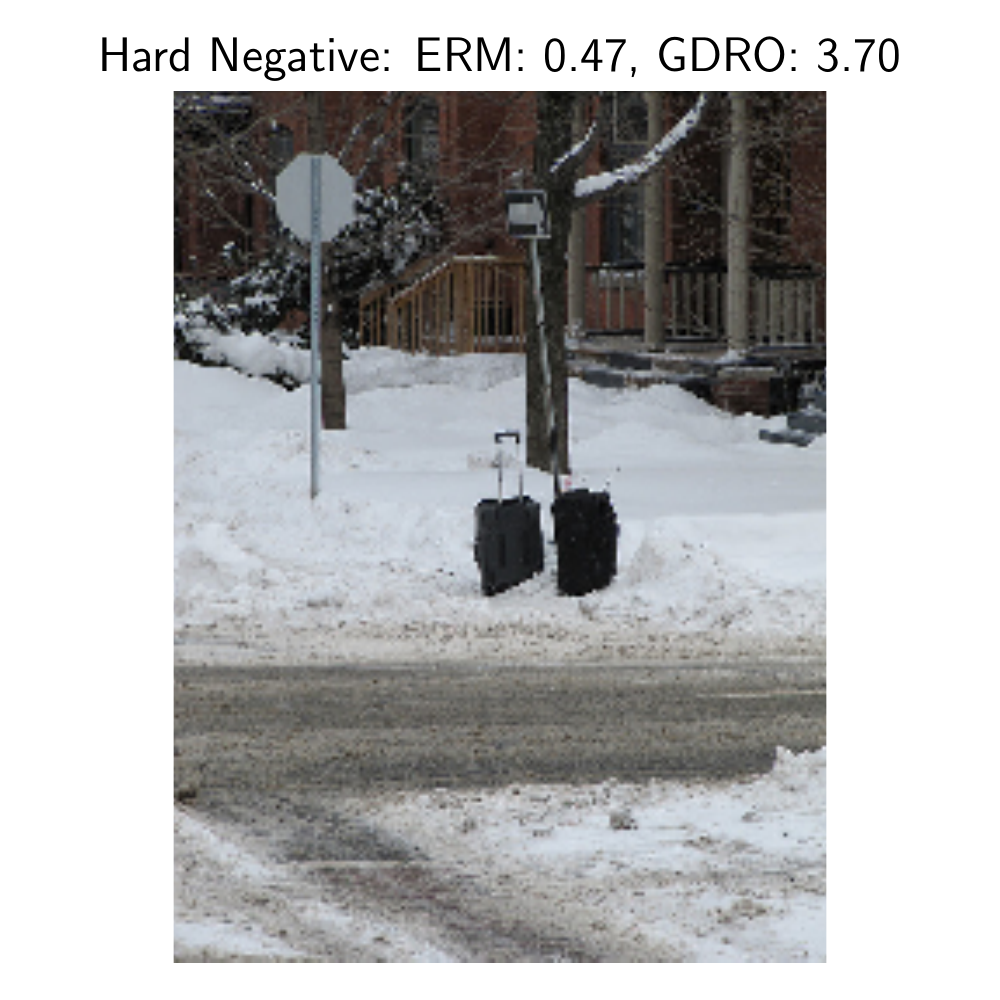}
% \caption{Second subfigure} \label{fig:b}
\end{subfigure}
% \hspace*{\fill}
% \hspace*{-2cm}
% \begin{subfigure}{0.33\textwidth}
% \includegraphics[width=\linewidth]{figs/results/qual_examples/example_loss_diff_ERM_over_GDRO_car_Easy_Positive.pdf}
% % \caption{Second subfigure} \label{fig:b}
% \end{subfigure}

\caption{Test set examples with largest difference in NLL between ERM and GDRO methods, where ERM is better, among (L) hard positives and (R) hard negatives
from the \rococo-CE \texttt{car} category.
Titles show NLL for each method on that image.
% Top row: GDRO better, bottom row: ERM better.
% From L to R: .
} \label{fig:app-results-qual-examples-car}
\end{wrapfigure}

In Fig. \ref{fig:results-qual-examples-car}, we show the images with the largest gaps in NLL on the \texttt{car} task among two groups: hard positives and hard negatives (in \rococo-CE), where GDRO is better than ERM.
% These examples give us some insight into the strengths and weaknesses of the GDRO method.
% The hard positive and hard negative where GDRO most overperformed (top row) are both images we might expect, where the object label and context do not match: a living room with a tiny toy car on the shelf, and a normal street scene that happens to not have any cars.
% in the hard positive, the image is a living room (an unusual context) and the ``car'' is a toy car in the top right hand corner of the image.
% In the hard negative, the image is a normal street scene (usual context) that happens to not have any cars.
% These are both exactly the type of OOC image that we would hope GDRO performs well on.
Here, in Fig. \ref{fig:app-results-qual-examples-car} we look at the images where GDRO most underperforms ERM.
These help us consider some pitfalls of the GDRO objective, which incentivizes performance on two small subgroups: ``cars without roads'' and ``roads without cars''.
% We can also gain some insight by looking at where GDRO most underperforms relative to ERM.
GDRO's worst hard positive is reminiscent of a ``road without a car'': the background (labelled ``pavement'') looks road-like but the focal object is not a prototypical car (the car is in the top left, background).
We hypothesize that GDRO's incentive to perform well on ``roads without cars'' pushed its prediction negative, whereas ERM was able to take advantage of context and correctly output a positive.
% GDRO's worst easy positive is similar: there is clearly a road, but the car is hard to see (small, in the distance), which may encourage a negative prediction (``roads without cars'').
GDRO's worst hard negative tells an opposite story: the snowy background looks \textit{less} road-like, but the stop sign and black objects in the snow suggest a car may be present.
We hypothesize that GDRO's incentive to perform well on ``cars without roads'', may have pushed its prediction wrongly \textit{positive} in this instance.
% Its best hard positive is similar: the car is hard to see (edge of frame), but the context looks sidewalk-like, which may encourage a (correct) positive prediction (``cars without roads'').

% The easy positives where GDRO and ERM most differed also follow this pattern.
% % Finally, we can consider how GDRO may handle borderline cases from looking at the easy positives where GDRO and ERM most differed.
% Both images are borderline as the car is difficult to see (above: edge of the frame; below: small, in the distance), but in the top image (GDRO better), the context looks like a sidewalk rather than a road, which may encourage a positive prediction (``cars without roads'').
% However, in the bottom image, where there is clearly a road, this context may encourage a negative prediction (``roads without cars'').

Qualitatively, it appears that GDRO uses context in potentially more unintuitive ways than ERM does.
While we hope that robust methods will be more interpretable since they may ignore irrelevant signals in the training data \citep{ross2018improving,noack2021empirical,hamon2020robustness}, these two goals may sometimes be at odds.
Indeed, we observe that robust objectives may \textit{unnecessarily} encourage predictions which diverge from the context.
% In this sense, we can think of this as an analogue of ``excessive invariance'' \citep{jacobsen2018excessive} --- the model becomes overly invariant to certain types of context shifts.

\end{document}